%% file: paper.tex
\pdfoutput=1 
\documentclass[11pt,a4paper]{article}
\usepackage[utf8]{inputenc}

\usepackage[hyperref]{emnlp2020}
\aclfinalcopy
\title{Learning Context-Free Languages with Nondeterministic Stack RNNs}

\author{Brian DuSell \\
  University of Notre Dame \\
  \texttt{bdusell1@nd.edu} \\
  \And
  David Chiang \\
  University of Notre Dame \\
  \texttt{dchiang@nd.edu}}
  
\date{}

\usepackage{latexsym}

\usepackage{microtype}
\usepackage{amsmath}
\usepackage{mathtools}
\usepackage{amssymb}

\usepackage{txfonts}

\usepackage{tikz}
\usetikzlibrary{automata,arrows,shapes,calc,positioning,patterns}
\tikzset{every edge/.style={draw,->,>=stealth',shorten >=1pt,auto,semithick}}
\tikzset{initial text={},double distance=2pt}
\tikzset{state/.append style={minimum size=18pt,inner sep=1pt}}

\usepackage{pgfplots}
\usepgfplotslibrary{groupplots,dateplot}
\pgfplotsset{compat=newest}

\usepackage[noend]{algpseudocode}
\usepackage{algorithm}

\usepackage{graphicx}
\usepackage{paralist}

\usepackage{dsfont}
\newcommand{\indicator}[1]{\mathds{1}[#1]}

\newcommand{\topdist}{\tau}
\newcommand{\outdist}{\mathbf{y}}
\newcommand{\trans}[5]{\ensuremath{#1,#3\xrightarrow{#2} #4,#5}}
\newcommand{\transweight}[5]{\ensuremath{\delta(#1,#3\xrightarrow{#2} #4,#5)}}
\newcommand{\transtensor}[5]{\ensuremath{\Delta[#1][#2,#3\rightarrow #4,#5]}}

\newcommand{\iprev}{i\mathord-1}
\newcommand{\jprev}{j\mathord-1}
\newcommand{\actions}{\text{Act}(\Gamma)}

\DeclareMathOperator*{\join}{{\ooalign{$\bigcirc$\cr\hidewidth$\raisebox{2pt}{\texttt{,}}$\hidewidth\cr}}}
\newcommand{\wt}{\mathrel/}
\newcommand{\rev}{\text{R}}

\newcommand{\ourmodel}{Nondeterministic Stack RNN}

\newcommand{\om}{NS-RNN}

\begin{document}
\maketitle
\begin{abstract}
We present a differentiable stack data structure that simultaneously and tractably encodes an exponential number of stack configurations, based on Lang’s algorithm for simulating nondeterministic pushdown automata. We call the combination of this data structure with a recurrent neural network (RNN) controller a \ourmodel. We compare our model against existing stack RNNs on various formal languages, demonstrating that our model converges more reliably to algorithmic behavior on deterministic tasks, and achieves lower cross-entropy on inherently nondeterministic tasks.
\end{abstract}

\section{Introduction}
\label{sec:introduction}

Although recent neural models of language have made advances in learning syntactic behavior, research continues to suggest that inductive bias plays a key role in data efficiency and human-like syntactic generalization \cite{schijndel+al:2019,hu+al:2020}. Based on the long-held observation that language exhibits hierarchical structure, previous work has proposed coupling recurrent neural networks (RNNs) with differentiable stack data structures \cite{joulin+mikolov:2015,grefenstette+al:2015} to give them some of the computational power of pushdown automata (PDAs), the class of automata that recognize context-free languages (CFLs). However, previously proposed differentiable stack data structures only model deterministic stacks, which store only one version of the stack contents at a time, theoretically limiting the power of these stack RNNs to the deterministic~CFLs.

A sentence's syntactic structure often cannot be fully resolved until its conclusion (if ever), requiring a human listener to track multiple possibilities while hearing the sentence. Past work in psycholinguistics has suggested that models that keep multiple candidate parses in memory at once can explain human reading times better than models which assume harsher computational constraints. This ability also plays an important role in calculating expectations that facilitate more efficient language processing \cite{levy:2008}. Current neural language models do not track multiple parses, if they learn syntax generalizations at all \cite{futrell+al:2018,wilcox+al:2019,mccoy+al:2020}.

We propose a new differentiable stack data structure that explicitly models a nondeterministic PDA, adapting an algorithm by \citet{lang:1974} and reformulating it in terms of tensor operations. The algorithm is able to represent an exponential number of stack configurations at once using cubic time and quadratic space complexity. As with existing stack RNN architectures, we combine this data structure with an RNN controller, and we call the resulting model a \ourmodel{} (\om).

We predict that nondeterminism can help language processing in two ways. First, it will improve trainability, since all possible sequences of stack operations contribute to the objective function, not just the sequence used by the current model. Second, it will improve expressivity, as it is able to model concurrent parses in ways that a deterministic stack cannot. We demonstrate these claims by comparing the \om{} to deterministic stack RNNs on formal language modeling tasks of varying complexity. To show that nondeterminism aids training, we show that the \om{} achieves lower cross-entropy, in fewer parameter updates, on some deterministic CFLs. To show that nondeterminism improves expressivity, we show that the \om{} achieves lower cross-entropy on nondeterministic CFLs, including the ``hardest context-free language" \cite{greibach:1973}, a language which is at least as difficult to parse as any other CFL and inherently requires nondeterminism. Our code is available at \url{https://github.com/bdusell/nondeterministic-stack-rnn}.

\section{Background and Motivation}

In all differentiable stack-augmented networks that we are aware of (including ours), a network called the \emph{controller}, which is some kind of RNN (typically an LSTM), is augmented with a differentiable stack, which has no parameters of its own. At each time step, the controller emits weights for various stack operations, which at minimum include push and pop. To maintain differentiability, the weights need to be continuous; different designs for the stack interpret fractionally-weighted operations differently. The stack then executes the fractional operations and produces a stack \emph{reading}, which is a vector that represents the top of the updated stack. The stack reading is used as an extra input to the next hidden state update.

Designs for differentiable stacks have proceeded generally along two lines. One approach, which we call \emph{superposition} \citep{joulin+mikolov:2015}, treats fractional weights as probabilities. The other, which we call \emph{stratification} \citep{sun+al:1995,grefenstette+al:2015}, treats fractional weights as ``thicknesses.''

\paragraph{Superposition}
In the model of \citet{joulin+mikolov:2015}, the controller emits at each time step a probability distribution over three stack operations: push a new vector, pop the top vector, and no-op. The stack simulates all three operations at once, setting each stack element to the weighted interpolation of the elements above, at, and below it in the previous time step, weighted by push, no-op, and pop probabilities respectively. Thus, each stack element is a superposition of possible values for that element. Because stack elements depend only on a fixed number of elements from the previous time step, the stack update can largely be parallelized. \Citet{yogatama+al:2018} developed an extension to this model that allows a variable number of pops per time step, up to a fixed limit $K$. \Citet{suzgun+:2019} also proposed a modification of the controller parameterization.

\paragraph{Stratification}
The model proposed by \citet{sun+al:1995} and later studied by \citet{grefenstette+al:2015} takes a different approach, assigning a \textit{strength} between 0 and 1 to each stack element. If the stack elements were the layers of a cake, then the strengths would represent the thickness of each layer. At each time step, the controller emits a push weight between 0 and 1 which determines the strength of a new vector pushed onto the stack, and a pop weight between 0 and 1 which determines how much to slice off the top of the stack. The stack reading is computed by examining the top layer of unit thickness and interpolating the vectors proportional to their strengths. This relies on $\min$ and $\max$ operations, which can have zero gradients. In practice, the model can get trapped in local optima and requires random restarts \cite{hao+al:2018}. This model also affords less opportunity for parallelization because of the interdependence of stack elements within the same time step. \Citet{hao+al:2018} proposed an extension that uses memory buffers to allow variable-length transductions.

\paragraph{Nondeterminism}

In all the above models, the stack is essentially deterministic in design. In order to recognize a nondeterministic CFL like $\{ww^\rev\}$ from left to right, it must be possible, at each time step, for the stack to track all prefixes of the input string read so far. None of the foregoing models, to our knowledge, can represent a set of possiblities like this. Even for deterministic CFLs, this has consequences for trainability; at each time step, training can only update the model from the vantage point of a single stack configuration, making the model prone to getting stuck in local minima.

To overcome this weakness, we propose incorporating a nondeterministic stack, which affords the model a global view of the space of possible ways to use the stack. Our controller emits a probability distribution over stack operations, as in the superposition approach. However, whereas superposition only maintains the per-element marginal distributions over the stack elements, we propose to maintain the full distribution over the whole stack contents. We marginalize the distribution as late as possible, when the controller queries the stack for the current top stack symbol.

In the following sections, we explain our model and compare it against those of \citet{joulin+mikolov:2015} and \citet{grefenstette+al:2015}. Despite taking longer in wall-clock time to train, our model learns to solve the tasks optimally with a higher rate of success.

\section{Pushdown Automata}

In this section, we give a definition of nondeterministic PDAs (\S\ref{sec:pdadef}), 
describe how to process strings with nondeterministic PDAs in cubic time (\S\ref{sec:lang}), and reformulate this algorithm in terms of tensor operations (\S\ref{sec:tensor}).

\subsection{Notation}

Let $\epsilon$ be the empty string. Let $\indicator{\phi}$ be $1$ when proposition $\phi$ is true, $0$ otherwise. 
If $A$ is a matrix, let $A_{i:}$ and $A_{:j}$ be the $i$th row and $j$th column, respectively, and define analogous notation for tensors.

\subsection{Definition}
\label{sec:pdadef}

A \textit{weighted pushdown automaton (PDA)} is a tuple $M = (Q, \Sigma, \Gamma, \delta, q_0, \bot
)$, where:
\begin{compactitem}
\item $Q$ is a finite set of states.
\item $\Sigma$ is a finite input alphabet.
\item $\Gamma$ is a finite stack alphabet.
\item $\delta \colon Q
\times \Gamma \times \Sigma \times Q \times \Gamma^\ast \rightarrow \mathbb{R}_{\geq 0}$ maps transitions, which we write as $\trans qaxry$, to weights.
\item $q_0 \in Q$ is the start state.
\item $\bot \in \Gamma$ is the initial stack symbol.
\end{compactitem}
In this paper, we do not allow non-scanning transitions (that is, those where $a = \epsilon$). Although this does not reduce the weak generative capacity of PDAs \citep{autebert+:1997}, it could affect their ability to learn; we leave exploration of non-scanning transitions for future work.

For simplicity, we will assume that all transitions have one of the three forms:
\begin{align*}
&\trans q a x r xy && \text{push $y$ on top of $x$} \\
&\trans q a x r y && \text{replace $x$ with $y$} \\
&\trans q a x r \epsilon && \text{pop $x$.}
\end{align*}
This also does not reduce the weak generative capacity of PDAs.

Given an input string $w \in \Sigma^\ast$ of length $n$, a \emph{configuration} is a triple $(i, q, \beta)$, where $i \in [0, n]$ is an input position indicating that all symbols up to and including $w_i$ have been scanned, $q \in Q$ is a state, and $\beta \in \Gamma^\ast$ is the content of the stack (written bottom to top). For all $i, q, r, \beta, x, y$, we say that $(\iprev, q, \beta x)$ \emph{yields} $(i, r, \beta y)$ if $\transweight q{w_i}xry > 0$. A \emph{run} is a sequence of configurations starting with $(0, q_0, \bot)$ where each configuration (except the last) yields the next configuration.

Because our model does not use the PDA to accept or reject strings, we omit the usual definitions for the language accepted by a PDA. This is also why our definition lacks accept states.

As an example, consider the following PDA, for the language $\{ww^\rev \mid w \in \{\texttt{0}, \texttt{1}\}^\ast\}$:
\begin{align*}
M &= (Q, \Sigma, \Gamma, \delta, q_1, \bot) \\
Q &= \{q_1, q_2\} \\
\Sigma &= \{\texttt{0}, \texttt{1}\} \\
\Gamma &= \{\texttt{0}, \texttt{1}, \bot\}
\end{align*}
where $\delta$ contains the transitions
\begin{align*}
q_1, x &\xrightarrow{a} q_1, xa & x &\in \Gamma, a \in \Sigma \\
q_1, a &\xrightarrow{a} q_2, \epsilon & a &\in \Sigma \\
q_2, a &\xrightarrow{a} q_2, \epsilon & a &\in \Sigma.
\end{align*}
This PDA has a possible configuration with an empty stack ($\bot$) iff the input string read so far is of the form $ww^\rev$.

To make a weighted PDA probabilistic, we require that all transition weights be nonnegative and, for all $a, q, x$:
\begin{align*}
    \displaystyle\sum_{r \in Q} \sum_{y \in \Gamma^\ast} \transweight qaxry &= 1.
\end{align*}
Whereas many definitions make the model generate symbols \citep{abney+al:1999}, our definition makes the PDA operations conditional on the input symbol $a$. The difference is not very important, because the RNN controller will eventually assume responsibility for reading and writing symbols, but our definition makes the shift to an RNN controller below slightly simpler.

\subsection{Recognition}
\label{sec:lang}

\citet{lang:1974} gives an algorithm for simulating all runs of a nondeterministic PDA, related to Earley's algorithm \citep{earley:1970}. At any point in time, there can be exponentially many possibilities for the contents of the stack. In spite of this, Lang's algorithm is able to represent the set of all possibilities using only quadratic space. As this set is regular, its representation can be thought of as a weighted finite automaton, which we call the \emph{stack WFA}, similar to the graph-structured stack used in GLR parsing \cite{tomita:1987}.

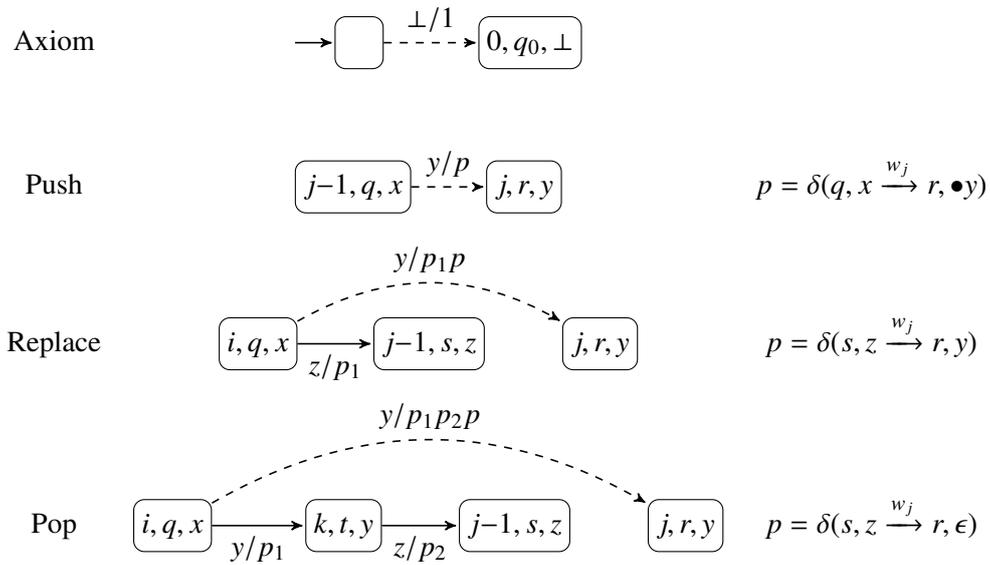
\begin{figure*}
\tikzset{state/.append style={rectangle,rounded corners,inner sep=3pt,anchor=base,execute at begin node={\strut}}}
\tikzset{x=2.25cm,baseline=0}
\renewcommand{\arraystretch}{4}
\begin{center}
\begin{tabular}{ccc}
Axiom & 
\begin{tikzpicture}
\node[initial,state](q) at (0,0) {};
\node[state](r) at (1,0) {$0,q_0,\bot$};
\draw[dashed] (q) edge node {$\bot/1$} (r);
\end{tikzpicture}
& \\
Push & 
\begin{tikzpicture}
\node[state](q1) at (1,0) {$\jprev,q,x$};
\node[state](q2) at (2,0) {$j,r,y$};
\draw[dashed] (q1) edge node {$y / p$} (q2);
\end{tikzpicture} &
$p = \transweight q{w_j}xr{\bullet y}$
\\
Replace &
\begin{tikzpicture}
\node[state](q0) at (0,0) {$i,q,x$};
\node[state](q1) at (1,0) {$\jprev,s,z$};
\node[state](q2) at (2,0) {$j,r,y$};
\draw (q0) edge node[below] {$z / p_1$} (q1);
\draw[dashed,bend left] (q0) edge node {$y / p_1 p$}(q2);
\end{tikzpicture} &
$p = \transweight s{w_j}zry$
\\
Pop &
\begin{tikzpicture}
\node[state](q0) at (0,0) {$i,q,x$};
\node[state](q1) at (1,0) {$k,t,y$};
\node[state](q2) at (2,0) {$\jprev,s,z$};
\node[state](q3) at (3,0) {$j,r,y$};
\draw (q0) edge node[below] {$y / p_1$} (q1);
\draw (q1) edge node[below] {$z / p_2$} (q2);
\draw[dashed,bend left] (q0) edge node {$y / p_1 p_2 p$} (q3);
\end{tikzpicture} &
$p = \transweight s{w_j}zr\epsilon$
\end{tabular}
\end{center}
\caption{Lang's algorithm drawn as operations on the stack WFA. Solid edges indicate existing transitions; dashed edges indicate transitions that are added as a result of the stack operation.}
\label{fig:stackops}
\end{figure*}

Figure~\ref{fig:stackops} depicts Lang's algorithm as a set of inference rules, similar to a deductive parser \citep{shieber+:1995,goodman:1999}, although the visual presentation is rather different. Each inference rule is drawn as a fragment of the stack WFA. If the transitions drawn with solid lines are present in the stack WFA, and the side conditions in the right column are met, then the transition drawn with a dashed line can be added to the stack WFA. The algorithm repeatedly applies inference rules to add states and transitions to the stack WFA; no states or transitions are ever deleted.

\begin{figure*}
\tikzset{state/.append style={rectangle,rounded corners,inner sep=3pt,anchor=base,execute at begin node={\strut}}}
\tikzset{label/.style={anchor=base,execute at begin node={\strut}}}
\tikzset{x=2cm,baseline=0pt,node distance=2cm}
\begin{center}
\begin{tabular}{ll}
$j=0$ &
\begin{tikzpicture}
\node[initial,state] (start) {};
\node[accepting,state,right of=start] (0q1bot) {$0, q_1, \bot$};
\draw[dashed] (start) edge node {$\bot$} (0q1bot);
\end{tikzpicture}
\\[0.5cm]
$j=1$ &
\begin{tikzpicture}
\node[initial,state] (start) {};
\node[state,right of=start] (0q1bot) {$0, q_1, \bot$};
\draw (start) edge node {$\bot$} (0q1bot);
\node[accepting,state,right of=0q1bot](1q10) {$1, q_1, \texttt{0}$};
\draw[dashed] (0q1bot) edge node {$\texttt{0}$} (1q10);
\coordinate (p) at (12cm,0);
\node[label] at (1q10.base -| p) {$\trans{q_1}{\texttt{0}}{\bot}{q_1}{\texttt{0}}$};
\end{tikzpicture}
\\[0.5cm]
$j=2$ &
\begin{tikzpicture}
\node[initial,state] (start) {};
\node[state,right of=start] (0q1bot) {$0, q_1, \bot$};
\draw (start) edge node {$\bot$} (0q1bot);
\node[state,right of=0q1bot](1q10) {$1, q_1, \texttt{0}$};
\draw (0q1bot) edge node {$\texttt{0}$} (1q10);
\node[accepting,state,right of=1q10](2q11) {$2, q_1, \texttt{1}$};
\draw[dashed] (1q10) edge node {$\texttt{1}$} (2q11);
\coordinate (p) at (12cm,0);
\node[anchor=base] at (2q11.base -| p) {$\trans{q_1}{\texttt{1}}{\texttt{0}}{q_1}{\texttt{1}}$};
\end{tikzpicture}
\\[0.5cm]
$j=3$ &
\begin{tikzpicture}
\node[initial,state] (start) {};
\node[state,right of=start] (0q1bot) {$0, q_1, \bot$};
\draw (start) edge node {$\bot$} (0q1bot);
\node[state,right of=0q1bot](1q10) {$1, q_1, \texttt{0}$};
\draw (0q1bot) edge node {$\texttt{0}$} (1q10);
\node[state,right of=1q10](2q11) {$2, q_1, \texttt{1}$};
\draw (1q10) edge node {$\texttt{1}$} (2q11);
\node[accepting,state,right of=2q11](3q11) {$3, q_1, \texttt{1}$};
\draw[dashed] (2q11) edge node {$\texttt{1}$} (3q11);
\node[accepting,state,below=0.5cm of 3q11](3q20) {$3, q_2, \texttt{0}$};
\draw[dashed,out=-30,in=180] (0q1bot) edge node {$\texttt{0}$} (3q20);
\coordinate (p) at (12cm,0);
\node[label] at (3q11.base -| p) {$\trans{q_1}{\texttt{1}}{\texttt{1}}{q_1}{\texttt{1}}$};
\node[label] at (3q20.base -| p) {$\trans{q_1}{\texttt{1}}{\texttt{1}}{q_2}{\epsilon}$};
\end{tikzpicture}
\\[1.7cm]
$j=4$ &
\begin{tikzpicture}
\node[initial,state] (start) {};
\node[state,right of=start] (0q1bot) {$0, q_1, \bot$};
\draw (start) edge node {$\bot$} (0q1bot);
\node[state,right of=0q1bot](1q10) {$1, q_1, \texttt{0}$};
\draw (0q1bot) edge node {$\texttt{0}$} (1q10);
\node[state,right of=1q10](2q11) {$2, q_1, \texttt{1}$};
\draw (1q10) edge node {$\texttt{1}$} (2q11);
\node[state,right of=2q11](3q11) {$3, q_1, \texttt{1}$};
\draw (2q11) edge node {$\texttt{1}$} (3q11);
\node[state,below=0.5cm of 3q11](3q20) {$3, q_2, \texttt{0}$};
\draw[out=-30,in=180] (0q1bot) edge node {$\texttt{0}$} (3q20);
\node[accepting,state,right of=3q11](4q10) {$4, q_1, \texttt{0}$};
\draw[dashed] (3q11) edge node {$\texttt{0}$} (4q10);
\node[accepting,state,below=0.5cm of 4q10](4q2bot) {$4, q_2, \bot$};
\draw[dashed,every edge,rounded corners=5mm] (start) -- ($(start|-4q2bot)+(1.5,-0.75)$) to node {$\bot$} ($(4q2bot)+(-0.5,-0.75)$) -- (4q2bot);
\coordinate (p) at (12cm,0);
\node[label] at (4q10.base -| p) {$\trans{q_1}{\texttt{0}}{\texttt{1}}{q_1}{\texttt{0}}$};
\node[label] at (4q2bot.base -| p) {$\trans{q_2}{\texttt{0}}{\texttt{0}}{q_2}{\epsilon}$};
\end{tikzpicture}
\end{tabular}
\end{center}
\caption{Run of Lang's algorithm on our example PDA and the string $\texttt{0110}$. The PDA transitions used are shown at right.}
\label{fig:lang_example}
\end{figure*}
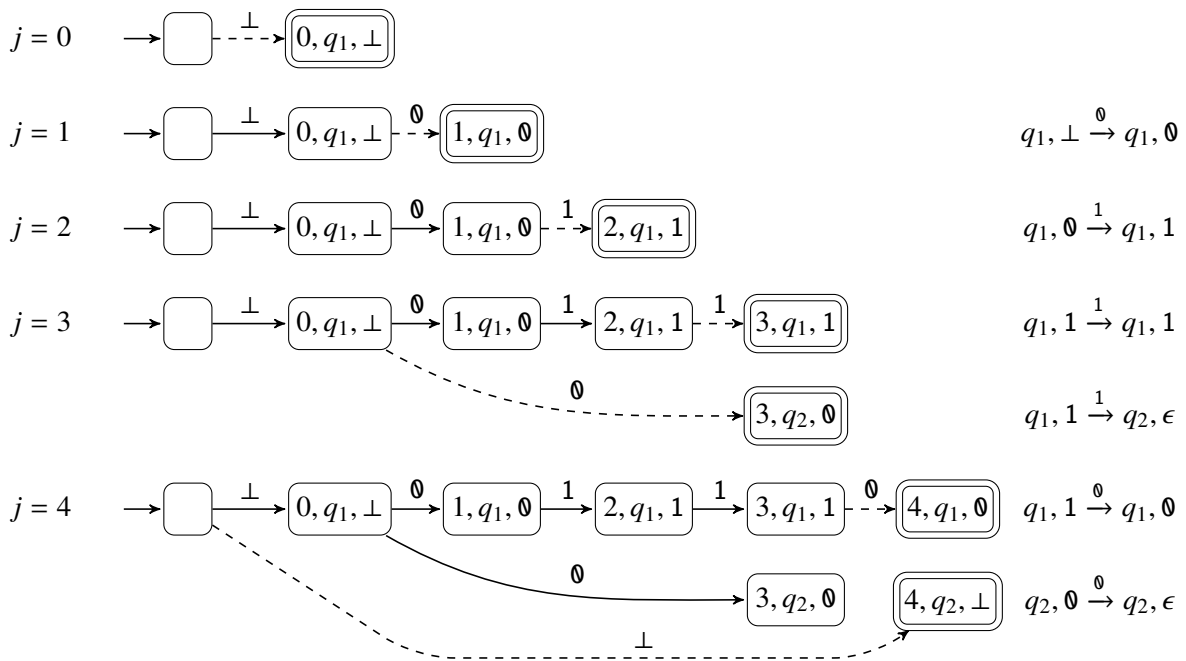

Each state of the stack WFA is of the form $(i, q, x)$, where $i$ is a position in the input string, $q$ is a PDA state, and $x$ is the top stack symbol. We briefly explain each of the inference rules:
\begin{asparadesc}
\item[Axiom] creates an initial state and pushes $\bot$ onto the stack.
\item[Push] pushes a $y$ on top of an $x$. Unlike Lang's original algorithm, this inference rule applies whether or not state $(\jprev, q, x)$ is reachable.
\item[Replace] pops a $z$ and pushes a $y$, by backing up the $z$ transition (without deleting it) and adding a new $y$ transition.
\item[Pop] pops a $z$, by backing up the $z$ transition as well as the preceding $y$ transition (without deleting them) and adding a new $y$ transition.
\end{asparadesc}

The set of accept states of the stack WFA changes from time step to time step; at step $j$, the accept states are $\{(j, q, x) \mid q \in Q, x \in \Gamma\}$. The language recognized by the stack WFA at time $j$ is the set of possible stack contents at time $j$.

An example run of the algorithm is shown in Figure~\ref{fig:lang_example}, using our example PDA and the string $\texttt{0110}$. At time step $j=3$, the PDA reads $\texttt{1}$ and either pushes a $\texttt{1}$ (path ending in state $(3,q_1,\texttt{1})$) or pops a $\texttt{1}$ (path ending in state $(3,q_2,\texttt{0})$). Similarly at time step $j=4$, and the existence of a state with top stack symbol $\bot$ indicates that the string is of the form $ww^\rev$.

The total running time of the algorithm is proportional to the number of ways that the inference rules can be instantiated. Since the Pop rule contains three string positions ($i$, $j$, and $k$), the time complexity is $O(n^3)$. The total space requirement is characterized by the number of possible WFA transitions. Since transitions connect two states, each with a string position ($i$ and $j$), the space complexity is $O(n^2)$.

\subsection{Inner and Forward Weights}
\label{sec:tensor}

To implement this algorithm in a typical neural-network framework, we reformulate it in terms of tensor operations. We use the assumption that all transitions are scanning, although it would be possible to extend the model to handle non-scanning transitions using matrix inversions \citep{stolcke:1995}.

Define $\actions = \bullet\Gamma \cup \Gamma \cup \{\epsilon\}$ to be a set of possible stack actions: if $y \in \Gamma$, then $\bullet y$ means ``push $y$,'' $y$ means ``replace with $y$,'' and $\epsilon$ means ``pop.''

Given an input string $w$, we pack the transition weights of the PDA into a tensor $\Delta$ with dimensions $n \times |Q| \times |\Gamma| \times |Q| \times |\actions|$:
\begin{equation}
\begin{aligned}
\transtensor jqxr{\bullet y} &= \transweight q{w_j}{x}{r}{x y} \\
\transtensor jszry &= \transweight{s}{w_j}{z}{r}{y} \\
\transtensor jszr\epsilon &= \transweight{s}{w_j}{z}{r}{\epsilon}.
\end{aligned}
\label{eq:delta}
\end{equation}

We compute the transition weights of the stack WFA (except for the initial transition) as a tensor of \emph{inner weights} $\gamma$, with dimensions $n \times n \times |Q| \times |\Gamma| \times |Q| \times |\Gamma|$. Each element, which we write as $\gamma[i \xrightarrow{} j][q, x \xrightarrow{} r, y]$, is the weight of the stack WFA transition
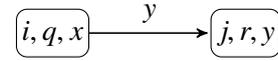
\begin{center}
\begin{tikzpicture}
\tikzset{state/.append style={rectangle,rounded corners,inner sep=2pt}}
\node[state](q) at (0,0) {$i,q,x$};
\node[state](r) at (1in,0) {$j,r,y$};
\draw (q) edge node {$y$} (r);
\end{tikzpicture}
\end{center}
The equations defining $\gamma$ are shown in Figure \ref{fig:equations}. Because these equations are a recurrence relation, we cannot compute $\gamma$ all at once, but (for example) in order of increasing $j$. 
\begin{figure*}
For $1 \leq i < j \leq n$,
\begin{equation*}
\begin{split}
    &\gamma[i \xrightarrow{} j][q, x \xrightarrow{} r, y] = \\
&\qquad \begin{aligned}    
    &\mathds{1}[i=\jprev] \; \transtensor jqxr{\bullet y} && \text{Push} \\
                      & + \sum_{s,z} \gamma[i \xrightarrow{} \jprev][q, x \xrightarrow{} s, z] \; \transtensor jszry && \text{Replace} \\
                      & + \sum_{k=i+1}^{j-2} \sum_{t} \sum_{s,z}  \gamma[i \xrightarrow{} k][q, x \xrightarrow{} t, y] \; \gamma[k \xrightarrow{} \jprev][t, y \xrightarrow{} s, z] \; \transtensor jszr\epsilon && \text{Pop}
\end{aligned}                      
\end{split}
\end{equation*}
    \caption{Equations for computing inner weights.}
    \label{fig:equations}
\end{figure*}

Additionally, we compute a tensor $\alpha$ of \emph{forward weights} of the stack WFA. This tensor has dimensions $n \times |Q| \times |\Gamma|$, and its elements are defined by the recurrence
\begin{align*}
    \alpha[1][r, y] &= \indicator{r = q_0 \wedge y = \bot} \\
    \alpha[j][r, y] &= 
    \begin{multlined}[t]
    \! \sum_{i=1}^{j-1} \sum_{q,x} \alpha[i][q, x] \, \gamma[i \xrightarrow{} j][q, x \xrightarrow{} r, y] \hspace{-6pt} \\
(2 \leq j \leq n).
\end{multlined}
\end{align*}
The weight $\alpha[j][r, y]$ is the total weight of reaching a configuration $(r, j, \beta y)$ for any $\beta$ from the initial configuration, and we can use $\alpha$ to compute the probability distribution over top stack symbols at time step $j$:
\begin{align*}
    \topdist^{(j)}(y) &= \frac{ \sum_r \alpha[j][r, y] }{ \sum_{y'} \sum_r \alpha[j][r, y'] }.
\end{align*}

\section{Neural Pushdown Automata}

Now we couple the tensor formulation of Lang's algorithm for nondeterministic PDAs with an RNN controller.

\subsection{Model}
\label{sec:controllerinterface}

The controller can be any type of RNN; in our experiments, we used a LSTM RNN. At each time step, it computes a hidden vector $\mathbf{h}^{(j)}$ with $d$ dimensions from the previous hidden vector, an input vector $\mathbf{x}^{(j)}$, and the distribution over current top stack symbols, $\tau^{(j)}$, defined above:
\begin{align*}
\textbf{h}^{(j)} &= R\left(\textbf{h}^{(j-1)}, \, \begin{bmatrix} \mathbf{x}^{(j)} \\ \topdist^{(j)} \end{bmatrix} \right) \\
\intertext{where $R$ can be any RNN unit. This state is used to compute an output vector $\outdist^{(j)}$ as usual:}
\outdist^{(j)} &= \text{softmax}\left(\mathbf{A} \mathbf{h}^{(j)} + \mathbf{b}\right) \\
\intertext{where $\mathbf{A}$ and $\mathbf{b}$ are parameters with dimensions $|\Sigma| \times d$ and $|\Sigma|$, respectively. In addition, the state is used to compute a conditional distribution over actions, $\Delta[j]$:}
\mathbf{z}^{(j)}_{qxry} &= \exp\left(\mathbf{C}_{qxry:} \mathbf{h}^{(j)} + \mathbf{D}_{qxry}\right) \\
\transtensor jqxry &= \frac{\mathbf{z}^{(j)}_{qxry}}{\sum_{r',y'} \mathbf{z}^{(j)}_{qxr'y'}}
\end{align*}
where $\mathbf{C}$ and $\mathbf{D}$ are tensors of parameters with dimensions $|Q| \times |\Gamma| \times |Q| \times |\actions| \times d$ and $|Q| \times |\Gamma| \times |Q| \times |\actions|$, respectively. (This is just an affine transformation followed by a softmax over $r$ and $y$.)
These equations replace equations~(\ref{eq:delta}).

\subsection{Implementation}

We implemented the \om{} using PyTorch \citep{pytorch}, and doing so efficiently required a few crucial tricks. The first was a workaround to update the $\gamma$ and $\alpha$ tensors in-place in a way that was compatible with PyTorch's automatic differentiation; this was necessary to achieve the theoretical quadratic space complexity. The second was an efficient implementation of a differentiable \texttt{einsum} operation\footnote{\url{https://github.com/bdusell/semiring-einsum}} that supports the log semiring (as well as other semirings), which allowed us to implement the equations of Figure \ref{fig:equations} in a reasonably fast, memory-efficient way that avoids underflow. Our \texttt{einsum} implementation splits the operation into fixed-size blocks where the multiplication and summation of terms can be fully parallelized. This enforces a reasonable upper bound on memory usage while suffering only a slight decrease in speed compared to fully parallelizing the entire \texttt{einsum} operation.

\section{Experiments}

In this section, we describe our experiments comparing our \om{} and three baseline language models on several formal languages.

\subsection{Tasks}

\begin{asparadesc}
    \item[Marked reversal] The language of palindromes with an explicit middle marker, with strings of the form $w\texttt{\#}w^\rev$, where $w \in \{ \texttt{0}, \texttt{1} \}^{*}$. This task should be easily solvable by a model with a deterministic stack, as the model can push the string $w$ to the stack, change states upon reading $\texttt{\#}$, and predict $w^\rev$ by popping $w$ from the stack in reverse.
    \item[Unmarked reversal] The language of (even-length) palindromes without a middle marker, with strings of the form $ww^\rev$, where $w \in \{ \texttt{0}, \texttt{1} \}^{*}$. When the length of $w$ can vary, a language model reading the string from left to right must use nondeterminism to guess where the boundary between $w$ and $w^\rev$ lies. At each position, it must either push the input symbol to the stack, or else guess that the middle point has been reached and start popping symbols from the stack. An optimal language model will interpolate among all possible split points to produce a final prediction.
    \item[Padded reversal] Like the unmarked reversal language, but with a long stretch of repeated symbols in the middle, with strings of the form $wa^pw^\rev$, where $w \in \{ \texttt{0}, \texttt{1} \}^{*}$, $a \in \{ \texttt{0}, \texttt{1} \}$, and $p \geq 0$. The purpose of the padding is to confuse a language model attempting to guess where the middle of the palindrome is based on the content of the string. In the general case of unmarked reversal, a language model can disregard split points where a valid palindrome does not occur locally. Since all substrings of $a^p$ are palindromes, the language model must deal with a larger number of candidates simultaneously.
    \item[Dyck language] The language $D_2$ of strings with two kinds of balanced brackets.
    \item[Hardest CFL] Designed by \citet{greibach:1973} to be at least as difficult to parse as any other CFL:
\begin{equation*}
\begin{split}
L_0 &= \{ x_1 \texttt{,} y_1 \texttt{,} z_1 \texttt{;} \cdots x_n \texttt{,} y_n \texttt{,} z_n \texttt{;} \mid {} \\
&\qquad n \geq 0, \\
&\qquad y_1 \cdots y_n \in \texttt{\$}D_2, \\
&\qquad x_i, z_i \in \{\texttt{,}, \texttt{\$}, \texttt{(}, \texttt{)}, \texttt{[}, \texttt{]}\}^\ast \}.
\end{split}
\end{equation*}
Intuitively, $L_0$ contains strings formed by dividing a member of $\texttt{\$}D_2$ into pieces ($y_i$) and interleaving them with ``decoy'' pieces (substrings of $x_i$ and $z_i$). While processing the string, the machine has to nondeterministically guess whether each piece is genuine or a decoy. Greibach shows that for any CFL $L$, there is a string homomorphism $h$ such that a parser for $L_0$ can be run on $h(w)$ to find a parse for $w$. See Appendix~\ref{sec:hardest_cfl} for more information.
\end{asparadesc}

\subsection{Data}

For each task, we construct a probabilistic context-free grammar (PCFG) for the language (see Appendix \ref{sec:grammars} for the full grammars and their parameters). We then randomly sample a training set of 10,000 examples from the PCFG, filtering samples so that the length of a string is in the interval $[40, 80]$ (see Appendix \ref{sec:lengthsample} for our sampling method). The training set remains the same throughout the training process and is not re-sampled from epoch to epoch, since we want to test how well the model can infer the probability distribution from a finite sample. 

We sample a validation set of 1,000 examples from the same distribution and a test set with string lengths varying from 40 to 100, with 100 examples per length. The validation set is randomized in each experiment, but for each task, the test set remains the same across all models and random restarts. For simplicity, we do not filter training samples from the validation or test sets, assuming that the chance of overlap is very small.

\subsection{Evaluation}
\label{sec:evaluation}

Since, in these languages, the next symbol cannot always be predicted deterministically from previous symbols, we do not use prediction accuracy as in previous work. Instead, we compute per-symbol cross-entropy on a set of strings $S$. Let $p$ be any distribution over strings; then:
\begin{align*}
    H(S, p) &= \frac{\sum_{w \in S} -\log p(w)}{\sum_{w \in S} |w|}.
\end{align*}
We compute the cross-entropy for both the stack RNN and the distribution from which $S$ is sampled and report the difference. This can be seen as an approximation of the KL divergence of the stack RNN from the true distribution. 

Technically, because the RNN models do not predict the end of the string, they estimate $p(w \mid |w|)$, not $p(w)$. However, they do not actually use any knowledge of the length, so it seems reasonable to compare the RNN's estimate of $p(w \mid |w|)$ with the true $p(w)$. (This is why, when we bin by length in Figure~\ref{fig:test}, some of the differences are negative.)

A benefit of using cross-entropy instead of prediction accuracy is that we can easily incorporate new tasks as long as they are expressed as a PCFG. We do not, for example, need to define a language-dependent subsequence of symbols to evaluate on.

\subsection{Baselines}

We compare our \om{} against three baselines: an LSTM, the Stack LSTM of \citet{joulin+mikolov:2015} (``JM"), and the Stack LSTM of \citet{grefenstette+al:2015} (``Gref"). We deviate slightly from the original definitions of these models in order to standardize the controller-stack interface to the one defined in Section \ref{sec:controllerinterface}, and to isolate the effects of differences in the stack data structure, rather than the controller mechanism. For all three stack models, we use an LSTM controller whose initial hidden state is fixed to 0, and we use only one stack for the JM and Gref models. (In early experiments, we found that using multiple stacks did not make a meaningful difference in performance.) For JM, we include a bias term in the layers that compute the stack actions and network output. We do allow the no-op operation, and the stack reading consists of only the top stack cell. For Gref, we set the controller output~$\mathbf{o}'_t$ equal to the hidden state $\mathbf{h}_t$, so we compute the stack actions, pushed vector, and network output directly from the hidden state. We encode all input symbols as one-hot vectors; there are no embedding layers.

\subsection{Hyperparameters}

For all models, we use a single-layer LSTM with 20 hidden units. We selected this number because we found that an LSTM of this size could not completely solve the marked reversal task, indicating that the hidden state is a memory bottleneck. For each task, we perform a hyperparameter grid search for each model. We search for the initial learning rate, which has a large impact on performance, from the set $\{0.01, 0.005, 0.001, 0.0005\}$. For JM and Gref, we search for stack embedding sizes in $\{2, 20, 40\}$. We manually choose a small number of PDA states and stack symbol types for the \om{} for each task. For marked reversal, unmarked reversal, and Dyck, we use 2 states and 2 stack symbol types. For padded reversal, we use 3 states and 2 stack symbol types. For the hardest CFL, we use 3 states and 3 stack symbol types.

As noted by \citet{grefenstette+al:2015}, initialization can play a large role in whether a Stack LSTM converges on algorithmic behavior or becomes trapped in a local optimum. To mitigate this, for each hyperparameter setting in the grid search, we run five random restarts and select the hyperparameter setting with the lowest average difference in cross entropy on the validation set. This gives us a picture not only of the model's performance, but of its rate of success. We initialize all fully-connected layers except for the recurrent LSTM layer with Xavier uniform initialization \citep{glorot+bengio:2010}, and all other parameters uniformly from $[-0.1, 0.1]$.

We train all models with Adam \citep{kingma+ba:2015} and clip gradients whose magnitude is above~5. We use mini-batches of size~10; to generate a batch, we first select a length and then sample~10 strings of that length. We train models until convergence, multiplying the learning rate by 0.9 after~5 epochs of no improvement in cross-entropy on the validation set, and stopping after 10 epochs of no improvement.

{
\definecolor{color0}{rgb}{0.12156862745098,0.466666666666667,0.705882352941177}
\definecolor{color1}{rgb}{1,0.498039215686275,0.0549019607843137}
\definecolor{color2}{rgb}{0.172549019607843,0.627450980392157,0.172549019607843}
\definecolor{color3}{rgb}{0.83921568627451,0.152941176470588,0.156862745098039}
\pgfplotsset{lines/.style={semithick}}
\pgfplotsset{line0/.style={lines, color0, mark=triangle*, mark options={rotate=30}}}
\pgfplotsset{line1/.style={lines, color1, mark=triangle*, mark options={rotate=120}}}
\pgfplotsset{line2/.style={lines, color2, mark=triangle*, mark options={rotate=210}}}
\pgfplotsset{line3/.style={lines, color3, mark=triangle*, mark options={rotate=300}}}
\tikzset{bars/.style={opacity=0.12}}

\pgfplotsset{every axis/.style={
    width=3.5in,height=2.4in,
    title style={yshift=-4.5ex},
    legend cell align={left},
    legend style={at={(0.5,-0.33)},anchor=north,draw=none,/tikz/every even column/.append style={column sep=0.4cm}},
    legend columns=-1,
    tick style={color=black},
    tick align=outside,
    tick pos=left,
    ymin=0,ytick distance=0.1,
    scaled ticks=false,
    ticklabel style={/pgf/number format/fixed,/pgf/number format/precision=5},
    ylabel style={at={(axis description cs:-0.15,0.5)}},
}}

\begin{figure*}
\begin{minipage}[t]{\columnwidth}
    \centering
    \pgfplotsset{every axis/.append style={
        xmin=0,xmax=160,xtick distance=50,
        mark repeat=32,
    }}
    \pgfplotsset{line0/.append style={mark phase=0}}
    \pgfplotsset{line1/.append style={mark phase=8}}
    \pgfplotsset{line2/.append style={mark phase=16}}
    \pgfplotsset{line3/.append style={mark phase=24}}
    \tikzset{linelabel/.style={black,inner sep=2pt,font={\footnotesize}}}
    {\pgfplotsset{every axis/.append style={xticklabels={,,}}}
    \scalebox{0.8}{\input{figures/train-marked-reversal.tex}} \\
    \scalebox{0.8}{\input{figures/train-unmarked-reversal.tex}} \\
    \scalebox{0.8}{\input{figures/train-padded-reversal.tex}} \\
    \scalebox{0.8}{\input{figures/train-dyck.tex}} \\
    }
    \scalebox{0.8}{\input{figures/train-hardest-cfl.tex}}
    \caption{Cross-entropy difference in nats between model and source distribution on validation set, as a function of training time. Lines are averages of five random restarts, and shaded regions are standard deviations. After a random restart converges, the value of its last epoch is used in the average for later epochs.}
    \label{fig:train}
\end{minipage}%
\hspace{\columnsep}%
\begin{minipage}[t]{\columnwidth}
    \centering
    \pgfplotsset{every axis/.append style={
        xmin=40, xmax=112,
        xtick={40,60,80,100},
        mark repeat=8,
    }}
    \pgfplotsset{line0/.append style={mark phase=0}}
    \pgfplotsset{line1/.append style={mark phase=2}}
    \pgfplotsset{line2/.append style={mark phase=4}}
    \pgfplotsset{line3/.append style={mark phase=6}}
    \tikzset{linelabel/.style={black,inner sep=2pt,font={\footnotesize}}}
    {\pgfplotsset{every axis/.append style={xticklabels={,,}}}
    \scalebox{0.8}{\input{figures/test-marked-reversal.tex}} \\
    \scalebox{0.8}{\input{figures/test-unmarked-reversal.tex}} \\
    \scalebox{0.8}{\input{figures/test-padded-reversal.tex}} \\
    \scalebox{0.8}{\input{figures/test-dyck.tex}} \\
    }
    \scalebox{0.8}{\input{figures/test-hardest-cfl.tex}}
    \caption{Cross-entropy difference in nats on the test set, binned by string length. Some models achieve a negative difference, for reasons explained in \S\ref{sec:evaluation}. Each line is the average of the same five random restarts shown in Figure~\ref{fig:train}.}
    \label{fig:test}
\end{minipage}
\end{figure*}

}

\section{Results}

We show plots of the difference in cross entropy on the validation set between each model and the source distribution in Figure \ref{fig:train}. For all tasks, stack-based models outperform the LSTM baseline, indicating that the tasks are effective benchmarks for differentiable stacks. For the marked reversal, unmarked reversal, and hardest CFL tasks, our model consistently achieves cross-entropy closer to the source distribution than any other model. Even for the marked reversal task, which can be solved deterministically, the \om{}, besides achieving lower cross-entropy on average, learns to solve the task in fewer updates and with much higher reliability across random restarts. In the case of the mildly nondeterministic unmarked reversal and highly nondeterministic hardest CFL tasks, the \om{} converges on the lowest validation cross-entropy. On the Dyck language, which is a deterministic task, all stack models converge quickly on the source distribution. We hypothesize that this is because the Dyck language represents a case where stack usage is locally advantageous everywhere, so it is particularly conducive for learning stack-like behavior. On the other hand, we note that our model struggles on padded reversal, in which stack-friendly signals are intentionally made very distant. Although the \om{} outperforms the LSTM baseline, the JM model solves the task most effectively, though still imperfectly.

In order to show how each model performs when evaluated on strings longer than those seen during training, in Figure \ref{fig:test}, we show cross-entropy on separately sampled test data as a function of string length. All test sets are identical across models and random restarts, and there are 100 samples per length. The \om{} consistently does well on string lengths it was trained on, but it is sometimes surpassed by other stack models on strings that are outside the distribution of lengths it was trained on. This suggests that the \om{} conforms more tightly to the real distribution seen during training.

\section{Conclusion}

We presented the \om{}, a neural language model with a differentiable stack that explicitly models nondeterminism. We showed that it offers improved trainability and modeling power over previous stack-based neural language models; the \om{} learns to solve some deterministic tasks more effectively than other stack-LSTMs, and achieves the best results on a challenging nondeterministic context-free language. However, we note that the \om{} struggled on a task where signals in the data were distant, and did not generalize to longer lengths as well as other stack-LSTMs; we hope to address these shortcomings in future work. We believe that the \om{} will prove to be a powerful tool for learning and modeling ambiguous syntax in natural language.

\section*{Acknowledgements}

This research was supported in part by a Google Faculty Research Award. We would like to thank Justin DeBenedetto and Darcey Riley for their helpful comments, and the Center for Research Computing at the University of Notre Dame for providing the computing infrastructure for our experiments.

\bibliographystyle{acl_natbib}
\bibliography{references}

\clearpage

\appendix

\section{The Hardest CFL}
\label{sec:hardest_cfl}

\Citet{greibach:1973} describes a CFL, $L_0$, which is the ``hardest" CFL in the sense that an efficient parser for $L_0$ is also an efficient parser for any other CFL $L$. It is defined as follows. (We deviate from Greibach's original notation for the sake of clarity.) Every string in $L_0$ is of the following form:
\begin{align*}
    \alpha_1 \texttt{;} \alpha_2 \texttt{;} \cdots \alpha_n \texttt{;} \in L_0
\end{align*}
that is, a sequence of strings $\alpha_i$, each terminated by~$\texttt{;}$. No $\alpha_i$ can contain $\texttt{;}$. Each $\alpha_i$, in turn, is divided into three parts, separated by commas:
\begin{align*}
    \alpha_i = x_i \texttt{,} y_i \texttt{,} z_i
\end{align*}
The middle part, $y_i$, is a substring of a string in $D_2$. The brackets in $y_i$ do not need to be balanced, but all of the $y_i$'s concatenated must form a string in $D_2$, prefixed by $\texttt{\$}$. The catch is that $x_i$ and $z_i$ can be any sequence of bracket, comma, and $\texttt{\$}$ symbols, so it is impossible to tell, in a single $\alpha_i$, where $y_i$ begins and ends. A parser must nondeterministically guess where each $y_i$ is, and cannot verify a guess until the end of the string is reached.

The design of $L_0$ is justified as follows. Suppose we have a parser for $L_0$ which, as part of its output, identifies the start and end of each $y_i$. Given a CFG~$G$ in Greibach normal form (GNF), we can adapt the parser for $L_0$ to parse $\mathcal{L}(G)$ by constructing a string homomorphism $h$, such that $w \in \mathcal{L}(G)$ iff $h(w) \in L_0$, and the concatenated $y_i$'s encode a leftmost derivation of $w$ under $G$. 

The homomorphism $h$ always exists and can be constructed from $G$ as follows. Let the nonterminals of $G$ be $V = \{A_1, \ldots, A_{|V|}\}$. Recall that in GNF, every rule is of the form $A_i \rightarrow a A_{j_1} \cdots A_{j_m}$ and $S$ does not appear on any right-hand side. Define
\begin{align*}
\text{push}(A_i) &= \texttt{(}\texttt{[}^i\texttt{(} \\
\text{pop}(A_i) &= \begin{cases}
\; \texttt{)}\texttt{]}^i\texttt{)} & A_i \neq S \\
\; \texttt{\$} & A_i = S.
\end{cases}
\end{align*}
We encode each rule of $G$ as
\begin{multline*}
\text{rule}(A_i \rightarrow a A_{j_1} \cdots A_{j_m}) = \\
\text{pop}(A_i) \; \text{push}(A_{j_1}) \cdots \text{push}(A_{j_m}).
\end{multline*}
Finally, we can define $h$ as
\begin{equation*}
h(b) = \left(\join_{(A \rightarrow b\gamma) \in G} \text{rule}(A \rightarrow b\gamma)\right) \texttt{;}
\end{equation*}
where $\join$ concatenates strings together delimited by commas. Then there is a valid string of $y_i$'s iff there is a valid derivation of $w$ with respect to~$G$.

\section{PCFGs for Generating Data}
\label{sec:grammars}

We list here the production rules and weights for the PCFG used for each of our tasks. Let $f(\mu) = 1 - \frac{1}{\mu + 1}$, which is the probability of failure associated with a negative binomial distribution with a mean of $\mu$ failures before one success. For a recursive PCFG rule, a probability of $f(\mu)$ results in an average of $\mu$ applications of the recursive rule.

\subsection{Marked reversal}

We set $\mu = 60$.
\begin{alignat*}{3}
S & \rightarrow \texttt{0} S \texttt{0} && \wt \tfrac{1}{2} f(\mu) \\
S & \rightarrow \texttt{1} S \texttt{1} && \wt \tfrac{1}{2} f(\mu) \\
S & \rightarrow \texttt{\#} && \wt 1 - f(\mu)
\end{alignat*}

\subsection{Unmarked reversal}

We set $\mu = 60$.
\begin{alignat*}{3}
S & \rightarrow \texttt{0} S \texttt{0} && \wt \tfrac{1}{2} f(\mu) \\
S & \rightarrow \texttt{1} S \texttt{1} && \wt \tfrac{1}{2} f(\mu) \\
S & \rightarrow \epsilon && \wt 1 - f(\mu)
\end{alignat*}

\subsection{Padded reversal}

Let $\mu_c$ be the mean length of the reversed content, and let $\mu_p$ be the mean padding length. We set $\mu_c = 60$ and $\mu_p = 30$.
\begin{alignat*}{3}
S & \rightarrow \texttt{0} S \texttt{0} && \wt \tfrac{1}{2} f(\mu_c) \\
S & \rightarrow \texttt{1} S \texttt{1} && \wt \tfrac{1}{2} f(\mu_c) \\
S & \rightarrow T_0 && \wt \tfrac{1}{2} (1 - f(\mu_c)) \\
S & \rightarrow T_1 && \wt \tfrac{1}{2} (1 - f(\mu_c)) \\
T_0 & \rightarrow \texttt{0} T_0 && \wt f(\mu_p) \\
T_0 & \rightarrow \epsilon && \wt 1 - f(\mu_p) \\
T_1 & \rightarrow \texttt{1} T_1 && \wt f(\mu_p) \\
T_1 & \rightarrow \epsilon && \wt 1 - f(\mu_p)
\end{alignat*}

\subsection{Dyck language}

Let $\mu_s$ be the mean number of splits, and let $\mu_n$ be the mean nesting depth. We set $\mu_s = 1$ and $\mu_n = 40$.
\begin{alignat*}{3}
S & \rightarrow ST && \wt f(\mu_s) \\
S & \rightarrow T && \wt 1 - f(\mu_s) \\
T & \rightarrow \texttt{(} S \texttt{)} && \wt \tfrac{1}{2} f(\mu_n) \\
T & \rightarrow \texttt{[} S \texttt{]} && \wt \tfrac{1}{2} f(\mu_n) \\
T & \rightarrow \texttt{(} \texttt{)} && \wt \tfrac{1}{2} (1 - f(\mu_n)) \\
T & \rightarrow \texttt{[} \texttt{]} && \wt \tfrac{1}{2} (1 - f(\mu_n))
\end{alignat*}

\subsection{Hardest CFL}

Let $\mu_c$ be the mean number of commas, $\mu_{sf}$ be the mean short filler length, $\mu_{lf}$ be the mean long filler length, $p_s$ be the probability of a semicolon, $\mu_s$ be the mean number of bracket splits, and $\mu_n$ be the mean bracket nesting depth. We set $\mu_c = 0.5$, $\mu_{sf} = 0.5$, $\mu_{lf} = 2$, $p_s = 0.25$, $\mu_s = 1.5$, and $\mu_n = 3$.
\begin{alignat*}{3}
S' & \rightarrow R \texttt{\$} Q\, S L \texttt{;} && \wt 1 \\
L & \rightarrow L' \texttt{,} U && \wt 1 \\
L' & \rightarrow \texttt{,} V L' && \wt f(\mu_c) \\
L' & \rightarrow \epsilon && \wt 1 - f(\mu_c) \\
R & \rightarrow U \texttt{,} R' && \wt 1 \\
R' & \rightarrow R' V \texttt{,} && \wt f(\mu_c) \\
R' & \rightarrow \epsilon && \wt 1 - f(\mu_c) \\
U & \rightarrow WU && \wt f(\mu_{sf}) \\
U & \rightarrow \epsilon && \wt 1 - f(\mu_{sf}) \\
V & \rightarrow WV && \wt f(\mu_{lf} - 1) \\
V & \rightarrow W && \wt 1 - f(\mu_{lf} - 1) \\
W & \rightarrow \texttt{(} && \wt 0.2 \\
W & \rightarrow \texttt{)} && \wt 0.2 \\
W & \rightarrow \texttt{[} && \wt 0.2 \\
W & \rightarrow \texttt{]} && \wt 0.2 \\
W & \rightarrow \texttt{\$} && \wt 0.2 \\
Q & \rightarrow L \texttt{;} R && \wt p_s \\
Q & \rightarrow \epsilon && \wt 1 - p_s \\
S & \rightarrow SQ\,T && \wt f(\mu_s) \\
S & \rightarrow T && \wt 1 - f(\mu_s) \\
T & \rightarrow \texttt{(} Q\,SQ \texttt{)} && \wt \tfrac{1}{2} f(\mu_n) \\
T & \rightarrow \texttt{[} Q\,SQ \texttt{]} && \wt \tfrac{1}{2} f(\mu_n) \\
T & \rightarrow \texttt{(} Q \texttt{)} && \wt \tfrac{1}{2} (1 - f(\mu_n)) \\
T & \rightarrow \texttt{[} Q \texttt{]} && \wt \tfrac{1}{2} (1 - f(\mu_n))
\end{alignat*}

\section{Sampling Strings with Fixed Length from a PCFG}
\label{sec:lengthsample}

For practical reasons, we restrict strings we sample from PCFGs to those whose lengths lie within a certain interval, say $[\ell_{\mathrm{min}}, \ell_{\mathrm{max}}]$. The lengths of strings sampled randomly from PCFGs tend to have high variance, and we often want data sets to consist of strings of a certain length (e.g. longer strings in the test set than in the training set). 

To do this, we first sample a length $\ell$ uniformly from $[\ell_{\mathrm{min}}, \ell_{\mathrm{max}}]$. Then we use an efficient dynamic programming algorithm to sample strings directly from the distribution of strings in the PCFG with length $\ell$. This algorithm is adapted from an algorithm presented by \citet{aguinaga:2018} for sampling graphs of a specific size from a hyperedge replacement grammar.

The algorithm operates in two phases. The first (Algorithm \ref{alg:table}) computes a table $T$ such that every entry $T[A, \ell]$ contains the total probability of sampling a string from the PCFG with length $\ell$. The second (Algorithm \ref{alg:sample}) uses $T$ to randomly sample a string from the PCFG (using $S$ as the nonterminal parameter $X$), restricted to those with a length of exactly $\ell$.

Let $\mathrm{nonterminals}(\beta)$ be an ordered sequence consisting of the nonterminals in $\beta$. Let $\Call{Compositions}{\ell, n}$ be a function that returns a (possibly empty) list of all compositions of $\ell$ that are of length $n$ (that is, all ordered sequences of $n$ positive integers that add up to $\ell$).

\begin{algorithm}
    \caption{Computing the probability table $T$}
    \begin{algorithmic}[1]
        \Require $G$ has no $\epsilon$-rules or unary rules
        
        \Function{ComputeWeights}{$G, T, X, \ell$}
            \ForAll{rules $X \rightarrow \beta ~ / ~ p$ in $G$}
                \State $N \gets \mathrm{nonterminals}(\beta)$
                \State $\ell' = \ell - |\beta| + |N|$
                \For{$C$ in $\Call{Compositions}{\ell', |N|}$}
                    \State $\displaystyle t[\beta, C] \gets p \times \prod_{i=1}^{|N|} T[N_i, C_{i}]$
                \EndFor
            \EndFor
            \State \Return $t$
        \EndFunction
        \Function{ComputeTable}{$G, n$}
            \For{$\ell$ from $1$ to $n$}
                \ForAll{nonterminals $X$}
                    \State $t \gets \Call{ComputeWeights}{G, T, X, \ell}$
                    \State $\displaystyle T[X, \ell] = \sum_{\beta, C} t[\beta, C]$
                \EndFor
            \EndFor
            \State \Return $T$
        \EndFunction
    \end{algorithmic}
    \label{alg:table}
\end{algorithm}

\begin{algorithm}
    \caption{Sampling a string using $T$}
    \begin{algorithmic}[1]
        \Require $T$ is the output of $\Call{ComputeTable}{G, \ell}$
        \Function{SampleSized}{$G, T, X, \ell$}
            \If{$T[X,\ell] = 0$}
                \State \textbf{error} \label{alg:sample:error}
            \EndIf
            \State $t \gets \Call{ComputeWeights}{G, T, X, \ell}$ 
            \State sample $(\beta, C)$ with probability $\displaystyle\frac{t[\beta, C]}{T[X, \ell]}$
            \State $s \gets \epsilon$
            \State $i \gets 1$
            \For{$j$ from $1$ to $|\beta|$}
                \If{$\beta_j$ is a terminal}
                    \State append $\beta_j$ to $s$
                \Else
                    \State $s' \gets \Call{SampleSized}{G, T, \beta_j, C_i}$ \label{alg:sample:expand}
                    \State append $s'$ to $s$
                    \State $i \gets i + 1$
                \EndIf
            \EndFor
            \State \Return $s$
        \EndFunction
    \end{algorithmic}
    \label{alg:sample}
\end{algorithm}

Because this algorithm only works on PCFGs that are free of $\epsilon$-rules and unary rules, we automatically refactor our PCFGs to remove them before providing them to the algorithm. 

Some of our PCFGs do not generate any strings for certain lengths, which is detected at line \ref{alg:sample:error} of Algorithm \ref{alg:sample}. In this case, we restart the sampling procedure from the beginning. This means that the distribution we are effectively sampling from is as follows. Let $G(w)$ be the probability of $w$ under PCFG $G$, and let $G(\ell)$ be the probability of all strings of length $\ell$, that is,
\[ G(\ell) = \sum_{\mathclap{\text{$w$ s.t. $|w| = \ell$}}} G(w). \]
Then the distribution we are sampling from is
\begin{align*}
    p_{\text{sample}}(w) &= \frac{1}{|\{ \ell \mid G(\ell) > 0 \}|} \frac{G(w)}{G(|w|)}.
\end{align*}

When computing the lower-bound cross-entropy of the validation and test sets, we must compute $p_{\text{sample}}(w)$ for each string $w$. Finding $G(w)$ requires re-parsing $w$ with respect to $G$ and summing the probabilities of all valid parses using the Inside algorithm. We can look up the value of $G(|w|)$ in the table entry $T[S, |w|]$ produced in the sampling algorithm.

\end{document}

%% file: figures/train-marked-reversal.tex
\begin{tikzpicture}

\begin{axis}[
title={Marked Reversal},
ylabel={Difference in Cross Entropy},
ymin=-0.02,
ymax=0.3,
]
\path [draw=color0, fill=color0, bars]
(axis cs:1,0.353673750077879)
--(axis cs:1,0.34754654072987)
--(axis cs:2,0.294524401463921)
--(axis cs:3,0.253353446380917)
--(axis cs:4,0.237543206197081)
--(axis cs:5,0.220343080630412)
--(axis cs:6,0.214432583842409)
--(axis cs:7,0.200377489994699)
--(axis cs:8,0.194427103006061)
--(axis cs:9,0.176919635140158)
--(axis cs:10,0.175627844833888)
--(axis cs:11,0.170580395851693)
--(axis cs:12,0.15431047016515)
--(axis cs:13,0.150728035646473)
--(axis cs:14,0.151283680344691)
--(axis cs:15,0.145588520725096)
--(axis cs:16,0.135069138723789)
--(axis cs:17,0.144670504213307)
--(axis cs:18,0.134955589325975)
--(axis cs:19,0.128439228488949)
--(axis cs:20,0.124721614155629)
--(axis cs:21,0.132534494933052)
--(axis cs:22,0.124850563086242)
--(axis cs:23,0.119579196850507)
--(axis cs:24,0.122437098311523)
--(axis cs:25,0.121323158634233)
--(axis cs:26,0.108900250341376)
--(axis cs:27,0.11285352187155)
--(axis cs:28,0.110548971493658)
--(axis cs:29,0.110931872771361)
--(axis cs:30,0.101872500688099)
--(axis cs:31,0.109991972469736)
--(axis cs:32,0.109047348266205)
--(axis cs:33,0.109993803951562)
--(axis cs:34,0.114109767486245)
--(axis cs:35,0.107353681340196)
--(axis cs:36,0.108505297282248)
--(axis cs:37,0.0998696655475769)
--(axis cs:38,0.0921588176540202)
--(axis cs:39,0.09735578604175)
--(axis cs:40,0.101364915725705)
--(axis cs:41,0.103179084142108)
--(axis cs:42,0.091612850973036)
--(axis cs:43,0.0842346468605437)
--(axis cs:44,0.0971969960995176)
--(axis cs:45,0.0897928399806266)
--(axis cs:46,0.085134648168014)
--(axis cs:47,0.0947177962908353)
--(axis cs:48,0.077972001054773)
--(axis cs:49,0.0875889471858167)
--(axis cs:50,0.0868063231807163)
--(axis cs:51,0.0924492182031165)
--(axis cs:52,0.0784388821301177)
--(axis cs:53,0.0805304855689681)
--(axis cs:54,0.0909635851774984)
--(axis cs:55,0.077893086629202)
--(axis cs:56,0.0783709061000549)
--(axis cs:57,0.0710371877233257)
--(axis cs:58,0.0848573349484985)
--(axis cs:59,0.0714667852733552)
--(axis cs:60,0.0643218148460951)
--(axis cs:61,0.090183160551065)
--(axis cs:62,0.0739927674303491)
--(axis cs:63,0.0824025265699943)
--(axis cs:64,0.0739067838174311)
--(axis cs:65,0.0683320538349124)
--(axis cs:66,0.0706554658900999)
--(axis cs:67,0.0623846813803426)
--(axis cs:68,0.0692381576235189)
--(axis cs:69,0.0693594169041535)
--(axis cs:70,0.0634176863100624)
--(axis cs:71,0.0652421404760835)
--(axis cs:72,0.0667632465114062)
--(axis cs:73,0.062694111458639)
--(axis cs:74,0.0682201706198055)
--(axis cs:75,0.0613142305962104)
--(axis cs:76,0.0617152401458379)
--(axis cs:77,0.0642997727853288)
--(axis cs:78,0.0664024212364341)
--(axis cs:79,0.0693094422565547)
--(axis cs:80,0.0588860583358159)
--(axis cs:81,0.0616089208899019)
--(axis cs:82,0.0635588892240455)
--(axis cs:83,0.0610154864016929)
--(axis cs:84,0.0581302076339238)
--(axis cs:85,0.0547100761595312)
--(axis cs:86,0.0588873629521677)
--(axis cs:87,0.0597512276262564)
--(axis cs:88,0.0637267011616965)
--(axis cs:89,0.0600270486389673)
--(axis cs:90,0.0568646453541526)
--(axis cs:91,0.0593652956011758)
--(axis cs:92,0.0754052876743408)
--(axis cs:93,0.0693623488325427)
--(axis cs:94,0.0604903884590932)
--(axis cs:95,0.0627242651814002)
--(axis cs:96,0.054095351047819)
--(axis cs:97,0.0686316420032564)
--(axis cs:98,0.0645021146415361)
--(axis cs:99,0.0597713466878687)
--(axis cs:100,0.0707970152650292)
--(axis cs:101,0.064389677682854)
--(axis cs:102,0.0626323774824801)
--(axis cs:103,0.074347854836528)
--(axis cs:104,0.0638936594388654)
--(axis cs:105,0.0628707449348732)
--(axis cs:106,0.0639643168776609)
--(axis cs:107,0.0723263279067783)
--(axis cs:107,0.101868780196428)
--(axis cs:107,0.101868780196428)
--(axis cs:106,0.102200918255786)
--(axis cs:105,0.102384615968705)
--(axis cs:104,0.1022122312397)
--(axis cs:103,0.102485490268839)
--(axis cs:102,0.10242695654919)
--(axis cs:101,0.102134561642422)
--(axis cs:100,0.101694605970248)
--(axis cs:99,0.0977921773737023)
--(axis cs:98,0.101505449610438)
--(axis cs:97,0.100322802624342)
--(axis cs:96,0.0971673135788828)
--(axis cs:95,0.0999484800971984)
--(axis cs:94,0.0975503956788043)
--(axis cs:93,0.0997819543462217)
--(axis cs:92,0.103762213530298)
--(axis cs:91,0.0986903583216406)
--(axis cs:90,0.0975868580472429)
--(axis cs:89,0.0980844902861639)
--(axis cs:88,0.0991998103532298)
--(axis cs:87,0.0944689330280121)
--(axis cs:86,0.0954418972961335)
--(axis cs:85,0.0931268005312888)
--(axis cs:84,0.0943875664277658)
--(axis cs:83,0.0966017129307514)
--(axis cs:82,0.096654378799538)
--(axis cs:81,0.0985193816784131)
--(axis cs:80,0.094971899341972)
--(axis cs:79,0.0993626629574361)
--(axis cs:78,0.103564255833216)
--(axis cs:77,0.113379015985574)
--(axis cs:76,0.0969947059582843)
--(axis cs:75,0.0873723374297481)
--(axis cs:74,0.0750548416698376)
--(axis cs:73,0.078636015478961)
--(axis cs:72,0.0712844199946199)
--(axis cs:71,0.0780463390153105)
--(axis cs:70,0.0949461025836982)
--(axis cs:69,0.0919248554604209)
--(axis cs:68,0.0909242278986738)
--(axis cs:67,0.0885954644889681)
--(axis cs:66,0.08531552250597)
--(axis cs:65,0.0814255732152373)
--(axis cs:64,0.101050427198112)
--(axis cs:63,0.0935316540423787)
--(axis cs:62,0.0813275463720995)
--(axis cs:61,0.104081663035348)
--(axis cs:60,0.0961095380124058)
--(axis cs:59,0.100508758517091)
--(axis cs:58,0.114069170864826)
--(axis cs:57,0.0930907391088727)
--(axis cs:56,0.0992797495679904)
--(axis cs:55,0.0999281517735052)
--(axis cs:54,0.123440405744345)
--(axis cs:53,0.0939106024152591)
--(axis cs:52,0.101112784364234)
--(axis cs:51,0.109305415698343)
--(axis cs:50,0.102458806503265)
--(axis cs:49,0.112897689874119)
--(axis cs:48,0.11913817667861)
--(axis cs:47,0.12214845429188)
--(axis cs:46,0.106773833922589)
--(axis cs:45,0.110938885361574)
--(axis cs:44,0.113275911378961)
--(axis cs:43,0.106554272919833)
--(axis cs:42,0.100759424082337)
--(axis cs:41,0.111701479033739)
--(axis cs:40,0.12249789490327)
--(axis cs:39,0.114686288209362)
--(axis cs:38,0.125994387935778)
--(axis cs:37,0.131189739969197)
--(axis cs:36,0.130909066948824)
--(axis cs:35,0.139339304610995)
--(axis cs:34,0.127082057367994)
--(axis cs:33,0.129178053804601)
--(axis cs:32,0.143690580159467)
--(axis cs:31,0.123149705085215)
--(axis cs:30,0.139098315997678)
--(axis cs:29,0.123981630409256)
--(axis cs:28,0.127248546322141)
--(axis cs:27,0.162664057714828)
--(axis cs:26,0.152834718181675)
--(axis cs:25,0.163904475580815)
--(axis cs:24,0.140229910861553)
--(axis cs:23,0.158521761251458)
--(axis cs:22,0.150057608054474)
--(axis cs:21,0.147194662016366)
--(axis cs:20,0.162835673459189)
--(axis cs:19,0.15106525012989)
--(axis cs:18,0.162568340702128)
--(axis cs:17,0.158161676429098)
--(axis cs:16,0.171360042656786)
--(axis cs:15,0.193093594558366)
--(axis cs:14,0.195682107267547)
--(axis cs:13,0.192321691046221)
--(axis cs:12,0.197112772435544)
--(axis cs:11,0.19321781878617)
--(axis cs:10,0.202952171505759)
--(axis cs:9,0.219512880201798)
--(axis cs:8,0.230922276318888)
--(axis cs:7,0.241027740556684)
--(axis cs:6,0.261083463466756)
--(axis cs:5,0.276211609567659)
--(axis cs:4,0.293338154881496)
--(axis cs:3,0.309260560025469)
--(axis cs:2,0.316597572916683)
--(axis cs:1,0.353673750077879)
--cycle;

\path [draw=color1, fill=color1, bars]
(axis cs:1,0.357538792866588)
--(axis cs:1,0.341384095438919)
--(axis cs:2,0.0223259611932708)
--(axis cs:3,-0.0183175006357914)
--(axis cs:4,-0.0188986226846877)
--(axis cs:5,-0.0153276909224273)
--(axis cs:6,-0.0134739320365028)
--(axis cs:7,-0.01448053296697)
--(axis cs:8,-0.0141849020323762)
--(axis cs:9,-0.0152383050059114)
--(axis cs:10,-0.0135216434907315)
--(axis cs:11,-0.0127072585569931)
--(axis cs:12,-0.0121048401000492)
--(axis cs:13,-0.011930136513077)
--(axis cs:14,-0.0118928598891521)
--(axis cs:15,-0.00889237945545536)
--(axis cs:16,-0.0105424188755201)
--(axis cs:17,-0.00890794871397674)
--(axis cs:18,-0.00994628794586168)
--(axis cs:19,-0.00976500740063643)
--(axis cs:20,-0.0103537328496926)
--(axis cs:21,-0.0114649138242207)
--(axis cs:22,-0.00998533016385636)
--(axis cs:23,-0.00812325522504752)
--(axis cs:24,-0.00664553889579277)
--(axis cs:25,-0.00629949846008021)
--(axis cs:26,-0.00991929353608231)
--(axis cs:27,-0.00780222186604036)
--(axis cs:28,-0.00738364297260118)
--(axis cs:29,-0.00501121004315285)
--(axis cs:30,-0.0101227509447322)
--(axis cs:31,-0.00798427399189051)
--(axis cs:32,-0.00642592263257152)
--(axis cs:33,-0.00322182028603189)
--(axis cs:34,-0.00371523426793574)
--(axis cs:35,-0.00325380457973772)
--(axis cs:36,-0.00376716908390502)
--(axis cs:37,-0.00281629994949614)
--(axis cs:38,-0.00307118141728851)
--(axis cs:39,-0.00359069644279095)
--(axis cs:40,-0.0021655319188351)
--(axis cs:41,-0.00231580827534405)
--(axis cs:42,-0.00482276173352846)
--(axis cs:43,-0.00431362861572284)
--(axis cs:44,-0.00442140192226455)
--(axis cs:45,-0.00196312471333192)
--(axis cs:46,-0.0027260627706073)
--(axis cs:47,-0.00134212827028326)
--(axis cs:48,-0.00661439794830205)
--(axis cs:49,-0.00148907647364367)
--(axis cs:50,-0.00298463580232647)
--(axis cs:51,-0.0027371283382432)
--(axis cs:52,-0.00176772011815952)
--(axis cs:53,-0.00102540347012263)
--(axis cs:54,-0.0032761091666831)
--(axis cs:55,-0.00128360356664033)
--(axis cs:56,-0.000745302643055834)
--(axis cs:57,0.000399143697384356)
--(axis cs:58,-0.00335527123330278)
--(axis cs:59,-4.48466127226099e-05)
--(axis cs:60,-0.000952496462175628)
--(axis cs:61,0.000253509470302449)
--(axis cs:62,-0.000673958933381721)
--(axis cs:63,-0.000754975824351183)
--(axis cs:64,-0.00453079319311005)
--(axis cs:65,-0.000341776444592573)
--(axis cs:66,-0.00391235431696366)
--(axis cs:67,-0.000636410801111285)
--(axis cs:68,-0.000118232238622409)
--(axis cs:69,-0.0028461230582653)
--(axis cs:70,-0.000986445960195072)
--(axis cs:71,-0.000431405966494554)
--(axis cs:72,-0.000386762294031472)
--(axis cs:73,-0.000291976143961155)
--(axis cs:74,-0.00204653590206104)
--(axis cs:75,-0.000344062779528122)
--(axis cs:76,-0.000429710846005946)
--(axis cs:77,-0.000600264997658195)
--(axis cs:78,-0.000408192779682168)
--(axis cs:79,-0.000278191576012055)
--(axis cs:80,-0.000265819822706743)
--(axis cs:81,-0.000490304219162477)
--(axis cs:82,-0.000432223167453782)
--(axis cs:83,-0.000265598961460292)
--(axis cs:84,-0.000288171087609965)
--(axis cs:85,-0.000316755820366492)
--(axis cs:86,-0.000260629074353841)
--(axis cs:87,-0.000295397249500735)
--(axis cs:88,-0.000300261261743331)
--(axis cs:89,-0.000395367321395203)
--(axis cs:90,-0.000297368258611756)
--(axis cs:91,-0.000456125968370621)
--(axis cs:92,-0.000268773920675994)
--(axis cs:93,-0.000743088960978001)
--(axis cs:94,-0.000746783964627962)
--(axis cs:95,-0.000267599605176777)
--(axis cs:96,-0.000648223796268783)
--(axis cs:97,-0.000264583911612847)
--(axis cs:98,-0.000657997746851206)
--(axis cs:99,-0.000259911555416811)
--(axis cs:100,-0.000286189915151662)
--(axis cs:101,-0.00114748977044739)
--(axis cs:102,-0.000528576987132233)
--(axis cs:103,-0.000368334026984482)
--(axis cs:104,-0.000509358648196581)
--(axis cs:104,0.070587182247418)
--(axis cs:104,0.070587182247418)
--(axis cs:103,0.0720637913135693)
--(axis cs:102,0.0704287326527048)
--(axis cs:101,0.0862043376275002)
--(axis cs:100,0.0735971059729978)
--(axis cs:99,0.0754441597238037)
--(axis cs:98,0.0695060344426032)
--(axis cs:97,0.0744775273982977)
--(axis cs:96,0.0821220727336203)
--(axis cs:95,0.0742977024483082)
--(axis cs:94,0.0830428840908723)
--(axis cs:93,0.0689971686739007)
--(axis cs:92,0.0761888104956113)
--(axis cs:91,0.0710676046583159)
--(axis cs:90,0.073303179148731)
--(axis cs:89,0.0717196990280285)
--(axis cs:88,0.0772959551046103)
--(axis cs:87,0.0733512884059433)
--(axis cs:86,0.0755636535959818)
--(axis cs:85,0.0776982101651978)
--(axis cs:84,0.0735408378445005)
--(axis cs:83,0.0744121920817462)
--(axis cs:82,0.0713080876047577)
--(axis cs:81,0.0804127895261335)
--(axis cs:80,0.0743987106875282)
--(axis cs:79,0.0738501506486783)
--(axis cs:78,0.0715699772487087)
--(axis cs:77,0.081639934102561)
--(axis cs:76,0.0796279176527714)
--(axis cs:75,0.0724161751452409)
--(axis cs:74,0.0917953236945654)
--(axis cs:73,0.0770650136976528)
--(axis cs:72,0.0789945245776014)
--(axis cs:71,0.0717662722038771)
--(axis cs:70,0.083569780818244)
--(axis cs:69,0.0943294734756945)
--(axis cs:68,0.076508167399674)
--(axis cs:67,0.0815548643885163)
--(axis cs:66,0.0891186670395014)
--(axis cs:65,0.0757758569097662)
--(axis cs:64,0.0936624975829231)
--(axis cs:63,0.0813276207346665)
--(axis cs:62,0.0781360242860999)
--(axis cs:61,0.0730816314094014)
--(axis cs:60,0.0845151853638821)
--(axis cs:59,0.0761043515848936)
--(axis cs:58,0.106861721544664)
--(axis cs:57,0.0715798714970604)
--(axis cs:56,0.0830403482436683)
--(axis cs:55,0.082127958496004)
--(axis cs:54,0.105320621555124)
--(axis cs:53,0.0816550989905659)
--(axis cs:52,0.0931598376571539)
--(axis cs:51,0.0915993670977843)
--(axis cs:50,0.0953727209600094)
--(axis cs:49,0.0898048991112359)
--(axis cs:48,0.12792448997754)
--(axis cs:47,0.0876459957930231)
--(axis cs:46,0.0971387012585411)
--(axis cs:45,0.0950610437221726)
--(axis cs:44,0.10192806900711)
--(axis cs:43,0.111865336648968)
--(axis cs:42,0.123308065911233)
--(axis cs:41,0.098658523355252)
--(axis cs:40,0.0940619639215975)
--(axis cs:39,0.110209914518345)
--(axis cs:38,0.106148125930689)
--(axis cs:37,0.103136488393296)
--(axis cs:36,0.111103791163057)
--(axis cs:35,0.10665763908202)
--(axis cs:34,0.108421083945544)
--(axis cs:33,0.10770531554946)
--(axis cs:32,0.116501613438472)
--(axis cs:31,0.140463599150213)
--(axis cs:30,0.138852825806469)
--(axis cs:29,0.110604955824013)
--(axis cs:28,0.12896403390713)
--(axis cs:27,0.126342027334378)
--(axis cs:26,0.140730340849225)
--(axis cs:25,0.123977128511205)
--(axis cs:24,0.138171810230905)
--(axis cs:23,0.133423121319587)
--(axis cs:22,0.143709678333505)
--(axis cs:21,0.147350851107521)
--(axis cs:20,0.154202371395142)
--(axis cs:19,0.162442477951751)
--(axis cs:18,0.15463682468186)
--(axis cs:17,0.162501664957795)
--(axis cs:16,0.165778293759546)
--(axis cs:15,0.162555824280449)
--(axis cs:14,0.162320087249746)
--(axis cs:13,0.173657823828885)
--(axis cs:12,0.179641334259512)
--(axis cs:11,0.184562386689511)
--(axis cs:10,0.191442393702478)
--(axis cs:9,0.209426580390372)
--(axis cs:8,0.213050784265486)
--(axis cs:7,0.215857973781597)
--(axis cs:6,0.22559851539309)
--(axis cs:5,0.240919670483519)
--(axis cs:4,0.245573838425567)
--(axis cs:3,0.252830529954017)
--(axis cs:2,0.293673109650485)
--(axis cs:1,0.357538792866588)
--cycle;

\path [draw=color2, fill=color2, bars]
(axis cs:1,0.330156787312008)
--(axis cs:1,0.315360898692503)
--(axis cs:2,0.113963986098758)
--(axis cs:3,0.104386392686984)
--(axis cs:4,0.0412567343259562)
--(axis cs:5,0.0278532853004703)
--(axis cs:6,0.0267364497589354)
--(axis cs:7,0.0241708837668635)
--(axis cs:8,-0.0156553065499249)
--(axis cs:9,-0.0134130572104531)
--(axis cs:10,-0.00708485197334668)
--(axis cs:11,-0.0097612970079915)
--(axis cs:12,-0.00566328842594137)
--(axis cs:13,-0.0100690282853787)
--(axis cs:14,-0.0121223750344448)
--(axis cs:15,-0.0151310759706026)
--(axis cs:16,-0.0104429640035183)
--(axis cs:17,-0.0096118563791187)
--(axis cs:18,-0.00981945977108943)
--(axis cs:19,-0.00912079776531745)
--(axis cs:20,-0.00910676601207508)
--(axis cs:21,-0.013664408627846)
--(axis cs:22,-0.0082808818940395)
--(axis cs:23,-0.00788027300173758)
--(axis cs:24,-0.0076077123992658)
--(axis cs:25,-0.0077712769218272)
--(axis cs:26,-0.00649337362872605)
--(axis cs:27,-0.00291378408258473)
--(axis cs:28,-0.00291710150469949)
--(axis cs:29,-0.00421385726965443)
--(axis cs:30,-0.00494572643154242)
--(axis cs:31,-0.00586750443665851)
--(axis cs:32,-0.00637496772882643)
--(axis cs:33,-0.00750708349778111)
--(axis cs:34,-0.00489185795070343)
--(axis cs:35,-0.00397048129749001)
--(axis cs:36,-0.00501909222325041)
--(axis cs:37,0.000747995939570548)
--(axis cs:38,-0.004210419921444)
--(axis cs:39,-0.00397940446252196)
--(axis cs:40,-0.00544173538221691)
--(axis cs:41,-0.00399156687095691)
--(axis cs:42,-0.00503267466413114)
--(axis cs:43,-0.00580268363005819)
--(axis cs:44,0.00113225265632976)
--(axis cs:45,-0.00390936885754144)
--(axis cs:46,-0.00437060912308019)
--(axis cs:47,-0.00532682794639183)
--(axis cs:48,-0.00433528919982641)
--(axis cs:49,-0.00111272940162427)
--(axis cs:50,-0.00196448600482865)
--(axis cs:51,-0.00287424226328053)
--(axis cs:52,-0.00348162141710365)
--(axis cs:53,-0.00205053081741415)
--(axis cs:54,-0.00398551036097437)
--(axis cs:55,-0.00359691019923123)
--(axis cs:56,-0.00507670163628322)
--(axis cs:57,-0.00439414343342537)
--(axis cs:58,-0.00261293475428357)
--(axis cs:59,-0.00291407514951883)
--(axis cs:60,-0.00488011093114372)
--(axis cs:61,-0.00441476857224905)
--(axis cs:62,-0.0025985648001537)
--(axis cs:63,-0.00251211106861816)
--(axis cs:64,-0.00283773770912894)
--(axis cs:65,-0.00252142162503856)
--(axis cs:66,-0.00289875705607717)
--(axis cs:67,-0.00250824653799216)
--(axis cs:68,-0.00234828923519874)
--(axis cs:69,-0.00428371812383212)
--(axis cs:70,-0.00387743254959901)
--(axis cs:71,-0.00357193176134427)
--(axis cs:72,-0.00275371902401372)
--(axis cs:73,-0.00311981502715024)
--(axis cs:74,-0.00279166014054237)
--(axis cs:75,-0.00374679498196258)
--(axis cs:76,-0.00314772018882867)
--(axis cs:77,-0.00241309351278395)
--(axis cs:78,-0.00240583755637726)
--(axis cs:79,-0.0022611427607049)
--(axis cs:80,-0.00250846914528637)
--(axis cs:81,-0.00430480770537901)
--(axis cs:82,-0.00225993709153711)
--(axis cs:83,-0.00232321275811138)
--(axis cs:84,-0.0027640420348772)
--(axis cs:85,-0.00231919787473505)
--(axis cs:86,-0.00243287733180381)
--(axis cs:87,-0.00226048433295109)
--(axis cs:88,-0.00303861875197039)
--(axis cs:89,-0.00267772282817188)
--(axis cs:90,-0.00229736916247591)
--(axis cs:91,-0.00226418930796027)
--(axis cs:92,-0.00227914600508256)
--(axis cs:93,-0.00233987778493074)
--(axis cs:94,-0.00291403934345483)
--(axis cs:95,-0.0023016701449614)
--(axis cs:96,-0.00225974528649566)
--(axis cs:97,-0.00315129168193574)
--(axis cs:98,-0.00243532253868268)
--(axis cs:99,-0.00235713179957321)
--(axis cs:100,-0.00228585615049276)
--(axis cs:101,-0.00235110712830185)
--(axis cs:102,-0.00256568865367045)
--(axis cs:103,-0.00227386102122119)
--(axis cs:104,-0.00239521333856351)
--(axis cs:104,0.0700920260242475)
--(axis cs:104,0.0700920260242475)
--(axis cs:103,0.0675037092403891)
--(axis cs:102,0.0721567725265147)
--(axis cs:101,0.0693873914332505)
--(axis cs:100,0.0679322087862381)
--(axis cs:99,0.0694914418315547)
--(axis cs:98,0.0706483164812254)
--(axis cs:97,0.0769350636451267)
--(axis cs:96,0.0662544939428305)
--(axis cs:95,0.0683713098566043)
--(axis cs:94,0.0752165402199179)
--(axis cs:93,0.0637632818551452)
--(axis cs:92,0.0677079728141792)
--(axis cs:91,0.0657045464541832)
--(axis cs:90,0.0645454515835408)
--(axis cs:89,0.073245925049613)
--(axis cs:88,0.0761431568962141)
--(axis cs:87,0.0666155876137503)
--(axis cs:86,0.0706161025506386)
--(axis cs:85,0.0687744612758572)
--(axis cs:84,0.0740063908256118)
--(axis cs:83,0.06885869319535)
--(axis cs:82,0.0661828930943001)
--(axis cs:81,0.083695202166741)
--(axis cs:80,0.0715370278079574)
--(axis cs:79,0.0659751026509379)
--(axis cs:78,0.0702456763324471)
--(axis cs:77,0.0703478148074052)
--(axis cs:76,0.0769105399464255)
--(axis cs:75,0.0806491577434199)
--(axis cs:74,0.0742386619535583)
--(axis cs:73,0.0767176884997672)
--(axis cs:72,0.0739183064740489)
--(axis cs:71,0.0796220801305893)
--(axis cs:70,0.0813907824352921)
--(axis cs:69,0.08358522427014)
--(axis cs:68,0.0693376678184383)
--(axis cs:67,0.0715345067881488)
--(axis cs:66,0.0750984922846776)
--(axis cs:65,0.0716821228299076)
--(axis cs:64,0.074616040085383)
--(axis cs:63,0.0715781387850361)
--(axis cs:62,0.0724911007993893)
--(axis cs:61,0.0842631911571413)
--(axis cs:60,0.086578856124601)
--(axis cs:59,0.0752168155736604)
--(axis cs:58,0.0726330383127008)
--(axis cs:57,0.0841573306267454)
--(axis cs:56,0.0904089195903538)
--(axis cs:55,0.0800269984583371)
--(axis cs:54,0.0817810215070661)
--(axis cs:53,0.0827608207673961)
--(axis cs:52,0.080701849901769)
--(axis cs:51,0.0743798189996432)
--(axis cs:50,0.0885589335845476)
--(axis cs:49,0.0849215848894649)
--(axis cs:48,0.0883626073907731)
--(axis cs:47,0.0930370679557531)
--(axis cs:46,0.0802877682268028)
--(axis cs:45,0.0815023899092909)
--(axis cs:44,0.122780372253641)
--(axis cs:43,0.102285567272731)
--(axis cs:42,0.0943731701504307)
--(axis cs:41,0.0906177149263995)
--(axis cs:40,0.0974189169426315)
--(axis cs:39,0.0999283556689253)
--(axis cs:38,0.0888043240474564)
--(axis cs:37,0.0918389116434714)
--(axis cs:36,0.104711845020119)
--(axis cs:35,0.0935617173110822)
--(axis cs:34,0.114841322253079)
--(axis cs:33,0.107135463437069)
--(axis cs:32,0.10129185477218)
--(axis cs:31,0.108880488048501)
--(axis cs:30,0.0997646029869813)
--(axis cs:29,0.124852296264203)
--(axis cs:28,0.105937506371327)
--(axis cs:27,0.108557238684535)
--(axis cs:26,0.127470897689258)
--(axis cs:25,0.112998458289906)
--(axis cs:24,0.11906097470332)
--(axis cs:23,0.13689243010462)
--(axis cs:22,0.118234580207794)
--(axis cs:21,0.145018916820665)
--(axis cs:20,0.149273271149229)
--(axis cs:19,0.129844051095108)
--(axis cs:18,0.141655710484114)
--(axis cs:17,0.142628787669325)
--(axis cs:16,0.145814039263896)
--(axis cs:15,0.161250536032775)
--(axis cs:14,0.161773008461992)
--(axis cs:13,0.156136407450448)
--(axis cs:12,0.148582927394413)
--(axis cs:11,0.17497269267231)
--(axis cs:10,0.160856438836505)
--(axis cs:9,0.182290017220841)
--(axis cs:8,0.182145380471844)
--(axis cs:7,0.225862100274339)
--(axis cs:6,0.242842083648941)
--(axis cs:5,0.251487775537332)
--(axis cs:4,0.267666565182244)
--(axis cs:3,0.306919896879491)
--(axis cs:2,0.333427603592662)
--(axis cs:1,0.330156787312008)
--cycle;

\path [draw=color3, fill=color3, bars]
(axis cs:1,0.321308770928158)
--(axis cs:1,0.306410912082113)
--(axis cs:2,0.118822365095419)
--(axis cs:3,0.109058982318633)
--(axis cs:4,0.0396620571883116)
--(axis cs:5,0.0271172957080684)
--(axis cs:6,0.023201402667724)
--(axis cs:7,-0.0251009345687635)
--(axis cs:8,-0.0225173920345292)
--(axis cs:9,0.00283724926604078)
--(axis cs:10,0.00277619204260372)
--(axis cs:11,-0.00162314053748265)
--(axis cs:12,0.0027109376250824)
--(axis cs:13,0.00321528410223498)
--(axis cs:14,0.00250117338937592)
--(axis cs:15,0.00313827434197937)
--(axis cs:16,0.00237588590986926)
--(axis cs:17,0.0014545747530442)
--(axis cs:18,0.00123707824107966)
--(axis cs:19,0.00146058793375234)
--(axis cs:20,0.00315317615931404)
--(axis cs:21,0.00258076405624184)
--(axis cs:22,0.0025365201171795)
--(axis cs:23,0.00221978885779127)
--(axis cs:24,0.00254739908824355)
--(axis cs:25,0.0027058950256329)
--(axis cs:26,0.00229950416600936)
--(axis cs:27,0.0012980545452333)
--(axis cs:28,0.00187032806131462)
--(axis cs:29,0.00157641369976768)
--(axis cs:30,0.00206739221543141)
--(axis cs:31,0.00191378933562661)
--(axis cs:32,0.00183675882135155)
--(axis cs:33,0.00238501964263276)
--(axis cs:34,0.0017106696436391)
--(axis cs:35,0.00190475028701485)
--(axis cs:36,0.00109050703935598)
--(axis cs:37,0.00185468176362666)
--(axis cs:38,0.00162220009140316)
--(axis cs:39,0.00171609222538846)
--(axis cs:40,0.00212434419735655)
--(axis cs:41,0.00156372002215229)
--(axis cs:42,0.00203333215341498)
--(axis cs:43,0.00125280556520889)
--(axis cs:44,0.0022791594479191)
--(axis cs:45,0.0020406870307783)
--(axis cs:46,0.00207789521090765)
--(axis cs:47,0.00147544237195981)
--(axis cs:48,0.00178808889238572)
--(axis cs:49,0.00193350050451671)
--(axis cs:50,0.0014413032801368)
--(axis cs:51,0.00206807409374529)
--(axis cs:52,0.0015430323053328)
--(axis cs:53,0.00160093232547033)
--(axis cs:54,0.00182471434133381)
--(axis cs:55,0.00219956153538063)
--(axis cs:56,0.00209076945608913)
--(axis cs:57,0.00151870121004731)
--(axis cs:58,0.00154388518186174)
--(axis cs:59,0.00169379615768761)
--(axis cs:60,0.00129742330581223)
--(axis cs:61,0.00165248041698372)
--(axis cs:62,0.00147005003917839)
--(axis cs:63,0.00183413853303019)
--(axis cs:64,0.00189414343795566)
--(axis cs:65,0.00159007843326585)
--(axis cs:66,0.00144726947948111)
--(axis cs:67,0.00131249183401409)
--(axis cs:68,0.00159283761419502)
--(axis cs:69,0.0022472061013264)
--(axis cs:70,0.00168834669012004)
--(axis cs:71,0.00157197714740717)
--(axis cs:71,0.00595961183086538)
--(axis cs:71,0.00595961183086538)
--(axis cs:70,0.00593749850643227)
--(axis cs:69,0.00586020329698623)
--(axis cs:68,0.00595553664773381)
--(axis cs:67,0.00601380455477199)
--(axis cs:66,0.00598489273671019)
--(axis cs:65,0.00595607299148292)
--(axis cs:64,0.00590261281769087)
--(axis cs:63,0.00591216059105027)
--(axis cs:62,0.00598016269852487)
--(axis cs:61,0.00594414897009218)
--(axis cs:60,0.006017129459198)
--(axis cs:59,0.00593650158840211)
--(axis cs:58,0.00596517199327714)
--(axis cs:57,0.00597022503544609)
--(axis cs:56,0.00587568987973536)
--(axis cs:55,0.00586426477992355)
--(axis cs:54,0.00591370972889452)
--(axis cs:53,0.0059539679163375)
--(axis cs:52,0.00596534207234531)
--(axis cs:51,0.00587841398073084)
--(axis cs:50,0.00598613938336218)
--(axis cs:49,0.00589666254556178)
--(axis cs:48,0.00591985187298961)
--(axis cs:47,0.00597905011231663)
--(axis cs:46,0.00602325868432242)
--(axis cs:45,0.00592809796298298)
--(axis cs:44,0.00677538840615079)
--(axis cs:43,0.00559225870936547)
--(axis cs:42,0.00657347132785782)
--(axis cs:41,0.00578063776461987)
--(axis cs:40,0.00564180392343047)
--(axis cs:39,0.00719859617566057)
--(axis cs:38,0.00547256790619142)
--(axis cs:37,0.00567799771786979)
--(axis cs:36,0.00538915687896708)
--(axis cs:35,0.00559543875966692)
--(axis cs:34,0.00551384480855031)
--(axis cs:33,0.00585461659205801)
--(axis cs:32,0.00639529987929277)
--(axis cs:31,0.00552468098151892)
--(axis cs:30,0.006914870367508)
--(axis cs:29,0.00454650694557047)
--(axis cs:28,0.00349547285521072)
--(axis cs:27,0.00725800808732159)
--(axis cs:26,0.006135432056766)
--(axis cs:25,0.00451229151930865)
--(axis cs:24,0.00451293773935495)
--(axis cs:23,0.00575769200866176)
--(axis cs:22,0.00691965457930375)
--(axis cs:21,0.00479044927247081)
--(axis cs:20,0.00907925169416978)
--(axis cs:19,0.00469209563354128)
--(axis cs:18,0.0120268164520117)
--(axis cs:17,0.00512538683360108)
--(axis cs:16,0.0103741248905355)
--(axis cs:15,0.00807839366193447)
--(axis cs:14,0.00738651777069026)
--(axis cs:13,0.00800785974040442)
--(axis cs:12,0.00468529212996968)
--(axis cs:11,0.0234595809733227)
--(axis cs:10,0.0159260997072221)
--(axis cs:9,0.0063247853996074)
--(axis cs:8,0.0837600935869111)
--(axis cs:7,0.170138715863723)
--(axis cs:6,0.241073668978299)
--(axis cs:5,0.259774154674626)
--(axis cs:4,0.276600301650933)
--(axis cs:3,0.319639705950313)
--(axis cs:2,0.334547721132102)
--(axis cs:1,0.321308770928158)
--cycle;

\addplot [line0]
table {%
1 0.350610145403875
2 0.305560987190302
3 0.281307003203193
4 0.265440680539289
5 0.248277345099036
6 0.237758023654583
7 0.220702615275691
8 0.212674689662474
9 0.198216257670978
10 0.189290008169824
11 0.181899107318931
12 0.175711621300347
13 0.171524863346347
14 0.173482893806119
15 0.169341057641731
16 0.153214590690288
17 0.151416090321202
18 0.148761965014052
19 0.13975223930942
20 0.143778643807409
21 0.139864578474709
22 0.137454085570358
23 0.139050479050983
24 0.131333504586538
25 0.142613817107524
26 0.130867484261525
27 0.137758789793189
28 0.118898758907899
29 0.117456751590309
30 0.120485408342889
31 0.116570838777476
32 0.126368964212836
33 0.119585928878082
34 0.120595912427119
35 0.123346492975596
36 0.119707182115536
37 0.115529702758387
38 0.109076602794899
39 0.106021037125556
40 0.111931405314488
41 0.107440281587924
42 0.0961861375276863
43 0.0953944598901882
44 0.105236453739239
45 0.1003658626711
46 0.0959542410453017
47 0.108433125291358
48 0.0985550888666915
49 0.100243318529968
50 0.0946325648419907
51 0.10087731695073
52 0.0897758332471759
53 0.0872205439921136
54 0.107201995460922
55 0.0889106192013536
56 0.0888253278340227
57 0.0820639634160992
58 0.0994632529066621
59 0.0859877718952233
60 0.0802156764292505
61 0.0971324117932063
62 0.0776601569012243
63 0.0879670903061865
64 0.0874786055077715
65 0.0748788135250748
66 0.077985494198035
67 0.0754900729346553
68 0.0800811927610963
69 0.0806421361822872
70 0.0791818944468803
71 0.071644239745697
72 0.0690238332530131
73 0.0706650634688
74 0.0716375061448216
75 0.0743432840129793
76 0.0793549730520611
77 0.0888393943854515
78 0.084983338534825
79 0.0843360526069954
80 0.076928978838894
81 0.0800641512841575
82 0.0801066340117917
83 0.0788085996662222
84 0.0762588870308448
85 0.07391843834541
86 0.0771646301241506
87 0.0771100803271343
88 0.0814632557574631
89 0.0790557694625656
90 0.0772257517006978
91 0.0790278269614082
92 0.0895837506023194
93 0.0845721515893822
94 0.0790203920689488
95 0.0813363726392993
96 0.0756313323133509
97 0.0844772223137992
98 0.0830037821259869
99 0.0787817620307855
100 0.0862458106176388
101 0.0832621196626379
102 0.0825296670158349
103 0.0884166725526833
104 0.0830529453392829
105 0.0826276804517892
106 0.0830826175667233
107 0.0870975540516034
} node[linelabel,pos=1,right] {LSTM};
\addlegendentry{LSTM}
\addplot [line1]
table {%
1 0.349461444152754
2 0.157999535421878
3 0.117256514659113
4 0.113337607870439
5 0.112795989780546
6 0.106062291678294
7 0.100688720407314
8 0.0994329411165548
9 0.0970941376922302
10 0.0889603751058731
11 0.085927564066259
12 0.0837682470797314
13 0.080863843657904
14 0.0752136136802968
15 0.0768317224124971
16 0.0776179374420128
17 0.076796858121909
18 0.0723452683679993
19 0.0763387352755574
20 0.0719243192727246
21 0.06794296864165
22 0.0668621740848245
23 0.0626499330472696
24 0.0657631356675563
25 0.0588388150255623
26 0.0654055236565714
27 0.0592699027341686
28 0.0607901954672643
29 0.0527968728904299
30 0.0643650374308682
31 0.0662396625791612
32 0.0550378454029501
33 0.0522417476317143
34 0.0523529248388043
35 0.051701917251141
36 0.053668311039576
37 0.0501600942218999
38 0.0515384722567001
39 0.053309609037777
40 0.0459482160013812
41 0.048171357539954
42 0.0592426520888523
43 0.0537758540166227
44 0.0487533335424226
45 0.0465489595044204
46 0.0472063192439669
47 0.0431519337613699
48 0.0606550460146192
49 0.0441579113187961
50 0.0461940425788415
51 0.0444311193797705
52 0.0456960587694972
53 0.0403148477602216
54 0.0510222561942206
55 0.0404221774646818
56 0.0411475228003062
57 0.0359895075972224
58 0.0517532251556805
59 0.0380297524860855
60 0.0417813444508532
61 0.0366675704398519
62 0.0387310326763591
63 0.0402863224551577
64 0.0445658521949065
65 0.0377170402325868
66 0.0426031563612688
67 0.0404592267937025
68 0.0381949675805258
69 0.0457416752087146
70 0.0412916674290244
71 0.0356674331186913
72 0.039303881141785
73 0.0383865187768458
74 0.0448743938962522
75 0.0360360561828564
76 0.0395991034033827
77 0.0405198345524514
78 0.0355808922345133
79 0.0367859795363331
80 0.0370664454324107
81 0.0399612426534855
82 0.035437932218652
83 0.037073296560143
84 0.0366263333784453
85 0.0386907271724157
86 0.037651512260814
87 0.0365279455782213
88 0.0384978469214335
89 0.0356621658533167
90 0.0365029054450596
91 0.0353057393449726
92 0.0379600182874677
93 0.0341270398564614
94 0.0411480500631222
95 0.0370150514215657
96 0.0407369244686758
97 0.0371064717433424
98 0.034424018347876
99 0.0375921240841935
100 0.036655458028923
101 0.0425284239285264
102 0.0349500778327863
103 0.0358477286432924
104 0.0350389117996107
} node[linelabel,pos=1,right] {Gref, JM};
\addlegendentry{Gref}
\addplot [line2]
table {%
1 0.322758843002256
2 0.22369579484571
3 0.205653144783237
4 0.1544616497541
5 0.139670530418901
6 0.134789266703938
7 0.125016492020601
8 0.0832450369609593
9 0.084438480005194
10 0.0768857934315793
11 0.0826056978321593
12 0.0714598194842356
13 0.0730336895825348
14 0.0748253167137736
15 0.0730597300310864
16 0.067685537630189
17 0.0665084656451029
18 0.0659181253565125
19 0.0603616266648951
20 0.070083252568577
21 0.0656772540964093
22 0.0549768491568773
23 0.0645060785514411
24 0.055726631152027
25 0.0526135906840392
26 0.060488762030266
27 0.0528217273009754
28 0.0515102024333138
29 0.0603192194972743
30 0.0474094382777194
31 0.0515064918059213
32 0.0474584435216766
33 0.049814189969644
34 0.0549747321511877
35 0.0447956180067961
36 0.049846376398434
37 0.046293453791521
38 0.0422969520630062
39 0.0479744756032017
40 0.0459885907802073
41 0.0433130740277213
42 0.0446702477431498
43 0.0482414418213362
44 0.0619563124549854
45 0.0387965105258747
46 0.0379585795518613
47 0.0438551200046806
48 0.0420136590954733
49 0.0419044277439203
50 0.0432972237898595
51 0.0357527883681813
52 0.0386101142423327
53 0.040355144974991
54 0.0388977555730459
55 0.0382150441295529
56 0.0426661089770353
57 0.03988159359666
58 0.0350100517792086
59 0.0361513702120708
60 0.0408493725967286
61 0.0399242112924461
62 0.0349462679996178
63 0.0345330138582089
64 0.035889151188127
65 0.0345803506024345
66 0.0360998676143002
67 0.0345131301250783
68 0.0334946892916198
69 0.039650753073154
70 0.0387566749428465
71 0.0380250741846225
72 0.0355822937250176
73 0.0367989367363085
74 0.035723500906508
75 0.0384511813807287
76 0.0368814098787984
77 0.0339673606473107
78 0.0339199193880349
79 0.0318569799451165
80 0.0345142793313355
81 0.039695197230681
82 0.0319614780013815
83 0.0332677402186193
84 0.0356211743953673
85 0.033227631700561
86 0.0340916126094174
87 0.0321775516403996
88 0.0365522690721218
89 0.0352841011107206
90 0.0311240412105324
91 0.0317201785731115
92 0.0327144134045483
93 0.0307117020351072
94 0.0361512504382315
95 0.0330348198558214
96 0.0319973743281674
97 0.0368918859815955
98 0.0341064969712714
99 0.0335671550159908
100 0.0328231763178727
101 0.0335181421524743
102 0.0347955419364221
103 0.032614924109584
104 0.033848406342842
};
\addlegendentry{JM}
\addplot [line3]
table {%
1 0.313859841505135
2 0.22668504311376
3 0.214349344134473
4 0.158131179419622
5 0.143445725191347
6 0.132137535823011
7 0.0725188906474798
8 0.0306213507761909
9 0.00458101733282409
10 0.00935114587491289
11 0.01091822021792
12 0.00369811487752604
13 0.0056115719213197
14 0.00494384558003309
15 0.00560833400195692
16 0.00637500540020238
17 0.00328998079332264
18 0.00663194734654569
19 0.00307634178364681
20 0.00611621392674191
21 0.00368560666435632
22 0.00472808734824163
23 0.00398874043322651
24 0.00353016841379925
25 0.00360909327247078
26 0.00421746811138768
27 0.00427803131627744
28 0.00268290045826267
29 0.00306146032266907
30 0.00449113129146971
31 0.00371923515857276
32 0.00411602935032216
33 0.00411981811734539
34 0.00361225722609471
35 0.00375009452334089
36 0.00323983195916153
37 0.00376633974074823
38 0.00354738399879729
39 0.00445734420052452
40 0.00388307406039351
41 0.00367217889338608
42 0.0043034017406364
43 0.00342253213728718
44 0.00452727392703495
45 0.00398439249688064
46 0.00405057694761504
47 0.00372724624213822
48 0.00385397038268767
49 0.00391508152503924
50 0.00371372133174949
51 0.00397324403723807
52 0.00375418718883905
53 0.00377745012090391
54 0.00386921203511417
55 0.00403191315765209
56 0.00398322966791225
57 0.0037444631227467
58 0.00375452858756944
59 0.00381514887304486
60 0.00365727638250511
61 0.00379831469353795
62 0.00372510636885163
63 0.00387314956204023
64 0.00389837812782327
65 0.00377307571237439
66 0.00371608110809565
67 0.00366314819439304
68 0.00377418713096441
69 0.00405370469915631
70 0.00381292259827616
71 0.00376579448913628
} node[linelabel,pos=1,right] {Ours};
\addlegendentry{Ours}
\legend{} 
\end{axis}

\end{tikzpicture}

%% file: figures/train-dyck.tex
\begin{tikzpicture}

\begin{axis}[
title={Dyck},
ylabel={Difference in Cross Entropy},
ymin=-0.02,
ymax=0.12,
]
\path [draw=color0, fill=color0, bars]
(axis cs:1,0.156984268722147)
--(axis cs:1,0.14934480093293)
--(axis cs:2,0.0917429690876853)
--(axis cs:3,0.075837583535259)
--(axis cs:4,0.0725228714818672)
--(axis cs:5,0.0578701734850149)
--(axis cs:6,0.0558247973267697)
--(axis cs:7,0.0451883801618302)
--(axis cs:8,0.0512878968594385)
--(axis cs:9,0.0448572272989751)
--(axis cs:10,0.0435611471058352)
--(axis cs:11,0.0349587853559547)
--(axis cs:12,0.0412216580541777)
--(axis cs:13,0.034622362519469)
--(axis cs:14,0.0341209847770745)
--(axis cs:15,0.0340408151913031)
--(axis cs:16,0.0341139981402237)
--(axis cs:17,0.0305265832453923)
--(axis cs:18,0.0305692443028597)
--(axis cs:19,0.0373186347787619)
--(axis cs:20,0.0291659517785153)
--(axis cs:21,0.0243210167609624)
--(axis cs:22,0.0330179212782273)
--(axis cs:23,0.0300990839080487)
--(axis cs:24,0.0253941880640112)
--(axis cs:25,0.0292565896366938)
--(axis cs:26,0.0246550341025851)
--(axis cs:27,0.0244318074763676)
--(axis cs:28,0.0223209327584013)
--(axis cs:29,0.0211929494110192)
--(axis cs:30,0.0208670584881956)
--(axis cs:31,0.021165386140297)
--(axis cs:32,0.0178834490951454)
--(axis cs:33,0.0207344641950368)
--(axis cs:34,0.02150644566995)
--(axis cs:35,0.0205855595657315)
--(axis cs:36,0.0209519465566307)
--(axis cs:37,0.0195387286425058)
--(axis cs:38,0.0222527208989331)
--(axis cs:39,0.0215396987633069)
--(axis cs:40,0.0172566996488724)
--(axis cs:41,0.0190874227481677)
--(axis cs:42,0.0258161986318417)
--(axis cs:43,0.0206646493769231)
--(axis cs:44,0.0167747739288708)
--(axis cs:45,0.0172508809441643)
--(axis cs:46,0.0202438584678348)
--(axis cs:47,0.0172009887517829)
--(axis cs:48,0.0184109134542486)
--(axis cs:49,0.0188173169048452)
--(axis cs:50,0.0200870751356416)
--(axis cs:51,0.0169674579061429)
--(axis cs:52,0.0189330782674886)
--(axis cs:53,0.0144561334255674)
--(axis cs:54,0.0169091607613264)
--(axis cs:55,0.017600418156315)
--(axis cs:56,0.0150213428103614)
--(axis cs:57,0.0156756457263919)
--(axis cs:58,0.0196778432184933)
--(axis cs:59,0.0197353359087499)
--(axis cs:60,0.0157086820189779)
--(axis cs:61,0.0144587433817934)
--(axis cs:62,0.0156934955479162)
--(axis cs:63,0.01324451287999)
--(axis cs:64,0.0134682281611133)
--(axis cs:65,0.0173264674812033)
--(axis cs:66,0.0175706957607636)
--(axis cs:67,0.0169540957481145)
--(axis cs:68,0.0145443843311354)
--(axis cs:69,0.0135727340212294)
--(axis cs:70,0.0132022840434427)
--(axis cs:71,0.0144712399929897)
--(axis cs:72,0.0142707561167278)
--(axis cs:73,0.0143607579631199)
--(axis cs:74,0.0135894567693498)
--(axis cs:75,0.017913466692221)
--(axis cs:76,0.0174076943449461)
--(axis cs:77,0.0204892623262186)
--(axis cs:78,0.0164791501903015)
--(axis cs:79,0.0154198295295093)
--(axis cs:80,0.014413555903227)
--(axis cs:81,0.0136478288012115)
--(axis cs:82,0.0173777909713154)
--(axis cs:83,0.0179967749966623)
--(axis cs:84,0.0139847825594378)
--(axis cs:85,0.0171412747408988)
--(axis cs:86,0.0175640313422425)
--(axis cs:87,0.0152815524440033)
--(axis cs:88,0.0157815665142679)
--(axis cs:89,0.013362855084808)
--(axis cs:90,0.0166777792389097)
--(axis cs:91,0.0142382356895709)
--(axis cs:92,0.0149595664979187)
--(axis cs:93,0.01869556894743)
--(axis cs:94,0.0145563061843703)
--(axis cs:95,0.0157365747536532)
--(axis cs:96,0.0223220781011648)
--(axis cs:97,0.0129817713301043)
--(axis cs:98,0.0133954621723251)
--(axis cs:99,0.0164288738639164)
--(axis cs:100,0.0173218178108594)
--(axis cs:101,0.0159535233547008)
--(axis cs:102,0.0168821708579157)
--(axis cs:103,0.0158589054630775)
--(axis cs:104,0.0172477083977919)
--(axis cs:105,0.016053737327497)
--(axis cs:106,0.0126474924518988)
--(axis cs:107,0.0155406810677557)
--(axis cs:108,0.0146504562449854)
--(axis cs:109,0.0133480349811598)
--(axis cs:110,0.0126518940643593)
--(axis cs:111,0.0139230391539057)
--(axis cs:112,0.0151187354505438)
--(axis cs:113,0.0146985274669201)
--(axis cs:114,0.0129107237253194)
--(axis cs:115,0.0125934146702128)
--(axis cs:116,0.0129043608657657)
--(axis cs:117,0.0137559583401765)
--(axis cs:118,0.0129420953423266)
--(axis cs:119,0.0149429600225638)
--(axis cs:120,0.0140451611400072)
--(axis cs:121,0.014087207948172)
--(axis cs:122,0.0138641942865537)
--(axis cs:123,0.0136060731102563)
--(axis cs:124,0.0121145391050163)
--(axis cs:125,0.0134232315865815)
--(axis cs:126,0.0133364190048866)
--(axis cs:127,0.015446094495208)
--(axis cs:128,0.0147258081683857)
--(axis cs:129,0.0125623272462939)
--(axis cs:130,0.0145525559182547)
--(axis cs:131,0.015799138369084)
--(axis cs:132,0.0135036582050288)
--(axis cs:133,0.0142105869196771)
--(axis cs:134,0.0125110381732562)
--(axis cs:135,0.0133102606897284)
--(axis cs:135,0.0337567566810437)
--(axis cs:135,0.0337567566810437)
--(axis cs:134,0.0339725207083645)
--(axis cs:133,0.0335213413273697)
--(axis cs:132,0.0337054526873056)
--(axis cs:131,0.0331317887443701)
--(axis cs:130,0.033434407543226)
--(axis cs:129,0.0339585052308488)
--(axis cs:128,0.0333909488050041)
--(axis cs:127,0.0332149613565564)
--(axis cs:126,0.0337497955134898)
--(axis cs:125,0.0337267421486408)
--(axis cs:124,0.0340815879938257)
--(axis cs:123,0.033678439547957)
--(axis cs:122,0.0336108546743954)
--(axis cs:121,0.0335530628088321)
--(axis cs:120,0.0335639147428592)
--(axis cs:119,0.0333370688864562)
--(axis cs:118,0.0338554303252595)
--(axis cs:117,0.033639106052708)
--(axis cs:116,0.0338656150943264)
--(axis cs:115,0.0339500208567536)
--(axis cs:114,0.0338638968159418)
--(axis cs:113,0.0333977647533013)
--(axis cs:112,0.033293960622901)
--(axis cs:111,0.0335955501469426)
--(axis cs:110,0.0339340828012185)
--(axis cs:109,0.033746706483267)
--(axis cs:108,0.0334098000255702)
--(axis cs:107,0.03319246383269)
--(axis cs:106,0.0339352814109973)
--(axis cs:105,0.0330732314432571)
--(axis cs:104,0.0328177151578539)
--(axis cs:103,0.0331179310799901)
--(axis cs:102,0.0328922098403889)
--(axis cs:101,0.0330961311877294)
--(axis cs:100,0.03280307142559)
--(axis cs:99,0.0329893404570979)
--(axis cs:98,0.0337341081754749)
--(axis cs:97,0.0338447354991483)
--(axis cs:96,0.0327766978408109)
--(axis cs:95,0.0331463657926594)
--(axis cs:94,0.033433462538845)
--(axis cs:93,0.0325666213730897)
--(axis cs:92,0.0333329764777005)
--(axis cs:91,0.0335142579339385)
--(axis cs:90,0.0329353940575537)
--(axis cs:89,0.0337427673322202)
--(axis cs:88,0.0331358756314386)
--(axis cs:87,0.0332544538606032)
--(axis cs:86,0.0327563778393555)
--(axis cs:85,0.0328390248137897)
--(axis cs:84,0.0335795341399723)
--(axis cs:83,0.0326778178494319)
--(axis cs:82,0.032792119854306)
--(axis cs:81,0.0336674576422824)
--(axis cs:80,0.0334695634922097)
--(axis cs:79,0.0332212353457059)
--(axis cs:78,0.0329783287037758)
--(axis cs:77,0.0324191622780583)
--(axis cs:76,0.0327863080114254)
--(axis cs:75,0.0326924213124505)
--(axis cs:74,0.0323995607775612)
--(axis cs:73,0.0323611834509288)
--(axis cs:72,0.0324757492430528)
--(axis cs:71,0.0323582664416127)
--(axis cs:70,0.0326337071624257)
--(axis cs:69,0.0324301468011741)
--(axis cs:68,0.0323767750474897)
--(axis cs:67,0.0328501141102168)
--(axis cs:66,0.0324232160642791)
--(axis cs:65,0.0323311274918177)
--(axis cs:64,0.0325292277566728)
--(axis cs:63,0.0323921925681469)
--(axis cs:62,0.0326617023939587)
--(axis cs:61,0.0324638861270193)
--(axis cs:60,0.0323412907688534)
--(axis cs:59,0.0359061321050012)
--(axis cs:58,0.032399438878528)
--(axis cs:57,0.0318695843406478)
--(axis cs:56,0.0316442656337976)
--(axis cs:55,0.0325045951970119)
--(axis cs:54,0.0358362104036084)
--(axis cs:53,0.0316651587583255)
--(axis cs:52,0.032705240701737)
--(axis cs:51,0.0314364004754326)
--(axis cs:50,0.0313525951106449)
--(axis cs:49,0.032777557804303)
--(axis cs:48,0.0319470773078859)
--(axis cs:47,0.0431896087368105)
--(axis cs:46,0.0325327480090504)
--(axis cs:45,0.0326792826883879)
--(axis cs:44,0.0327091494361188)
--(axis cs:43,0.0299824785896643)
--(axis cs:42,0.0349319987397075)
--(axis cs:41,0.0291408922685775)
--(axis cs:40,0.0348290772923408)
--(axis cs:39,0.0353970450459133)
--(axis cs:38,0.0345531981911091)
--(axis cs:37,0.0360245369196884)
--(axis cs:36,0.0332408762816602)
--(axis cs:35,0.0387701507745892)
--(axis cs:34,0.0377660472558123)
--(axis cs:33,0.0409631570792291)
--(axis cs:32,0.038427406469779)
--(axis cs:31,0.0333954166128957)
--(axis cs:30,0.0329091317697656)
--(axis cs:29,0.0347560092727475)
--(axis cs:28,0.0408460172917265)
--(axis cs:27,0.0406348189252076)
--(axis cs:26,0.0383685638161519)
--(axis cs:25,0.0378974238915205)
--(axis cs:24,0.0403112025189306)
--(axis cs:23,0.0486781787729245)
--(axis cs:22,0.0477649292982374)
--(axis cs:21,0.0413561149508492)
--(axis cs:20,0.0373958551237098)
--(axis cs:19,0.0470968712389332)
--(axis cs:18,0.0480571117629895)
--(axis cs:17,0.0512851690695325)
--(axis cs:16,0.0503434179505019)
--(axis cs:15,0.0493047544560833)
--(axis cs:14,0.0543324244691205)
--(axis cs:13,0.0475429467304361)
--(axis cs:12,0.0547031780725929)
--(axis cs:11,0.0493443426113589)
--(axis cs:10,0.068690190016487)
--(axis cs:9,0.0598773953694631)
--(axis cs:8,0.0633534142504679)
--(axis cs:7,0.0650280466351779)
--(axis cs:6,0.0607469152901408)
--(axis cs:5,0.0734658753173224)
--(axis cs:4,0.0822757735365871)
--(axis cs:3,0.102603253602815)
--(axis cs:2,0.137718742558142)
--(axis cs:1,0.156984268722147)
--cycle;

\path [draw=color1, fill=color1, bars]
(axis cs:1,0.175300923086401)
--(axis cs:1,0.0370223071181166)
--(axis cs:2,0.00403171318390066)
--(axis cs:3,-0.00921562755920926)
--(axis cs:4,-0.00410185347163009)
--(axis cs:5,0.000719795578231599)
--(axis cs:6,0.000937164198047191)
--(axis cs:7,8.90752764219857e-05)
--(axis cs:8,0.000355538535542042)
--(axis cs:9,0.000377328341327158)
--(axis cs:10,0.000589293779572218)
--(axis cs:11,-0.000721821191147351)
--(axis cs:12,-0.000349562642622616)
--(axis cs:13,-0.00116266906536631)
--(axis cs:14,-0.000102937292146267)
--(axis cs:15,-0.000528148533216924)
--(axis cs:16,0.000986495978347223)
--(axis cs:17,2.57141624500897e-05)
--(axis cs:18,-0.000137659746387644)
--(axis cs:19,-0.000658434299360286)
--(axis cs:20,-0.000588079986975142)
--(axis cs:21,-0.0014905754672511)
--(axis cs:22,-0.000932423794687341)
--(axis cs:23,-0.000845466158058301)
--(axis cs:24,-0.000528197164496787)
--(axis cs:25,-0.00204422325199754)
--(axis cs:26,-0.00172374203010517)
--(axis cs:27,-0.000376585935068493)
--(axis cs:28,-0.00146942706657891)
--(axis cs:29,-0.00051530344684699)
--(axis cs:30,-0.00112799686494255)
--(axis cs:31,-0.00281683956266757)
--(axis cs:32,-0.00350083634001391)
--(axis cs:33,-0.00210313780040624)
--(axis cs:34,-0.00171327854390584)
--(axis cs:35,-0.00280745547764305)
--(axis cs:36,-0.00169582999122378)
--(axis cs:37,-0.00204165194823207)
--(axis cs:38,-0.00226168485567005)
--(axis cs:39,-0.00234129537733196)
--(axis cs:40,-0.00237171494063939)
--(axis cs:41,-0.00245379724174849)
--(axis cs:42,-0.00201666878103677)
--(axis cs:42,0.0061433097078998)
--(axis cs:42,0.0061433097078998)
--(axis cs:41,0.00608231210628311)
--(axis cs:40,0.00608955108124162)
--(axis cs:39,0.00609267222460683)
--(axis cs:38,0.00610203983392251)
--(axis cs:37,0.00613813813737005)
--(axis cs:36,0.00623452908528106)
--(axis cs:35,0.00556416519171032)
--(axis cs:34,0.00608443417497724)
--(axis cs:33,0.00637106827041832)
--(axis cs:32,0.00527909734211386)
--(axis cs:31,0.00565125166041426)
--(axis cs:30,0.00733864578230213)
--(axis cs:29,0.00673883867444941)
--(axis cs:28,0.0066488244740049)
--(axis cs:27,0.00152621528195294)
--(axis cs:26,0.00163830478527107)
--(axis cs:25,0.000889086147270043)
--(axis cs:24,0.00219517725293105)
--(axis cs:23,0.0022632123537566)
--(axis cs:22,0.000224350595395227)
--(axis cs:21,0.00140670203749945)
--(axis cs:20,0.000986444974605382)
--(axis cs:19,0.00171026192427214)
--(axis cs:18,0.00293369153977541)
--(axis cs:17,0.00203306778289921)
--(axis cs:16,0.00496867162195053)
--(axis cs:15,0.00129567779814624)
--(axis cs:14,0.00169739205248832)
--(axis cs:13,0.0022714761424371)
--(axis cs:12,0.00296879712033552)
--(axis cs:11,0.0069502659493667)
--(axis cs:10,0.0052143525458467)
--(axis cs:9,0.00268298501306876)
--(axis cs:8,0.00849015834449567)
--(axis cs:7,0.0066046584202451)
--(axis cs:6,0.00927196545780464)
--(axis cs:5,0.0151505710099274)
--(axis cs:4,0.0396375001485449)
--(axis cs:3,0.0775856980192727)
--(axis cs:2,0.133104354267671)
--(axis cs:1,0.175300923086401)
--cycle;

\path [draw=color2, fill=color2, bars]
(axis cs:1,0.144768067027514)
--(axis cs:1,0.00807431888975634)
--(axis cs:2,0.0023519588680965)
--(axis cs:3,-0.00200588474439818)
--(axis cs:4,0.000523899843195775)
--(axis cs:5,-0.000416514294406105)
--(axis cs:6,-7.73900995656215e-05)
--(axis cs:7,-0.00181173434562975)
--(axis cs:8,0.00123634116767791)
--(axis cs:9,-0.00020892784207574)
--(axis cs:10,-0.00413330480194978)
--(axis cs:11,-0.00266398240273989)
--(axis cs:12,-0.00208029472014225)
--(axis cs:13,-0.00125844402458884)
--(axis cs:14,-0.0010446227966715)
--(axis cs:15,-0.00158796832315308)
--(axis cs:16,-0.00132146670861425)
--(axis cs:17,-0.00128676837107894)
--(axis cs:18,-0.00165848710599884)
--(axis cs:19,-0.00180126459546435)
--(axis cs:20,-0.00198974115415043)
--(axis cs:21,-0.0019682448704197)
--(axis cs:22,-0.0025976628557063)
--(axis cs:23,-0.00121653532924437)
--(axis cs:24,-0.00193369653721903)
--(axis cs:25,-0.00126943484415863)
--(axis cs:26,-0.00191639370963409)
--(axis cs:27,-0.00245958211336639)
--(axis cs:28,-0.00269439628111098)
--(axis cs:29,-0.000722211074157208)
--(axis cs:30,-0.0019186407291235)
--(axis cs:31,-0.00333557603327014)
--(axis cs:32,-0.00340956461942347)
--(axis cs:33,-0.00278918701287556)
--(axis cs:34,-0.00291308356953415)
--(axis cs:35,-0.00265452484747385)
--(axis cs:36,-0.00111804882341)
--(axis cs:37,-0.00251020761916328)
--(axis cs:38,-0.00164965611669901)
--(axis cs:39,-0.0025225607469596)
--(axis cs:40,-0.00337905972494013)
--(axis cs:41,-0.00192748501812881)
--(axis cs:42,-0.00262318755560545)
--(axis cs:43,-0.00261996180176103)
--(axis cs:44,-0.00128863563105201)
--(axis cs:45,-0.00183350729605034)
--(axis cs:46,-0.00208735361121708)
--(axis cs:47,-0.00208917338982386)
--(axis cs:48,-0.00196426310639748)
--(axis cs:49,-0.0019812152361034)
--(axis cs:50,-0.00214063927827628)
--(axis cs:51,-0.00204665520317532)
--(axis cs:52,-0.00241171552825369)
--(axis cs:53,-0.00200554479991577)
--(axis cs:54,-0.00195807758173594)
--(axis cs:55,-0.0021488357946915)
--(axis cs:56,-0.00234837786490978)
--(axis cs:57,-0.00221855738702779)
--(axis cs:58,-0.00211860526044815)
--(axis cs:59,-0.00215346984186695)
--(axis cs:60,-0.00199108434418247)
--(axis cs:61,-0.00289712474221101)
--(axis cs:62,-0.00226861140953152)
--(axis cs:63,-0.00267281142066674)
--(axis cs:63,0.0021230917937485)
--(axis cs:63,0.0021230917937485)
--(axis cs:62,-0.000912303699920346)
--(axis cs:61,0.00282152867388533)
--(axis cs:60,-0.000762632958187303)
--(axis cs:59,-0.000895707519710736)
--(axis cs:58,-0.000884038502013403)
--(axis cs:57,-0.000908237257411005)
--(axis cs:56,-0.000912182971523597)
--(axis cs:55,-0.000894397255839542)
--(axis cs:54,-0.000531866019355201)
--(axis cs:53,-0.000790314035162501)
--(axis cs:52,-0.000907995668095215)
--(axis cs:51,-0.000840219151844074)
--(axis cs:50,-0.00089191150871157)
--(axis cs:49,-0.000737578885303938)
--(axis cs:48,-0.000665965184805613)
--(axis cs:47,-0.000791500764470121)
--(axis cs:46,-0.000902145172459016)
--(axis cs:45,0.00235877461175354)
--(axis cs:44,0.00219676185131225)
--(axis cs:43,-0.00042035171796197)
--(axis cs:42,0.00307436175029619)
--(axis cs:41,-0.000198652655824212)
--(axis cs:40,0.00425928723724179)
--(axis cs:39,-1.13705922713616e-05)
--(axis cs:38,0.000933793364191866)
--(axis cs:37,-0.00060715836657048)
--(axis cs:36,-0.000771328668209056)
--(axis cs:35,-2.9580170495976e-05)
--(axis cs:34,0.00727263291388745)
--(axis cs:33,0.003769398893427)
--(axis cs:32,0.00669632254630055)
--(axis cs:31,0.00273717019936445)
--(axis cs:30,-2.88683800126876e-05)
--(axis cs:29,0.00293972004461024)
--(axis cs:28,0.00314956000115362)
--(axis cs:27,0.00262341216342987)
--(axis cs:26,0.000930024000227206)
--(axis cs:25,0.00123317522137223)
--(axis cs:24,0.00148343855140146)
--(axis cs:23,0.00101782365424334)
--(axis cs:22,0.00357309493196724)
--(axis cs:21,0.00137296670462469)
--(axis cs:20,0.00182631485642824)
--(axis cs:19,0.0046852548226015)
--(axis cs:18,-0.000238603162671289)
--(axis cs:17,0.00521497494877191)
--(axis cs:16,0.0068853567535003)
--(axis cs:15,0.00257761053134916)
--(axis cs:14,0.002290718305779)
--(axis cs:13,0.00248153177366255)
--(axis cs:12,0.000325916200272861)
--(axis cs:11,0.00340213966539354)
--(axis cs:10,0.0106265118024022)
--(axis cs:9,0.0067779936750446)
--(axis cs:8,0.00508352479409301)
--(axis cs:7,0.0107897395815477)
--(axis cs:6,0.00224150565907146)
--(axis cs:5,0.00479148574208739)
--(axis cs:4,0.00859879584487154)
--(axis cs:3,0.0261124810834366)
--(axis cs:2,0.0210454618272359)
--(axis cs:1,0.144768067027514)
--cycle;

\path [draw=color3, fill=color3, bars]
(axis cs:1,0.154213247677685)
--(axis cs:1,0.0736796455176367)
--(axis cs:2,0.0115016484977029)
--(axis cs:3,0.00672258511170405)
--(axis cs:4,-0.000388178155992294)
--(axis cs:5,0.000746015243105038)
--(axis cs:6,6.06024356565602e-05)
--(axis cs:7,-9.15677989137235e-05)
--(axis cs:8,0.000261812247747769)
--(axis cs:9,-0.000475157962966409)
--(axis cs:10,0.00120131062465536)
--(axis cs:11,-0.000873955043744651)
--(axis cs:12,-0.000565430100888418)
--(axis cs:13,-0.00158409810818527)
--(axis cs:14,-0.00200984909255187)
--(axis cs:15,-0.00272317867777944)
--(axis cs:16,-0.00223144840767886)
--(axis cs:17,-0.00168776381613219)
--(axis cs:18,-0.00274005871223259)
--(axis cs:19,-0.00175670161188015)
--(axis cs:20,-0.00223723682875043)
--(axis cs:21,-0.0025633620688221)
--(axis cs:22,-0.00220709357549013)
--(axis cs:23,-0.00216753568326472)
--(axis cs:24,-0.00295639919678787)
--(axis cs:25,-0.00133259507102877)
--(axis cs:26,-0.00248530613373476)
--(axis cs:27,-0.00206004275842613)
--(axis cs:28,-0.00209617200960686)
--(axis cs:29,-0.00235400997518022)
--(axis cs:30,-0.00517579095007693)
--(axis cs:31,-0.00276218437154495)
--(axis cs:32,-0.00217915559511581)
--(axis cs:33,-0.00283132349386607)
--(axis cs:34,-0.00104330006107204)
--(axis cs:35,-0.00329750897749045)
--(axis cs:36,-0.00044245208994979)
--(axis cs:37,-0.00265821435464081)
--(axis cs:38,-0.00213445183260153)
--(axis cs:39,-0.0025643477612391)
--(axis cs:40,-0.00235842568677105)
--(axis cs:41,-0.00358211561413124)
--(axis cs:42,-0.0032246505302541)
--(axis cs:43,-0.00336608699105569)
--(axis cs:44,-0.00325890276913554)
--(axis cs:45,-0.00347388914502733)
--(axis cs:46,-0.00324756954216559)
--(axis cs:47,-0.00323490734038213)
--(axis cs:48,-0.00326006034599732)
--(axis cs:48,0.00244859645585348)
--(axis cs:48,0.00244859645585348)
--(axis cs:47,0.00248623007197605)
--(axis cs:46,0.00246685897919246)
--(axis cs:45,0.00222444236678278)
--(axis cs:44,0.00245025529023242)
--(axis cs:43,0.00232028784520246)
--(axis cs:42,0.00250867240927067)
--(axis cs:41,0.00217610443602234)
--(axis cs:40,0.00259808891772278)
--(axis cs:39,0.00303474435870203)
--(axis cs:38,0.00265066695554451)
--(axis cs:37,0.0025661896474147)
--(axis cs:36,0.00443091741736964)
--(axis cs:35,0.00214376234401482)
--(axis cs:34,0.00927442109905337)
--(axis cs:33,0.00252607865646909)
--(axis cs:32,0.00199173964842513)
--(axis cs:31,0.000435386085358187)
--(axis cs:30,0.0103871882625804)
--(axis cs:29,0.00404457416641303)
--(axis cs:28,-0.000312541027670287)
--(axis cs:27,-0.000570323077218535)
--(axis cs:26,0.00144325297854892)
--(axis cs:25,0.00216769838578353)
--(axis cs:24,0.00620132841362093)
--(axis cs:23,0.00201323980788783)
--(axis cs:22,3.22807338790617e-05)
--(axis cs:21,0.00162565721292879)
--(axis cs:20,9.7774121541986e-05)
--(axis cs:19,0.000933733454773079)
--(axis cs:18,0.00249866499314599)
--(axis cs:17,0.00225660361725691)
--(axis cs:16,0.00282664022895993)
--(axis cs:15,0.00945251314970784)
--(axis cs:14,0.00463611295195422)
--(axis cs:13,0.00479821975189553)
--(axis cs:12,0.00100172899448968)
--(axis cs:11,0.00411989699862784)
--(axis cs:10,0.0124483601192432)
--(axis cs:9,0.00226819273716786)
--(axis cs:8,0.00248254438263842)
--(axis cs:7,0.0023208098485767)
--(axis cs:6,0.00595570309478192)
--(axis cs:5,0.00306683266414854)
--(axis cs:4,0.0111828999165741)
--(axis cs:3,0.014201819354702)
--(axis cs:2,0.0694238030359632)
--(axis cs:1,0.154213247677685)
--cycle;

\addplot [line0]
table {%
1 0.153164534827539
2 0.114730855822914
3 0.0892204185690368
4 0.0773993225092271
5 0.0656680244011686
6 0.0582858563084552
7 0.055108213398504
8 0.0573206555549532
9 0.0523673113342191
10 0.0561256685611611
11 0.0421515639836568
12 0.0479624180633853
13 0.0410826546249526
14 0.0442267046230975
15 0.0416727848236932
16 0.0422287080453628
17 0.0409058761574624
18 0.0393131780329246
19 0.0422077530088476
20 0.0332809034511126
21 0.0328385658559058
22 0.0403914252882324
23 0.0393886313404866
24 0.0328526952914709
25 0.0335770067641071
26 0.0315117989593685
27 0.0325333132007876
28 0.0315834750250639
29 0.0279744793418834
30 0.0268880951289806
31 0.0272804013765964
32 0.0281554277824622
33 0.030848810637133
34 0.0296362464628812
35 0.0296778551701604
36 0.0270964114191454
37 0.0277816327810971
38 0.0284029595450211
39 0.0284683719046101
40 0.0260428884706066
41 0.0241141575083726
42 0.0303740986857746
43 0.0253235639832937
44 0.0247419616824948
45 0.0249650818162761
46 0.0263883032384426
47 0.0301952987442967
48 0.0251789953810673
49 0.0257974373545741
50 0.0257198351231433
51 0.0242019291907877
52 0.0258191594846128
53 0.0230606460919465
54 0.0263726855824674
55 0.0250525066766634
56 0.0233328042220795
57 0.0237726150335199
58 0.0260386410485107
59 0.0278207340068756
60 0.0240249863939156
61 0.0234613147544064
62 0.0241775989709375
63 0.0228183527240685
64 0.022998727958893
65 0.0248287974865105
66 0.0249969559125213
67 0.0249021049291656
68 0.0234605796893126
69 0.0230014404112018
70 0.0229179956029342
71 0.0234147532173012
72 0.0233732526798903
73 0.0233609707070244
74 0.0229945087734555
75 0.0253029440023357
76 0.0250970011781858
77 0.0264542123021384
78 0.0247287394470386
79 0.0243205324376076
80 0.0239415596977183
81 0.0236576432217469
82 0.0250849554128107
83 0.0253372964230471
84 0.0237821583497051
85 0.0249901497773442
86 0.025160204590799
87 0.0242680031523032
88 0.0244587210728533
89 0.0235528112085141
90 0.0248065866482317
91 0.0238762468117547
92 0.0241462714878096
93 0.0256310951602599
94 0.0239948843616076
95 0.0244414702731563
96 0.0275493879709879
97 0.0234132534146263
98 0.0235647851739
99 0.0247091071605071
100 0.0250624446182247
101 0.0245248272712151
102 0.0248871903491523
103 0.0244884182715338
104 0.0250327117778229
105 0.0245634843853771
106 0.023291386931448
107 0.0243665724502229
108 0.0240301281352778
109 0.0235473707322134
110 0.0232929884327889
111 0.0237592946504241
112 0.0242063480367224
113 0.0240481461101107
114 0.0233873102706306
115 0.0232717177634832
116 0.023384987980046
117 0.0236975321964423
118 0.0233987628337931
119 0.02414001445451
120 0.0238045379414332
121 0.0238201353785021
122 0.0237375244804745
123 0.0236422563291067
124 0.023098063549421
125 0.0235749868676111
126 0.0235431072591882
127 0.0243305279258822
128 0.0240583784866949
129 0.0232604162385713
130 0.0239934817307403
131 0.0244654635567271
132 0.0236045554461672
133 0.0238659641235234
134 0.0232417794408103
135 0.023533508685386
} node[pos=1,right,linelabel] {LSTM};
\addlegendentry{LSTM}
\addplot [line1]
table {%
1 0.106161615102259
2 0.0685680337257861
3 0.0341850352300317
4 0.0177678233384574
5 0.0079351832940795
6 0.00510456482792592
7 0.00334686684833354
8 0.00442284844001886
9 0.00153015667719796
10 0.00290182316270946
11 0.00311422237910968
12 0.00130961723885645
13 0.000554403538535397
14 0.000797227380171028
15 0.000383764632464656
16 0.00297758380014888
17 0.00102939097267465
18 0.00139801589669388
19 0.000525913812455925
20 0.00019918249381512
21 -4.19367148758232e-05
22 -0.000354036599646057
23 0.00070887309784915
24 0.000833490044217133
25 -0.00057756855236375
26 -4.27186224170484e-05
27 0.000574814673442225
28 0.00258969870371299
29 0.00311176761380121
30 0.00310532445867979
31 0.00141720604887334
32 0.000889130501049973
33 0.00213396523500604
34 0.0021855778155357
35 0.00137835485703364
36 0.00226934954702864
37 0.00204824309456899
38 0.00192017748912623
39 0.00187568842363743
40 0.00185891807030112
41 0.00181425743226731
42 0.00206332046343152
};
\addlegendentry{Gref}
\addplot [line2]
table {%
1 0.0764211929586353
2 0.0116987103476662
3 0.0120532981695192
4 0.00456134784403366
5 0.00218748572384064
6 0.00108205777975292
7 0.00448900261795897
8 0.00315993298088546
9 0.00328453291648443
10 0.00324660350022623
11 0.000369078631326825
12 -0.000877189259934696
13 0.000611543874536857
14 0.000623047754553752
15 0.000494821104098042
16 0.00278194502244302
17 0.00196410328884649
18 -0.000948545134335066
19 0.00144199511356857
20 -8.17131488610956e-05
21 -0.000297639082897505
22 0.00048771603813047
23 -9.93558375005144e-05
24 -0.000225128992908785
25 -1.81298113931971e-05
26 -0.000493184854703443
27 8.19150250317424e-05
28 0.00022758186002132
29 0.00110875448522652
30 -0.000973754554568096
31 -0.000299202916952845
32 0.00164337896343854
33 0.00049010594027572
34 0.00217977467217665
35 -0.00134205250898491
36 -0.00094468874580953
37 -0.00155868299286688
38 -0.000357931376253573
39 -0.00126696566961548
40 0.000440113756150828
41 -0.00106306883697651
42 0.00022558709734537
43 -0.0015201567598615
44 0.000454063110130121
45 0.000262633657851596
46 -0.00149474939183805
47 -0.00144033707714699
48 -0.00131511414560155
49 -0.00135939706070367
50 -0.00151627539349393
51 -0.0014434371775097
52 -0.00165985559817445
53 -0.00139792941753913
54 -0.00124497180054557
55 -0.00152161652526552
56 -0.00163028041821669
57 -0.0015633973222194
58 -0.00150132188123078
59 -0.00152458868078884
60 -0.00137685865118489
61 -3.77980341628392e-05
62 -0.00159045755472593
63 -0.000274859813459116
} node[pos=1,linelabel,right] {Gref, JM, Ours};
\addlegendentry{JM}
\addplot [line3]
table {%
1 0.113946446597661
2 0.040462725766833
3 0.010462202233203
4 0.00539736088029088
5 0.00190642395362679
6 0.00300815276521924
7 0.00111462102483149
8 0.0013721783151931
9 0.000896517387100726
10 0.00682483537194927
11 0.0016229709774416
12 0.000218149446800631
13 0.00160706082185513
14 0.00131313192970117
15 0.0033646672359642
16 0.000297595910640536
17 0.00028441990056236
18 -0.000120696859543301
19 -0.000411484078553537
20 -0.00106973135360422
21 -0.000468852427946653
22 -0.00108740642080554
23 -7.71479376884487e-05
24 0.00162246460841653
25 0.00041755165737738
26 -0.00052102657759292
27 -0.00131518291782233
28 -0.00120435651863857
29 0.000845282095616406
30 0.00260569865625175
31 -0.00116339914309338
32 -9.37079733453361e-05
33 -0.000152622418698489
34 0.00411556051899067
35 -0.000576873316737814
36 0.00199423266370993
37 -4.60123536130563e-05
38 0.000258107561471488
39 0.000235198298731465
40 0.000119831615475863
41 -0.000703005589054451
42 -0.000357989060491715
43 -0.000522899572926616
44 -0.000404323739451562
45 -0.000624723389122273
46 -0.000390355281486565
47 -0.000374338634203042
48 -0.000405731945071919
};
\addlegendentry{Ours}
\legend{} 
\end{axis}

\end{tikzpicture}

%% file: figures/train-hardest-cfl.tex
\begin{tikzpicture}

\begin{axis}[
title={Hardest CFL},
xlabel={Epoch},
ylabel={Difference in Cross Entropy},
ymin=0,ymax=0.07,ytick distance=0.05,
]
\path [draw=color0, fill=color0, bars]
(axis cs:1,0.107221007651666)
--(axis cs:1,0.100791876819128)
--(axis cs:2,0.0809073640656782)
--(axis cs:3,0.0717718955510831)
--(axis cs:4,0.0655953375813264)
--(axis cs:5,0.0620095719122297)
--(axis cs:6,0.0576972229623348)
--(axis cs:7,0.0558762891172758)
--(axis cs:8,0.0527129239237168)
--(axis cs:9,0.0512610605131476)
--(axis cs:10,0.0486921739674852)
--(axis cs:11,0.0492835344428186)
--(axis cs:12,0.0479273654460881)
--(axis cs:13,0.046685089078529)
--(axis cs:14,0.0464183188235851)
--(axis cs:15,0.0446068296299502)
--(axis cs:16,0.0449948709779444)
--(axis cs:17,0.0448817500654438)
--(axis cs:18,0.0431797471519886)
--(axis cs:19,0.0428694988945483)
--(axis cs:20,0.0422900072495229)
--(axis cs:21,0.0429211561216892)
--(axis cs:22,0.0423269368909165)
--(axis cs:23,0.0421276102631389)
--(axis cs:24,0.0402880268525567)
--(axis cs:25,0.0398624088192422)
--(axis cs:26,0.0399709149012658)
--(axis cs:27,0.0393199348994033)
--(axis cs:28,0.0397031796719075)
--(axis cs:29,0.0397568640062022)
--(axis cs:30,0.0409635877302343)
--(axis cs:31,0.0406992729136074)
--(axis cs:32,0.0382783208431217)
--(axis cs:33,0.0385560332931561)
--(axis cs:34,0.03995228634974)
--(axis cs:35,0.0382730350172494)
--(axis cs:36,0.0382947947097499)
--(axis cs:37,0.0367719004518759)
--(axis cs:38,0.0367090299050834)
--(axis cs:39,0.0378033942693738)
--(axis cs:40,0.0367542222886779)
--(axis cs:41,0.0374881720047231)
--(axis cs:42,0.0383937852108725)
--(axis cs:43,0.0366736331812837)
--(axis cs:44,0.0376383663064805)
--(axis cs:45,0.0389582562479773)
--(axis cs:46,0.037314659356516)
--(axis cs:47,0.0368457766391122)
--(axis cs:48,0.0357374948927334)
--(axis cs:49,0.0375682966388602)
--(axis cs:50,0.0368211980202161)
--(axis cs:51,0.0373866052642126)
--(axis cs:52,0.0393544758740038)
--(axis cs:53,0.0349141322798503)
--(axis cs:54,0.0367412907690889)
--(axis cs:55,0.0358730953455323)
--(axis cs:56,0.0358514408990812)
--(axis cs:57,0.036567519105619)
--(axis cs:58,0.0350571425437159)
--(axis cs:59,0.035980758103468)
--(axis cs:60,0.0348182799023274)
--(axis cs:61,0.0357334359722301)
--(axis cs:62,0.0348279064252335)
--(axis cs:63,0.0352410177654045)
--(axis cs:64,0.0340673363773529)
--(axis cs:65,0.0350138826474799)
--(axis cs:66,0.0351660101923138)
--(axis cs:67,0.0346118987898177)
--(axis cs:68,0.034387801240079)
--(axis cs:69,0.035083727671237)
--(axis cs:70,0.0351101808106007)
--(axis cs:71,0.0347932654655728)
--(axis cs:72,0.0374087053114869)
--(axis cs:73,0.0336334240671839)
--(axis cs:74,0.033879765931261)
--(axis cs:75,0.0342302156916497)
--(axis cs:76,0.0353287855880711)
--(axis cs:77,0.0340006275523045)
--(axis cs:78,0.0344887873515333)
--(axis cs:79,0.0345114510783562)
--(axis cs:80,0.0342958591447496)
--(axis cs:81,0.0342103019657174)
--(axis cs:82,0.033635047250124)
--(axis cs:83,0.033677132939134)
--(axis cs:84,0.0331379541588595)
--(axis cs:85,0.0339006998979118)
--(axis cs:86,0.0335147269411775)
--(axis cs:87,0.0332925648027697)
--(axis cs:88,0.0335589421993354)
--(axis cs:89,0.0339153702779194)
--(axis cs:90,0.0347877840924048)
--(axis cs:91,0.0336621221579046)
--(axis cs:92,0.0333038714302511)
--(axis cs:93,0.0333582641250657)
--(axis cs:94,0.0329417759308736)
--(axis cs:95,0.0329367311882738)
--(axis cs:96,0.0328647821125053)
--(axis cs:97,0.0328555129800681)
--(axis cs:98,0.0328515243575989)
--(axis cs:99,0.0330584979626145)
--(axis cs:100,0.0325445060674827)
--(axis cs:101,0.0328294809134733)
--(axis cs:102,0.0332406217638094)
--(axis cs:103,0.0327771011736077)
--(axis cs:104,0.0329469498875428)
--(axis cs:105,0.0328162908965713)
--(axis cs:106,0.032948450795214)
--(axis cs:107,0.0327715393100451)
--(axis cs:108,0.0327686396490625)
--(axis cs:109,0.0328232353849817)
--(axis cs:110,0.0328913885862501)
--(axis cs:111,0.033046566570045)
--(axis cs:112,0.0329915641207321)
--(axis cs:113,0.0330007758958691)
--(axis cs:114,0.0330358512470812)
--(axis cs:115,0.0330102495696189)
--(axis cs:116,0.0328720090588663)
--(axis cs:116,0.0396339338122609)
--(axis cs:116,0.0396339338122609)
--(axis cs:115,0.0398064163707656)
--(axis cs:114,0.0398464986800765)
--(axis cs:113,0.039792367361753)
--(axis cs:112,0.0397790818947131)
--(axis cs:111,0.0398642721199824)
--(axis cs:110,0.0395930048082706)
--(axis cs:109,0.0395770598189906)
--(axis cs:108,0.0395157512420035)
--(axis cs:107,0.0395213594356152)
--(axis cs:106,0.0396847768429556)
--(axis cs:105,0.0395063898277273)
--(axis cs:104,0.0396965186031606)
--(axis cs:103,0.0394142441613347)
--(axis cs:102,0.0407814893898602)
--(axis cs:101,0.0395332708533645)
--(axis cs:100,0.0391631037094088)
--(axis cs:99,0.0399721475682105)
--(axis cs:98,0.039521141915704)
--(axis cs:97,0.0397235096480331)
--(axis cs:96,0.0396974878130142)
--(axis cs:95,0.0396597366713203)
--(axis cs:94,0.0396850192935559)
--(axis cs:93,0.0396486412342825)
--(axis cs:92,0.0399589484075342)
--(axis cs:91,0.039436321047703)
--(axis cs:90,0.0402439037157695)
--(axis cs:89,0.0411107224686047)
--(axis cs:88,0.0394862609620944)
--(axis cs:87,0.0397728132435042)
--(axis cs:86,0.0403049861144778)
--(axis cs:85,0.040343414846145)
--(axis cs:84,0.0403697131453746)
--(axis cs:83,0.0407105683911807)
--(axis cs:82,0.0404581552306126)
--(axis cs:81,0.0405943465343554)
--(axis cs:80,0.0411905166712324)
--(axis cs:79,0.0404404866720274)
--(axis cs:78,0.0402786764921439)
--(axis cs:77,0.0399848109529291)
--(axis cs:76,0.0403298862620759)
--(axis cs:75,0.0408415439909676)
--(axis cs:74,0.0399020437762406)
--(axis cs:73,0.0408069458079643)
--(axis cs:72,0.0405486465416583)
--(axis cs:71,0.0413245530646927)
--(axis cs:70,0.0408030751316224)
--(axis cs:69,0.0407932303860494)
--(axis cs:68,0.0420997695373969)
--(axis cs:67,0.0411243529425725)
--(axis cs:66,0.0409571114085143)
--(axis cs:65,0.0411689201144491)
--(axis cs:64,0.0414278066445793)
--(axis cs:63,0.0409124452910489)
--(axis cs:62,0.0414849623329108)
--(axis cs:61,0.0421980122662005)
--(axis cs:60,0.0414691786395155)
--(axis cs:59,0.0419340551155122)
--(axis cs:58,0.0412483396782506)
--(axis cs:57,0.0421288498461471)
--(axis cs:56,0.0423735634638065)
--(axis cs:55,0.0424657048422455)
--(axis cs:54,0.0418145182821736)
--(axis cs:53,0.0427047091190673)
--(axis cs:52,0.0449373234716885)
--(axis cs:51,0.0427586583727292)
--(axis cs:50,0.0419154958093645)
--(axis cs:49,0.0432395451330251)
--(axis cs:48,0.0425135390000722)
--(axis cs:47,0.0444851236081923)
--(axis cs:46,0.0449261844662697)
--(axis cs:45,0.0428376340339838)
--(axis cs:44,0.0434120454030783)
--(axis cs:43,0.0429622737424548)
--(axis cs:42,0.045368625409599)
--(axis cs:41,0.0434009522191144)
--(axis cs:40,0.0443173858548915)
--(axis cs:39,0.0436464514923333)
--(axis cs:38,0.0442937036477132)
--(axis cs:37,0.0440697320629388)
--(axis cs:36,0.0456816217168607)
--(axis cs:35,0.0445279008127154)
--(axis cs:34,0.0450046754845898)
--(axis cs:33,0.0448471850946075)
--(axis cs:32,0.0446675182304723)
--(axis cs:31,0.0467858093105389)
--(axis cs:30,0.0457530097817669)
--(axis cs:29,0.0475026418961949)
--(axis cs:28,0.0464831425400655)
--(axis cs:27,0.0490506466362354)
--(axis cs:26,0.0470974682060475)
--(axis cs:25,0.0473301539440796)
--(axis cs:24,0.0471056576259823)
--(axis cs:23,0.0471207448100772)
--(axis cs:22,0.0485354666319688)
--(axis cs:21,0.0497428899783802)
--(axis cs:20,0.0490258605700508)
--(axis cs:19,0.04962469124744)
--(axis cs:18,0.0502080902355469)
--(axis cs:17,0.0500141678617177)
--(axis cs:16,0.0531344266048316)
--(axis cs:15,0.0511702600608287)
--(axis cs:14,0.0529021568470916)
--(axis cs:13,0.0541672788519161)
--(axis cs:12,0.054284399586951)
--(axis cs:11,0.0556838688316683)
--(axis cs:10,0.056335080171008)
--(axis cs:9,0.0567582657709666)
--(axis cs:8,0.0611988367550108)
--(axis cs:7,0.0622137963854745)
--(axis cs:6,0.0631221477651491)
--(axis cs:5,0.0666251346683403)
--(axis cs:4,0.0705366383464296)
--(axis cs:3,0.0766997109609573)
--(axis cs:2,0.0879561982389557)
--(axis cs:1,0.107221007651666)
--cycle;

\path [draw=color1, fill=color1, bars]
(axis cs:1,0.101611824187643)
--(axis cs:1,0.0932090348822386)
--(axis cs:2,0.0741432467929814)
--(axis cs:3,0.0655362976986739)
--(axis cs:4,0.0593530290217524)
--(axis cs:5,0.0531034543792949)
--(axis cs:6,0.0522347096562705)
--(axis cs:7,0.0498122683898964)
--(axis cs:8,0.0459528694608532)
--(axis cs:9,0.0442930643148413)
--(axis cs:10,0.0424108092883005)
--(axis cs:11,0.0430032963956618)
--(axis cs:12,0.0404739714319508)
--(axis cs:13,0.0400413401882929)
--(axis cs:14,0.0392571053683784)
--(axis cs:15,0.0389499037522053)
--(axis cs:16,0.0369903802846245)
--(axis cs:17,0.0369880588486427)
--(axis cs:18,0.0366009733429152)
--(axis cs:19,0.0364518758285749)
--(axis cs:20,0.0377719118562306)
--(axis cs:21,0.0378617259220904)
--(axis cs:22,0.0364494184217308)
--(axis cs:23,0.0353941270273475)
--(axis cs:24,0.0344908078576766)
--(axis cs:25,0.0347757335283657)
--(axis cs:26,0.0334583721340777)
--(axis cs:27,0.0346586605337886)
--(axis cs:28,0.0333732186124947)
--(axis cs:29,0.0340144173632988)
--(axis cs:30,0.0332093190927099)
--(axis cs:31,0.0337245609189565)
--(axis cs:32,0.0335095726903102)
--(axis cs:33,0.0333861657046552)
--(axis cs:34,0.0330841859680055)
--(axis cs:35,0.0326476934317068)
--(axis cs:36,0.0326214558764513)
--(axis cs:37,0.0334995647054816)
--(axis cs:38,0.0331934596991732)
--(axis cs:39,0.0325393098864452)
--(axis cs:40,0.0327273167282002)
--(axis cs:41,0.0324917476922221)
--(axis cs:42,0.0320688297984874)
--(axis cs:43,0.0318182678868758)
--(axis cs:44,0.0329827950857762)
--(axis cs:45,0.031398849594882)
--(axis cs:46,0.0316767989174668)
--(axis cs:47,0.0308022514899255)
--(axis cs:48,0.030335277983476)
--(axis cs:49,0.0323487832635316)
--(axis cs:50,0.0323069089050786)
--(axis cs:51,0.0310932084505407)
--(axis cs:52,0.0321442258595344)
--(axis cs:53,0.0299261206677732)
--(axis cs:54,0.032910356342708)
--(axis cs:55,0.0302918890377783)
--(axis cs:56,0.0309728996119791)
--(axis cs:57,0.0299094157325791)
--(axis cs:58,0.0326482933953427)
--(axis cs:59,0.0307387440203789)
--(axis cs:60,0.0319001872943361)
--(axis cs:61,0.0313616081555949)
--(axis cs:62,0.0290362933419487)
--(axis cs:63,0.0300006190282375)
--(axis cs:64,0.0304990777752841)
--(axis cs:65,0.0300497225291646)
--(axis cs:66,0.0305390872630378)
--(axis cs:67,0.0305308608086324)
--(axis cs:68,0.0304377656052218)
--(axis cs:69,0.0306356566905676)
--(axis cs:70,0.0302738042139656)
--(axis cs:71,0.0308369124698483)
--(axis cs:72,0.0296900112971212)
--(axis cs:73,0.0309259003975416)
--(axis cs:74,0.0301828095540916)
--(axis cs:75,0.0308821140665373)
--(axis cs:76,0.0304288443852512)
--(axis cs:77,0.0307567488180033)
--(axis cs:78,0.029838104667061)
--(axis cs:79,0.0300905344359896)
--(axis cs:80,0.0295848588658969)
--(axis cs:81,0.0304979569082429)
--(axis cs:82,0.0302296898520892)
--(axis cs:83,0.029838072186871)
--(axis cs:84,0.029658657864555)
--(axis cs:85,0.0301807326658039)
--(axis cs:86,0.0298402333511391)
--(axis cs:87,0.0295623870533399)
--(axis cs:88,0.0301658663074436)
--(axis cs:89,0.0307876889684387)
--(axis cs:90,0.0303378744229197)
--(axis cs:91,0.0304123044438695)
--(axis cs:92,0.029895169178293)
--(axis cs:93,0.0293431901754392)
--(axis cs:94,0.0294444146325805)
--(axis cs:95,0.0297003365757688)
--(axis cs:96,0.0295229744217005)
--(axis cs:97,0.0299314371773294)
--(axis cs:98,0.030100426614483)
--(axis cs:99,0.0299845820908962)
--(axis cs:100,0.0301237886025068)
--(axis cs:101,0.0308688099009315)
--(axis cs:102,0.0300996133139813)
--(axis cs:103,0.0302159495396669)
--(axis cs:104,0.0307326990316599)
--(axis cs:104,0.0359997468979645)
--(axis cs:104,0.0359997468979645)
--(axis cs:103,0.0359822185386046)
--(axis cs:102,0.035988066394619)
--(axis cs:101,0.0360215672586508)
--(axis cs:100,0.035986626804136)
--(axis cs:99,0.0359963917084204)
--(axis cs:98,0.0359880161359098)
--(axis cs:97,0.0360010048715889)
--(axis cs:96,0.0360497613753839)
--(axis cs:95,0.0360259646568978)
--(axis cs:94,0.036061404364223)
--(axis cs:93,0.0360773032284841)
--(axis cs:92,0.0360044113132603)
--(axis cs:91,0.0359793221557122)
--(axis cs:90,0.0359792831456792)
--(axis cs:89,0.0360074094876728)
--(axis cs:88,0.0359843963369224)
--(axis cs:87,0.0360441645127073)
--(axis cs:86,0.0360099504182846)
--(axis cs:85,0.0359836945558697)
--(axis cs:84,0.0360312257248825)
--(axis cs:83,0.0360101773698546)
--(axis cs:82,0.0359817156094665)
--(axis cs:81,0.0359813134739564)
--(axis cs:80,0.0360410495639895)
--(axis cs:79,0.0359886359648108)
--(axis cs:78,0.0360101739540247)
--(axis cs:77,0.0360029185456943)
--(axis cs:76,0.0359795376132605)
--(axis cs:75,0.0360242452162529)
--(axis cs:74,0.0359836001844279)
--(axis cs:73,0.0360338333311428)
--(axis cs:72,0.0360272480540481)
--(axis cs:71,0.0360155680389962)
--(axis cs:70,0.0359803884886143)
--(axis cs:69,0.0359895503467184)
--(axis cs:68,0.0349264951428899)
--(axis cs:67,0.0351398297556295)
--(axis cs:66,0.0350608322623086)
--(axis cs:65,0.0353139193544843)
--(axis cs:64,0.0351482842651578)
--(axis cs:63,0.0353042303558876)
--(axis cs:62,0.0367387589709355)
--(axis cs:61,0.035344906061294)
--(axis cs:60,0.0372193182408085)
--(axis cs:59,0.0344919290038604)
--(axis cs:58,0.0349087713757621)
--(axis cs:57,0.035892835231342)
--(axis cs:56,0.0372293584348576)
--(axis cs:55,0.0350506086818446)
--(axis cs:54,0.035942407834176)
--(axis cs:53,0.0349373956548663)
--(axis cs:52,0.0355111624504965)
--(axis cs:51,0.0387272753454462)
--(axis cs:50,0.0369326511711024)
--(axis cs:49,0.036993606155658)
--(axis cs:48,0.0375658612440268)
--(axis cs:47,0.0359018999557639)
--(axis cs:46,0.0355219672316064)
--(axis cs:45,0.0366397660677941)
--(axis cs:44,0.0383805522457019)
--(axis cs:43,0.0363213852092887)
--(axis cs:42,0.0368919497147543)
--(axis cs:41,0.0373219567870608)
--(axis cs:40,0.0379118890097274)
--(axis cs:39,0.0373553941335233)
--(axis cs:38,0.0372842992567844)
--(axis cs:37,0.0374992485940671)
--(axis cs:36,0.0381843773929797)
--(axis cs:35,0.038078999296237)
--(axis cs:34,0.0384989980553969)
--(axis cs:33,0.0376495503734212)
--(axis cs:32,0.0388364327560521)
--(axis cs:31,0.0394048790082566)
--(axis cs:30,0.0374997168675556)
--(axis cs:29,0.0393813897173765)
--(axis cs:28,0.0395646820965524)
--(axis cs:27,0.041136098684182)
--(axis cs:26,0.041261914560793)
--(axis cs:25,0.0399810482148214)
--(axis cs:24,0.0418830378269713)
--(axis cs:23,0.0428157337991019)
--(axis cs:22,0.0409067438008209)
--(axis cs:21,0.0453872706801419)
--(axis cs:20,0.0436477441774222)
--(axis cs:19,0.0434198985658819)
--(axis cs:18,0.0439823563765679)
--(axis cs:17,0.0453191897323674)
--(axis cs:16,0.044573642210471)
--(axis cs:15,0.047231038419687)
--(axis cs:14,0.0492461237638278)
--(axis cs:13,0.0467782680595345)
--(axis cs:12,0.0462847254619052)
--(axis cs:11,0.0492236233913537)
--(axis cs:10,0.0510855528451187)
--(axis cs:9,0.0488608803260631)
--(axis cs:8,0.0529810106743401)
--(axis cs:7,0.05314495325668)
--(axis cs:6,0.0561101426224653)
--(axis cs:5,0.0595263564290261)
--(axis cs:4,0.0668108865058028)
--(axis cs:3,0.0684840348689176)
--(axis cs:2,0.0815125296987103)
--(axis cs:1,0.101611824187643)
--cycle;

\path [draw=color2, fill=color2, bars]
(axis cs:1,0.102688036318593)
--(axis cs:1,0.0979577396046302)
--(axis cs:2,0.0789614464998856)
--(axis cs:3,0.067014249178115)
--(axis cs:4,0.0618047156069829)
--(axis cs:5,0.0565999882411249)
--(axis cs:6,0.0512984467222393)
--(axis cs:7,0.0484677548660322)
--(axis cs:8,0.0438455215663734)
--(axis cs:9,0.0417718725401147)
--(axis cs:10,0.0406284491414851)
--(axis cs:11,0.0404860022479925)
--(axis cs:12,0.0389211161623981)
--(axis cs:13,0.0370309406969662)
--(axis cs:14,0.0391108273060443)
--(axis cs:15,0.0355646733660079)
--(axis cs:16,0.0376700660088985)
--(axis cs:17,0.0343395043033846)
--(axis cs:18,0.0337418350553716)
--(axis cs:19,0.0330137330336286)
--(axis cs:20,0.0324866451129658)
--(axis cs:21,0.0322505225138877)
--(axis cs:22,0.0315388694160828)
--(axis cs:23,0.0310404339700432)
--(axis cs:24,0.0299997886262136)
--(axis cs:25,0.0324836669841441)
--(axis cs:26,0.0301932791105896)
--(axis cs:27,0.0315416782771143)
--(axis cs:28,0.0297356483248217)
--(axis cs:29,0.0302778967806648)
--(axis cs:30,0.0304875093932661)
--(axis cs:31,0.029045506658695)
--(axis cs:32,0.0305252250250413)
--(axis cs:33,0.0287476599484277)
--(axis cs:34,0.029541041657724)
--(axis cs:35,0.0278245287586128)
--(axis cs:36,0.0302494790588227)
--(axis cs:37,0.0293640640574497)
--(axis cs:38,0.0302556607491029)
--(axis cs:39,0.027552138898745)
--(axis cs:40,0.0281928813920117)
--(axis cs:41,0.0294498609272496)
--(axis cs:42,0.028647954926396)
--(axis cs:43,0.0281773000766484)
--(axis cs:44,0.0279925897901394)
--(axis cs:45,0.0291954172266946)
--(axis cs:46,0.0287462563735195)
--(axis cs:47,0.0285453324038575)
--(axis cs:48,0.0270883319926616)
--(axis cs:49,0.0274068591884619)
--(axis cs:50,0.0280092703076806)
--(axis cs:51,0.0280997697590023)
--(axis cs:52,0.026705931433735)
--(axis cs:53,0.0268654791511501)
--(axis cs:54,0.0290669539404411)
--(axis cs:55,0.0280177623767537)
--(axis cs:56,0.0269126147562777)
--(axis cs:57,0.0271567189786852)
--(axis cs:58,0.0273773433677189)
--(axis cs:59,0.0272530900749767)
--(axis cs:60,0.0277687816353635)
--(axis cs:61,0.0265585524336571)
--(axis cs:62,0.0277570594014203)
--(axis cs:63,0.0265848504665725)
--(axis cs:64,0.0270373972853831)
--(axis cs:65,0.027890932018281)
--(axis cs:66,0.0281226170458055)
--(axis cs:67,0.0275051129888559)
--(axis cs:68,0.0269563023835047)
--(axis cs:69,0.0268082053705455)
--(axis cs:70,0.0267166142711533)
--(axis cs:71,0.0278883441861973)
--(axis cs:72,0.0265062848246654)
--(axis cs:73,0.0266450305352899)
--(axis cs:74,0.0276150731384018)
--(axis cs:75,0.0267073196441527)
--(axis cs:76,0.0270505287286829)
--(axis cs:77,0.0278864336089648)
--(axis cs:78,0.0270887753545326)
--(axis cs:79,0.0264769750618395)
--(axis cs:80,0.0267505774833327)
--(axis cs:80,0.0368107762598839)
--(axis cs:80,0.0368107762598839)
--(axis cs:79,0.0368526711747175)
--(axis cs:78,0.0367662400266613)
--(axis cs:77,0.0367043539988049)
--(axis cs:76,0.0367708268399467)
--(axis cs:75,0.0368170765910669)
--(axis cs:74,0.0367171778220296)
--(axis cs:73,0.0368263689636076)
--(axis cs:72,0.0368479593529862)
--(axis cs:71,0.0367043001774727)
--(axis cs:70,0.0368157121463669)
--(axis cs:69,0.0371996571803245)
--(axis cs:68,0.0368681703819055)
--(axis cs:67,0.0368501645804866)
--(axis cs:66,0.0369709511131153)
--(axis cs:65,0.035347217669312)
--(axis cs:64,0.0354370230837963)
--(axis cs:63,0.0354467499064736)
--(axis cs:62,0.0361852459761922)
--(axis cs:61,0.0364140061614597)
--(axis cs:60,0.0359401006272998)
--(axis cs:59,0.0356643942776843)
--(axis cs:58,0.0362430344432878)
--(axis cs:57,0.0363166084301129)
--(axis cs:56,0.0353109787362808)
--(axis cs:55,0.0347908143998844)
--(axis cs:54,0.0357309637592327)
--(axis cs:53,0.0359372402853443)
--(axis cs:52,0.0358930495926539)
--(axis cs:51,0.0365474242026635)
--(axis cs:50,0.0352672904036777)
--(axis cs:49,0.036471945244891)
--(axis cs:48,0.0360984153897863)
--(axis cs:47,0.0366977957268172)
--(axis cs:46,0.0365177569134941)
--(axis cs:45,0.0364727963715936)
--(axis cs:44,0.0370078407494744)
--(axis cs:43,0.0367334152448949)
--(axis cs:42,0.037211632110294)
--(axis cs:41,0.0382923910355732)
--(axis cs:40,0.0377595176913151)
--(axis cs:39,0.0377296222933582)
--(axis cs:38,0.0381906024325763)
--(axis cs:37,0.0382528631444366)
--(axis cs:36,0.0372683410996778)
--(axis cs:35,0.0378167164185272)
--(axis cs:34,0.0375598589618611)
--(axis cs:33,0.0382772187964219)
--(axis cs:32,0.0390799776555625)
--(axis cs:31,0.0395292230835806)
--(axis cs:30,0.0417562393906822)
--(axis cs:29,0.0416257564788703)
--(axis cs:28,0.0401838560108173)
--(axis cs:27,0.0410973476383431)
--(axis cs:26,0.0421418797422522)
--(axis cs:25,0.0428697630362602)
--(axis cs:24,0.0413689834000007)
--(axis cs:23,0.0430401116581188)
--(axis cs:22,0.0443012931362599)
--(axis cs:21,0.0429416133106175)
--(axis cs:20,0.0445338241573023)
--(axis cs:19,0.0437790075026498)
--(axis cs:18,0.045097803951865)
--(axis cs:17,0.0448590333205625)
--(axis cs:16,0.0472088795406438)
--(axis cs:15,0.0467910844741887)
--(axis cs:14,0.0482862546429552)
--(axis cs:13,0.0477472773462741)
--(axis cs:12,0.0486868683778866)
--(axis cs:11,0.0495949136899227)
--(axis cs:10,0.0513196670926365)
--(axis cs:9,0.0525930359408148)
--(axis cs:8,0.0536929390507007)
--(axis cs:7,0.0566809050189093)
--(axis cs:6,0.0578152783819322)
--(axis cs:5,0.0622423857417646)
--(axis cs:4,0.0653747646858484)
--(axis cs:3,0.0729397176051805)
--(axis cs:2,0.0844677533999125)
--(axis cs:1,0.102688036318593)
--cycle;

\path [draw=color3, fill=color3, bars]
(axis cs:1,0.0894404174201978)
--(axis cs:1,0.0809850448132876)
--(axis cs:2,0.0642031437828166)
--(axis cs:3,0.0554387115254181)
--(axis cs:4,0.0482532372136349)
--(axis cs:5,0.04358124098021)
--(axis cs:6,0.0409036293125109)
--(axis cs:7,0.0394898877309006)
--(axis cs:8,0.0352520222413418)
--(axis cs:9,0.0367474742884562)
--(axis cs:10,0.0332779974487212)
--(axis cs:11,0.0322898837993606)
--(axis cs:12,0.0335964588848326)
--(axis cs:13,0.0326730852269508)
--(axis cs:14,0.031857766572583)
--(axis cs:15,0.0312475236816319)
--(axis cs:16,0.0302006131653286)
--(axis cs:17,0.0301144133513474)
--(axis cs:18,0.0311217534171984)
--(axis cs:19,0.030399282753694)
--(axis cs:20,0.0282754768181351)
--(axis cs:21,0.0283092334896575)
--(axis cs:22,0.0289693357281893)
--(axis cs:23,0.0287736682917178)
--(axis cs:24,0.0286691482851611)
--(axis cs:25,0.0299626483410199)
--(axis cs:26,0.0279113611968931)
--(axis cs:27,0.0274734085543186)
--(axis cs:28,0.0280627634911823)
--(axis cs:29,0.0265756384579702)
--(axis cs:30,0.0258348225078892)
--(axis cs:31,0.027353250041692)
--(axis cs:32,0.0258401800374265)
--(axis cs:33,0.0255245435644673)
--(axis cs:34,0.0256902836769487)
--(axis cs:35,0.0266244435050724)
--(axis cs:36,0.0278553943746229)
--(axis cs:37,0.0267520589232788)
--(axis cs:38,0.0261276683597204)
--(axis cs:39,0.0264557804264324)
--(axis cs:40,0.0260008403164793)
--(axis cs:41,0.0250506862294729)
--(axis cs:42,0.0246584385212344)
--(axis cs:43,0.0267632851905133)
--(axis cs:44,0.0262639611685998)
--(axis cs:45,0.0263141084915294)
--(axis cs:46,0.0262127395897091)
--(axis cs:47,0.0262605553927176)
--(axis cs:48,0.0257936654408506)
--(axis cs:49,0.0243006902729848)
--(axis cs:50,0.0241601614633458)
--(axis cs:51,0.0249501035568435)
--(axis cs:52,0.0242112275073189)
--(axis cs:53,0.0247794043293139)
--(axis cs:54,0.0241728590219885)
--(axis cs:55,0.0246675312918402)
--(axis cs:56,0.0239044973762135)
--(axis cs:57,0.0262033698347566)
--(axis cs:58,0.0239740879445412)
--(axis cs:59,0.024693466858382)
--(axis cs:60,0.0236357170786958)
--(axis cs:61,0.0234450861948455)
--(axis cs:62,0.0244053746552651)
--(axis cs:63,0.0233073673733129)
--(axis cs:64,0.0233328456051719)
--(axis cs:65,0.023961756741995)
--(axis cs:66,0.0238692558357411)
--(axis cs:67,0.0241422826976768)
--(axis cs:68,0.0237023322945251)
--(axis cs:69,0.0238876617516928)
--(axis cs:70,0.023881880360495)
--(axis cs:71,0.0230149451269366)
--(axis cs:72,0.023858143223257)
--(axis cs:73,0.0237703281485431)
--(axis cs:74,0.0234646704443061)
--(axis cs:75,0.0238445516202872)
--(axis cs:76,0.0232733333284333)
--(axis cs:77,0.0235433971493091)
--(axis cs:78,0.0250693599693202)
--(axis cs:79,0.0233560688154465)
--(axis cs:80,0.0238729262022426)
--(axis cs:81,0.0245478667398504)
--(axis cs:82,0.0235991365688809)
--(axis cs:82,0.0331780302836717)
--(axis cs:82,0.0331780302836717)
--(axis cs:81,0.0334843123048802)
--(axis cs:80,0.0332267970864291)
--(axis cs:79,0.0331516142393505)
--(axis cs:78,0.0339845671347933)
--(axis cs:77,0.0331707160676668)
--(axis cs:76,0.0331456163745186)
--(axis cs:75,0.0332206456654926)
--(axis cs:74,0.0331617014619809)
--(axis cs:73,0.03320582660314)
--(axis cs:72,0.0332235575935328)
--(axis cs:71,0.033135173148902)
--(axis cs:70,0.0332287964227883)
--(axis cs:69,0.0332301023799935)
--(axis cs:68,0.0331937716152441)
--(axis cs:67,0.0333006539975751)
--(axis cs:66,0.0332259856783449)
--(axis cs:65,0.0332479212927142)
--(axis cs:64,0.0331497889967403)
--(axis cs:63,0.0331479149575208)
--(axis cs:62,0.0334071634286256)
--(axis cs:61,0.0331518181303839)
--(axis cs:60,0.0331510399065486)
--(axis cs:59,0.0332376413723104)
--(axis cs:58,0.0331971689784791)
--(axis cs:57,0.0337938923341285)
--(axis cs:56,0.0331626109486597)
--(axis cs:55,0.033233217534744)
--(axis cs:54,0.033222984829868)
--(axis cs:53,0.0333270352434306)
--(axis cs:52,0.0332623994985984)
--(axis cs:51,0.0337274002351958)
--(axis cs:50,0.0355308669016069)
--(axis cs:49,0.0334588039697234)
--(axis cs:48,0.033560852826609)
--(axis cs:47,0.0334514218062641)
--(axis cs:46,0.0362406193454762)
--(axis cs:45,0.0364111015374357)
--(axis cs:44,0.0341478006753261)
--(axis cs:43,0.0343964546774745)
--(axis cs:42,0.0328164210464494)
--(axis cs:41,0.0346777819613997)
--(axis cs:40,0.0332576804664052)
--(axis cs:39,0.0334650629952629)
--(axis cs:38,0.0334617082321287)
--(axis cs:37,0.0345126027295184)
--(axis cs:36,0.0352707348185121)
--(axis cs:35,0.0342310417021626)
--(axis cs:34,0.0355302175359201)
--(axis cs:33,0.0362590750822412)
--(axis cs:32,0.0352449608443813)
--(axis cs:31,0.0357486898369806)
--(axis cs:30,0.0349460350237456)
--(axis cs:29,0.0359746887424467)
--(axis cs:28,0.0350072643491176)
--(axis cs:27,0.0379684178905277)
--(axis cs:26,0.0363920016954951)
--(axis cs:25,0.0367104564738349)
--(axis cs:24,0.0359448455865543)
--(axis cs:23,0.0370680678716163)
--(axis cs:22,0.036549691127053)
--(axis cs:21,0.0430338252661692)
--(axis cs:20,0.0377373798538428)
--(axis cs:19,0.0378071036641951)
--(axis cs:18,0.0393936120938568)
--(axis cs:17,0.0386897631567543)
--(axis cs:16,0.0378473164200245)
--(axis cs:15,0.0391198412194708)
--(axis cs:14,0.038580929861789)
--(axis cs:13,0.041547848194345)
--(axis cs:12,0.0422333346193413)
--(axis cs:11,0.0418353491915707)
--(axis cs:10,0.0435354653540408)
--(axis cs:9,0.0434469606647729)
--(axis cs:8,0.0469253699169778)
--(axis cs:7,0.0461614990007527)
--(axis cs:6,0.0469983870671614)
--(axis cs:5,0.0500793861154691)
--(axis cs:4,0.0533359282354136)
--(axis cs:3,0.0605708388046865)
--(axis cs:2,0.0677747432456676)
--(axis cs:1,0.0894404174201978)
--cycle;

\addplot [line0]
table {%
1 0.104006442235397
2 0.0844317811523169
3 0.0742358032560202
4 0.068065987963878
5 0.064317353290285
6 0.060409685363742
7 0.0590450427513752
8 0.0569558803393638
9 0.0540096631420571
10 0.0525136270692466
11 0.0524837016372434
12 0.0511058825165196
13 0.0504261839652226
14 0.0496602378353384
15 0.0478885448453894
16 0.049064648791388
17 0.0474479589635807
18 0.0466939186937677
19 0.0462470950709942
20 0.0456579339097869
21 0.0463320230500347
22 0.0454312017614427
23 0.044624177536608
24 0.0436968422392695
25 0.0435962813816609
26 0.0435341915536567
27 0.0441852907678193
28 0.0430931611059865
29 0.0436297529511986
30 0.0433582987560006
31 0.0437425411120731
32 0.041472919536797
33 0.0417016091938818
34 0.0424784809171649
35 0.0414004679149824
36 0.0419882082133053
37 0.0404208162574074
38 0.0405013667763983
39 0.0407249228808536
40 0.0405358040717847
41 0.0404445621119188
42 0.0418812053102358
43 0.0398179534618693
44 0.0405252058547794
45 0.0408979451409806
46 0.0411204219113928
47 0.0406654501236522
48 0.0391255169464028
49 0.0404039208859427
50 0.0393683469147903
51 0.0400726318184709
52 0.0421458996728461
53 0.0388094206994588
54 0.0392779045256312
55 0.0391694000938889
56 0.0391125021814439
57 0.0393481844758831
58 0.0381527411109832
59 0.0389574066094901
60 0.0381437292709215
61 0.0389657241192153
62 0.0381564343790722
63 0.0380767315282267
64 0.0377475715109661
65 0.0380914013809645
66 0.038061560800414
67 0.0378681258661951
68 0.038243785388738
69 0.0379384790286432
70 0.0379566279711116
71 0.0380589092651328
72 0.0389786759265726
73 0.0372201849375741
74 0.0368909048537508
75 0.0375358798413087
76 0.0378293359250735
77 0.0369927192526168
78 0.0373837319218386
79 0.0374759688751918
80 0.037743187907991
81 0.0374023242500364
82 0.0370466012403683
83 0.0371938506651574
84 0.036753833652117
85 0.0371220573720284
86 0.0369098565278276
87 0.036532689023137
88 0.0365226015807149
89 0.037513046373262
90 0.0375158439040872
91 0.0365492216028038
92 0.0366314099188926
93 0.0365034526796741
94 0.0363133976122148
95 0.0362982339297971
96 0.0362811349627598
97 0.0362895113140506
98 0.0361863331366515
99 0.0365153227654125
100 0.0358538048884458
101 0.0361813758834189
102 0.0370110555768348
103 0.0360956726674712
104 0.0363217342453517
105 0.0361613403621493
106 0.0363166138190848
107 0.0361464493728302
108 0.036142195445533
109 0.0362001476019862
110 0.0362421966972603
111 0.0364554193450137
112 0.0363853230077226
113 0.036396571628811
114 0.0364411749635788
115 0.0364083329701923
116 0.0362529714355636
} node[linelabel,right,pos=1] {LSTM};
\addlegendentry{LSTM}
\addplot [line1]
table {%
1 0.0974104295349409
2 0.0778278882458459
3 0.0670101662837958
4 0.0630819577637776
5 0.0563149054041605
6 0.0541724261393679
7 0.0514786108232882
8 0.0494669400675967
9 0.0465769723204522
10 0.0467481810667096
11 0.0461134598935077
12 0.043379348446928
13 0.0434098041239137
14 0.0442516145661031
15 0.0430904710859462
16 0.0407820112475477
17 0.0411536242905051
18 0.0402916648597416
19 0.0399358871972284
20 0.0407098280168264
21 0.0416244983011162
22 0.0386780811112758
23 0.0391049304132247
24 0.0381869228423239
25 0.0373783908715936
26 0.0373601433474354
27 0.0378973796089853
28 0.0364689503545236
29 0.0366979035403376
30 0.0353545179801328
31 0.0365647199636066
32 0.0361730027231812
33 0.0355178580390382
34 0.0357915920117012
35 0.0353633463639719
36 0.0354029166347155
37 0.0354994066497744
38 0.0352388794779788
39 0.0349473520099843
40 0.0353196028689638
41 0.0349068522396414
42 0.0344803897566209
43 0.0340698265480823
44 0.0356816736657391
45 0.034019307831338
46 0.0335993830745366
47 0.0333520757228447
48 0.0339505696137514
49 0.0346711947095948
50 0.0346197800380905
51 0.0349102418979935
52 0.0338276941550155
53 0.0324317581613198
54 0.034426382088442
55 0.0326712488598115
56 0.0341011290234183
57 0.0329011254819606
58 0.0337785323855524
59 0.0326153365121197
60 0.0345597527675723
61 0.0333532571084445
62 0.0328875261564421
63 0.0326524246920625
64 0.0328236810202209
65 0.0326818209418245
66 0.0327999597626732
67 0.0328353452821309
68 0.0326821303740559
69 0.033312603518643
70 0.0331270963512899
71 0.0334262402544223
72 0.0328586296755847
73 0.0334798668643422
74 0.0330832048692598
75 0.0334531796413951
76 0.0332041909992558
77 0.0333798336818488
78 0.0329241393105428
79 0.0330395852004002
80 0.0328129542149432
81 0.0332396351910996
82 0.0331057027307778
83 0.0329241247783628
84 0.0328449417947188
85 0.0330822136108368
86 0.0329250918847118
87 0.0328032757830236
88 0.033075131322183
89 0.0333975492280557
90 0.0331585787842994
91 0.0331958132997908
92 0.0329497902457767
93 0.0327102467019617
94 0.0327529094984018
95 0.0328631506163333
96 0.0327863678985422
97 0.0329662210244591
98 0.0330442213751964
99 0.0329904868996583
100 0.0330552077033214
101 0.0334451885797912
102 0.0330438398543002
103 0.0330990840391358
104 0.0333662229648122
} node[below right,linelabel] {Gref};
\addlegendentry{Gref}
\addplot [line2]
table {%
1 0.100322887961611
2 0.081714599949899
3 0.0699769833916478
4 0.0635897401464156
5 0.0594211869914448
6 0.0545568625520857
7 0.0525743299424708
8 0.0487692303085371
9 0.0471824542404648
10 0.0459740581170608
11 0.0450404579689576
12 0.0438039922701424
13 0.0423891090216202
14 0.0436985409744997
15 0.0411778789200983
16 0.0424394727747711
17 0.0395992688119735
18 0.0394198195036183
19 0.0383963702681392
20 0.0385102346351341
21 0.0375960679122526
22 0.0379200812761713
23 0.037040272814081
24 0.0356843860131072
25 0.0376767150102022
26 0.0361675794264209
27 0.0363195129577287
28 0.0349597521678195
29 0.0359518266297675
30 0.0361218743919741
31 0.0342873648711378
32 0.0348026013403019
33 0.0335124393724248
34 0.0335504503097925
35 0.03282062258857
36 0.0337589100792502
37 0.0338084636009431
38 0.0342231315908396
39 0.0326408805960516
40 0.0329761995416634
41 0.0338711259814114
42 0.032929793518345
43 0.0324553576607717
44 0.0325002152698069
45 0.0328341067991441
46 0.0326320066435068
47 0.0326215640653374
48 0.0315933736912239
49 0.0319394022166765
50 0.0316382803556792
51 0.0323235969808329
52 0.0312994905131944
53 0.0314013597182472
54 0.0323989588498369
55 0.031404288388319
56 0.0311117967462792
57 0.0317366637043991
58 0.0318101889055034
59 0.0314587421763305
60 0.0318544411313316
61 0.0314862792975584
62 0.0319711526888062
63 0.0310158001865231
64 0.0312372101845897
65 0.0316190748437965
66 0.0325467840794604
67 0.0321776387846713
68 0.0319122363827051
69 0.032003931275435
70 0.0317661632087601
71 0.032296322181835
72 0.0316771220888258
73 0.0317356997494488
74 0.0321661254802157
75 0.0317621981176098
76 0.0319106777843148
77 0.0322953938038848
78 0.0319275076905969
79 0.0316648231182785
80 0.0317806768716083
} coordinate[pos=1] (jm);
\draw[<-,shorten <=2pt] (jm) -- +(0.3cm,-0.2cm) node[linelabel,right] {JM};
\addlegendentry{JM}
\addplot [line3]
table {%
1 0.0852127311167427
2 0.0659889435142421
3 0.0580047751650523
4 0.0507945827245242
5 0.0468303135478395
6 0.0439510081898362
7 0.0428256933658266
8 0.0410886960791598
9 0.0400972174766145
10 0.038406731401381
11 0.0370626164954656
12 0.0379148967520869
13 0.0371104667106479
14 0.035219348217186
15 0.0351836824505514
16 0.0340239647926766
17 0.0344020882540508
18 0.0352576827555276
19 0.0341031932089445
20 0.0330064283359889
21 0.0356715293779133
22 0.0327595134276212
23 0.0329208680816671
24 0.0323069969358577
25 0.0333365524074274
26 0.0321516814461941
27 0.0327209132224231
28 0.0315350139201499
29 0.0312751636002085
30 0.0303904287658174
31 0.0315509699393363
32 0.0305425704409039
33 0.0308918093233542
34 0.0306102506064344
35 0.0304277426036175
36 0.0315630645965675
37 0.0306323308263986
38 0.0297946882959246
39 0.0299604217108477
40 0.0296292603914422
41 0.0298642340954363
42 0.0287374297838419
43 0.0305798699339939
44 0.0302058809219629
45 0.0313626050144825
46 0.0312266794675927
47 0.0298559885994909
48 0.0296772591337298
49 0.0288797471213541
50 0.0298455141824764
51 0.0293387518960196
52 0.0287368135029586
53 0.0290532197863722
54 0.0286979219259283
55 0.0289503744132921
56 0.0285335541624366
57 0.0299986310844426
58 0.0285856284615101
59 0.0289655541153462
60 0.0283933784926222
61 0.0282984521626147
62 0.0289062690419454
63 0.0282276411654169
64 0.0282413173009561
65 0.0286048390173546
66 0.028547620757043
67 0.028721468347626
68 0.0284480519548846
69 0.0285588820658431
70 0.0285553383916417
71 0.0280750591379193
72 0.0285408504083949
73 0.0284880773758415
74 0.0283131859531435
75 0.0285325986428899
76 0.0282094748514759
77 0.0283570566084879
78 0.0295269635520568
79 0.0282538415273985
80 0.0285498616443359
81 0.0290160895223653
82 0.0283885834262763
} node[linelabel,pos=1,below left] {Ours};
\addlegendentry{Ours}
\end{axis}

\end{tikzpicture}

%% file: figures/test-marked-reversal.tex
\begin{tikzpicture}

\begin{axis}[
title={Marked Reversal},
ylabel={Difference in Cross Entropy},
ymin=-0.02,
ymax=0.3,
]
\path [draw=color0, fill=color0, bars]
(axis cs:41,0.0194466518492119)
--(axis cs:41,-0.000816977529935828)
--(axis cs:43,-0.000768626156918924)
--(axis cs:45,-2.67686374417972e-05)
--(axis cs:47,-0.00236705821367008)
--(axis cs:49,-0.00192022537589498)
--(axis cs:51,0.0014665466484165)
--(axis cs:53,0.00668278136214589)
--(axis cs:55,0.0127382636916889)
--(axis cs:57,0.0218259640135858)
--(axis cs:59,0.0259448712963542)
--(axis cs:61,0.0374055663213408)
--(axis cs:63,0.0498827165997935)
--(axis cs:65,0.0554815259243539)
--(axis cs:67,0.0659826924051326)
--(axis cs:69,0.0751169835781432)
--(axis cs:71,0.0884876439935411)
--(axis cs:73,0.0954606384433024)
--(axis cs:75,0.109658465825182)
--(axis cs:77,0.118081988052704)
--(axis cs:79,0.125739569987985)
--(axis cs:81,0.135945136428749)
--(axis cs:83,0.153394935814441)
--(axis cs:85,0.159615887183495)
--(axis cs:87,0.169400825036114)
--(axis cs:89,0.177204679690069)
--(axis cs:91,0.184065740687101)
--(axis cs:93,0.192133915165529)
--(axis cs:95,0.199668202413858)
--(axis cs:97,0.205383372419119)
--(axis cs:99,0.21418669470891)
--(axis cs:99,0.250576759153981)
--(axis cs:99,0.250576759153981)
--(axis cs:97,0.24001519922826)
--(axis cs:95,0.22215561286552)
--(axis cs:93,0.211970232349639)
--(axis cs:91,0.199789874031641)
--(axis cs:89,0.184637282078477)
--(axis cs:87,0.175664463819005)
--(axis cs:85,0.163452792496097)
--(axis cs:83,0.155208830482369)
--(axis cs:81,0.144029433532732)
--(axis cs:79,0.134222172949516)
--(axis cs:77,0.128992364910702)
--(axis cs:75,0.121671547303014)
--(axis cs:73,0.104868602701178)
--(axis cs:71,0.097479838673422)
--(axis cs:69,0.0834248428146021)
--(axis cs:67,0.075244376032841)
--(axis cs:65,0.0651314569058632)
--(axis cs:63,0.0595204884313015)
--(axis cs:61,0.048896671587741)
--(axis cs:59,0.0437703103532028)
--(axis cs:57,0.0359669334700752)
--(axis cs:55,0.0298432751242393)
--(axis cs:53,0.026858468120091)
--(axis cs:51,0.020994929257043)
--(axis cs:49,0.0209293740420583)
--(axis cs:47,0.0172908649220349)
--(axis cs:45,0.0187134091549556)
--(axis cs:43,0.0184360457797796)
--(axis cs:41,0.0194466518492119)
--cycle;

\path [draw=color1, fill=color1, bars]
(axis cs:41,0.0193567009166357)
--(axis cs:41,-0.000780499348884044)
--(axis cs:43,-0.0114104363990373)
--(axis cs:45,-0.0182197292887715)
--(axis cs:47,-0.0177417049675659)
--(axis cs:49,-0.0168448619304006)
--(axis cs:51,-0.016286789262)
--(axis cs:53,-0.0139519246477533)
--(axis cs:55,-0.0130977919268024)
--(axis cs:57,-0.0129978769013005)
--(axis cs:59,-0.0123825080856555)
--(axis cs:61,-0.0121255313405142)
--(axis cs:63,-0.0133196127383394)
--(axis cs:65,-0.0136237510778479)
--(axis cs:67,-0.0152083475691835)
--(axis cs:69,-0.0155984462910427)
--(axis cs:71,-0.016287598972402)
--(axis cs:73,-0.0153783757291266)
--(axis cs:75,-0.0128844905538978)
--(axis cs:77,-0.00981055767983873)
--(axis cs:79,-0.00579366449298889)
--(axis cs:81,-0.00194134584544955)
--(axis cs:83,0.00221918502965596)
--(axis cs:85,0.0090009153888259)
--(axis cs:87,0.0132457809884857)
--(axis cs:89,0.0198388394734637)
--(axis cs:91,0.0232161346550613)
--(axis cs:93,0.0305060114779472)
--(axis cs:95,0.0355560453732349)
--(axis cs:97,0.0397188013710608)
--(axis cs:99,0.0483550665473422)
--(axis cs:99,0.289764092807972)
--(axis cs:99,0.289764092807972)
--(axis cs:97,0.279573405505056)
--(axis cs:95,0.248987630103511)
--(axis cs:93,0.232853239363834)
--(axis cs:91,0.21763823521478)
--(axis cs:89,0.196225954106879)
--(axis cs:87,0.185383579346519)
--(axis cs:85,0.168127242691502)
--(axis cs:83,0.160669584561582)
--(axis cs:81,0.138475329387178)
--(axis cs:79,0.124159073866407)
--(axis cs:77,0.110298262915972)
--(axis cs:75,0.101887706807094)
--(axis cs:73,0.0916661731664154)
--(axis cs:71,0.0807469252793299)
--(axis cs:69,0.0682433212571576)
--(axis cs:67,0.0608685579148065)
--(axis cs:65,0.0499885217686419)
--(axis cs:63,0.046685615388482)
--(axis cs:61,0.0365760466573828)
--(axis cs:59,0.031257730121865)
--(axis cs:57,0.0230178508652247)
--(axis cs:55,0.0177286915381851)
--(axis cs:53,0.0113877790828202)
--(axis cs:51,0.00909411190459193)
--(axis cs:49,0.00617594737908935)
--(axis cs:47,0.00468108473111681)
--(axis cs:45,0.00459100522295195)
--(axis cs:43,0.00956095921957255)
--(axis cs:41,0.0193567009166357)
--cycle;

\path [draw=color2, fill=color2, bars]
(axis cs:41,0.0190878671219198)
--(axis cs:41,0.00380443105406365)
--(axis cs:43,-0.00661438041323217)
--(axis cs:45,-0.0101865292557058)
--(axis cs:47,-0.00921222926817065)
--(axis cs:49,-0.00948219138090969)
--(axis cs:51,-0.0100869623387586)
--(axis cs:53,-0.0109525937961509)
--(axis cs:55,-0.0112974107912307)
--(axis cs:57,-0.0117384023479034)
--(axis cs:59,-0.0120785224315729)
--(axis cs:61,-0.0128965585983899)
--(axis cs:63,-0.0143642264928673)
--(axis cs:65,-0.014599172505381)
--(axis cs:67,-0.0162648378828156)
--(axis cs:69,-0.0155900785368988)
--(axis cs:71,-0.0157474771194657)
--(axis cs:73,-0.0142529835929606)
--(axis cs:75,-0.0122685408250926)
--(axis cs:77,-0.00884483215269483)
--(axis cs:79,-0.00499164497692336)
--(axis cs:81,-0.000392528907275244)
--(axis cs:83,0.00482227243671599)
--(axis cs:85,0.010640555514449)
--(axis cs:87,0.0180801478132484)
--(axis cs:89,0.0243967643469727)
--(axis cs:91,0.0318884337729505)
--(axis cs:93,0.0385650130795161)
--(axis cs:95,0.0464342007091042)
--(axis cs:97,0.0540782525926733)
--(axis cs:99,0.0610857677143592)
--(axis cs:99,0.205291598711663)
--(axis cs:99,0.205291598711663)
--(axis cs:97,0.195896551335634)
--(axis cs:95,0.191633374849879)
--(axis cs:93,0.181930389475974)
--(axis cs:91,0.169978051910764)
--(axis cs:89,0.162578646333089)
--(axis cs:87,0.149954553308394)
--(axis cs:85,0.147190647143084)
--(axis cs:83,0.135703521815667)
--(axis cs:81,0.121788467503016)
--(axis cs:79,0.112444166197652)
--(axis cs:77,0.102928950741101)
--(axis cs:75,0.0966409367657886)
--(axis cs:73,0.0864373456706603)
--(axis cs:71,0.0794028174008654)
--(axis cs:69,0.0654684510120717)
--(axis cs:67,0.0610775377340356)
--(axis cs:65,0.0497267587009826)
--(axis cs:63,0.0453641577144384)
--(axis cs:61,0.0393392614152586)
--(axis cs:59,0.0334152116287993)
--(axis cs:57,0.0271305151550293)
--(axis cs:55,0.0192341697776134)
--(axis cs:53,0.0181973736439538)
--(axis cs:51,0.014654088902919)
--(axis cs:49,0.0104443255381444)
--(axis cs:47,0.00980166720991308)
--(axis cs:45,0.0145833694259974)
--(axis cs:43,0.0127755772889998)
--(axis cs:41,0.0190878671219198)
--cycle;

\path [draw=color3, fill=color3, bars]
(axis cs:41,0.0121192978733589)
--(axis cs:41,0.000182942185246403)
--(axis cs:43,-0.00994335041423798)
--(axis cs:45,-0.014983980307888)
--(axis cs:47,-0.0141921120814764)
--(axis cs:49,-0.013258154580046)
--(axis cs:51,-0.0120903944548824)
--(axis cs:53,-0.0104931699735299)
--(axis cs:55,-0.00946161607098959)
--(axis cs:57,-0.00807147242790289)
--(axis cs:59,-0.0075108770865417)
--(axis cs:61,-0.00759058541116822)
--(axis cs:63,-0.00744987967017754)
--(axis cs:65,-0.00769422774441207)
--(axis cs:67,-0.00763541110240963)
--(axis cs:69,-0.00761032267621896)
--(axis cs:71,-0.00713491462837997)
--(axis cs:73,-0.00562313788964988)
--(axis cs:75,-0.00327830365186461)
--(axis cs:77,0.000177461162527003)
--(axis cs:79,0.00532952143662427)
--(axis cs:81,0.0110449449652165)
--(axis cs:83,0.0166252103939504)
--(axis cs:85,0.0220831319920724)
--(axis cs:87,0.0271106214799934)
--(axis cs:89,0.0291191672381051)
--(axis cs:91,0.0310170214478608)
--(axis cs:93,0.0321840604222034)
--(axis cs:95,0.0336195038623382)
--(axis cs:97,0.0334337630809231)
--(axis cs:99,0.0203559829036216)
--(axis cs:99,0.330392299715582)
--(axis cs:99,0.330392299715582)
--(axis cs:97,0.234959190815168)
--(axis cs:95,0.199238850890724)
--(axis cs:93,0.166907181767023)
--(axis cs:91,0.131477694404123)
--(axis cs:89,0.0978489247095132)
--(axis cs:87,0.0716374529354706)
--(axis cs:85,0.0486722779264903)
--(axis cs:83,0.0325827892424683)
--(axis cs:81,0.0207867213504315)
--(axis cs:79,0.0114606431780918)
--(axis cs:77,0.00373820673594402)
--(axis cs:75,-0.00205427436577286)
--(axis cs:73,-0.00486612749412289)
--(axis cs:71,-0.00545720052595976)
--(axis cs:69,-0.00551767373447782)
--(axis cs:67,-0.00524932957808678)
--(axis cs:65,-0.00510604062729396)
--(axis cs:63,-0.00541767112710054)
--(axis cs:61,-0.00582823200761881)
--(axis cs:59,-0.00639321430415565)
--(axis cs:57,-0.00686949110817299)
--(axis cs:55,-0.00711566968694599)
--(axis cs:53,-0.00724135810909173)
--(axis cs:51,-0.00763374353365316)
--(axis cs:49,-0.00804103141578053)
--(axis cs:47,-0.00799861952465342)
--(axis cs:45,-0.00699834618848702)
--(axis cs:43,-0.00154961332045941)
--(axis cs:41,0.0121192978733589)
--cycle;

\addplot [line0]
table {%
41 0.00931483715963805
43 0.00883370981143032
45 0.0093433202587569
47 0.00746190335418239
49 0.00950457433308165
51 0.0112307379527298
53 0.0167706247411185
55 0.0212907694079641
57 0.0288964487418305
59 0.0348575908247785
61 0.0431511189545409
63 0.0547016025155475
65 0.0603064914151085
67 0.0706135342189868
69 0.0792709131963726
71 0.0929837413334816
73 0.10016462057224
75 0.115665006564098
77 0.123537176481703
79 0.12998087146875
81 0.139987284980741
83 0.154301883148405
85 0.161534339839796
87 0.17253264442756
89 0.180920980884273
91 0.191927807359371
93 0.202052073757584
95 0.210911907639689
97 0.22269928582369
99 0.232381726931445
} node[linelabel,pos=1,right] {LSTM};
\addlegendentry{LSTM}
\addplot [line1]
table {%
41 0.00928810078387584
43 -0.000924738589732377
45 -0.00681436203290976
47 -0.00653031011822454
49 -0.0053344572756556
51 -0.00359633867870405
53 -0.00128207278246653
55 0.00231544980569132
57 0.0050099869819621
59 0.00943761101810473
61 0.0122252576584343
63 0.0166830013250713
65 0.018182385345397
67 0.0228301051728115
69 0.0263224374830575
71 0.032229663153464
73 0.0381438987186444
75 0.0445016081265979
77 0.0502438526180667
79 0.0591827046867093
81 0.0682669917708641
83 0.081444384795619
85 0.0885640790401637
87 0.0993146801675021
89 0.108032396790172
91 0.120427184934921
93 0.13167962542089
95 0.142271837738373
97 0.159646103438058
99 0.169059579677657
} node[linelabel,pos=0.9,above left] {Gref};
\addlegendentry{Gref}
\addplot [line2]
table {%
41 0.0114461490879917
43 0.00308059843788381
45 0.00219842008514581
47 0.000294718970871211
49 0.000481067078617348
51 0.00228356328208023
53 0.00362238992390143
55 0.00396837949319135
57 0.00769605640356296
59 0.0106683445986132
61 0.0132213514084343
63 0.0154999656107856
65 0.0175637930978008
67 0.02240634992561
69 0.0249391862375864
71 0.0318276701406999
73 0.0360921810388499
75 0.042186197970348
77 0.0470420592942031
79 0.0537262606103644
81 0.0606979692978703
83 0.0702628971261913
85 0.0789156013287666
87 0.0840173505608211
89 0.0934877053400311
91 0.100933242841857
93 0.110247701277745
95 0.119033787779492
97 0.124987401964154
99 0.133188683213011
} node[linelabel,pos=1,right] {JM};
\addlegendentry{JM}
\addplot [line3]
table {%
41 0.00615112002930267
43 -0.0057464818673487
45 -0.0109911632481875
47 -0.0110953658030649
49 -0.0106495929979133
51 -0.00986206899426778
53 -0.00886726404131081
55 -0.00828864287896779
57 -0.00747048176803794
59 -0.00695204569534867
61 -0.00670940870939352
63 -0.00643377539863904
65 -0.00640013418585301
67 -0.0064423703402482
69 -0.00656399820534839
71 -0.00629605757716987
73 -0.00524463269188639
75 -0.00266628900881873
77 0.00195783394923551
79 0.00839508230735806
81 0.015915833157824
83 0.0246039998182093
85 0.0353777049592813
87 0.049374037207732
89 0.0634840459738092
91 0.0812473579259921
93 0.0995456210946134
95 0.116429177376531
97 0.134196476948045
99 0.175374141309602
} node[linelabel,pos=0.8,below right] {Ours};
\addlegendentry{Ours}
\legend{} 
\end{axis}

\end{tikzpicture}

%% file: figures/test-unmarked-reversal.tex
\begin{tikzpicture}

\begin{axis}[
title={Unmarked Reversal},
ylabel={Difference in Cross Entropy},
ymax=0.328256274987751,
]
\path [draw=color0, fill=color0, bars]
(axis cs:40,0.181685486959855)
--(axis cs:40,0.144919119716879)
--(axis cs:42,0.132917774978263)
--(axis cs:44,0.127943320726502)
--(axis cs:46,0.121387345219)
--(axis cs:48,0.122128194137432)
--(axis cs:50,0.132533586469534)
--(axis cs:52,0.132479072253934)
--(axis cs:54,0.144345567446157)
--(axis cs:56,0.144524776842252)
--(axis cs:58,0.150252193554495)
--(axis cs:60,0.162091231147468)
--(axis cs:62,0.165384504096011)
--(axis cs:64,0.173169787250933)
--(axis cs:66,0.181898623133316)
--(axis cs:68,0.190448576470662)
--(axis cs:70,0.194177628243421)
--(axis cs:72,0.204372190368375)
--(axis cs:74,0.2071692753159)
--(axis cs:76,0.216394983158359)
--(axis cs:78,0.223277185062583)
--(axis cs:80,0.225172545849746)
--(axis cs:82,0.233323843922851)
--(axis cs:84,0.242654655065135)
--(axis cs:86,0.249554318216498)
--(axis cs:88,0.250486860593374)
--(axis cs:90,0.251551160263491)
--(axis cs:92,0.260102922715376)
--(axis cs:94,0.268237680956334)
--(axis cs:96,0.269947960950872)
--(axis cs:98,0.28047700598532)
--(axis cs:100,0.277905909427444)
--(axis cs:100,0.302969822172908)
--(axis cs:100,0.302969822172908)
--(axis cs:98,0.296636358388712)
--(axis cs:96,0.296435113669178)
--(axis cs:94,0.287988540715421)
--(axis cs:92,0.278940137523913)
--(axis cs:90,0.275860667479671)
--(axis cs:88,0.261295315704974)
--(axis cs:86,0.267222695050937)
--(axis cs:84,0.251826677735393)
--(axis cs:82,0.243407169653237)
--(axis cs:80,0.238575197970366)
--(axis cs:78,0.239088081858623)
--(axis cs:76,0.236144542249192)
--(axis cs:74,0.221648179433678)
--(axis cs:72,0.212098199129384)
--(axis cs:70,0.206497292273842)
--(axis cs:68,0.199506569560822)
--(axis cs:66,0.192706522378211)
--(axis cs:64,0.182363699382943)
--(axis cs:62,0.173709025481618)
--(axis cs:60,0.169939098143082)
--(axis cs:58,0.165617870720124)
--(axis cs:56,0.159483337052572)
--(axis cs:54,0.1645867370419)
--(axis cs:52,0.147282348546067)
--(axis cs:50,0.145870154947365)
--(axis cs:48,0.137591064210792)
--(axis cs:46,0.136827825724417)
--(axis cs:44,0.146046310796615)
--(axis cs:42,0.159087439486309)
--(axis cs:40,0.181685486959855)
--cycle;

\path [draw=color1, fill=color1, bars]
(axis cs:40,0.152577930432876)
--(axis cs:40,0.126020279759483)
--(axis cs:42,0.0991327373623337)
--(axis cs:44,0.083118957823784)
--(axis cs:46,0.0797348273862628)
--(axis cs:48,0.0832355007703479)
--(axis cs:50,0.0869632434116827)
--(axis cs:52,0.0937585417790436)
--(axis cs:54,0.102275146410409)
--(axis cs:56,0.112030364883507)
--(axis cs:58,0.114320023396926)
--(axis cs:60,0.12310501354434)
--(axis cs:62,0.130755810784822)
--(axis cs:64,0.134190948942749)
--(axis cs:66,0.143810713867485)
--(axis cs:68,0.15374463367516)
--(axis cs:70,0.157617805708657)
--(axis cs:72,0.16965631676395)
--(axis cs:74,0.176174527763642)
--(axis cs:76,0.185104297040207)
--(axis cs:78,0.201134559167872)
--(axis cs:80,0.196124788929592)
--(axis cs:82,0.205435463807105)
--(axis cs:84,0.211157919447028)
--(axis cs:86,0.225995451822541)
--(axis cs:88,0.229436585216726)
--(axis cs:90,0.235685741614872)
--(axis cs:92,0.244906748033027)
--(axis cs:94,0.254343848400868)
--(axis cs:96,0.260655246179298)
--(axis cs:98,0.267825172938296)
--(axis cs:100,0.272979582490333)
--(axis cs:100,0.290625895203769)
--(axis cs:100,0.290625895203769)
--(axis cs:98,0.291578509516094)
--(axis cs:96,0.28518044481445)
--(axis cs:94,0.274929663829397)
--(axis cs:92,0.262882469645121)
--(axis cs:90,0.262142817899123)
--(axis cs:88,0.252228270414008)
--(axis cs:86,0.250474898109138)
--(axis cs:84,0.238520330206179)
--(axis cs:82,0.229492124654654)
--(axis cs:80,0.228134795710833)
--(axis cs:78,0.22063064765718)
--(axis cs:76,0.214846566771948)
--(axis cs:74,0.20966259020384)
--(axis cs:72,0.198452283454294)
--(axis cs:70,0.19231968456307)
--(axis cs:68,0.184692661368273)
--(axis cs:66,0.180646216090064)
--(axis cs:64,0.168545159151088)
--(axis cs:62,0.157218688424863)
--(axis cs:60,0.150998704418085)
--(axis cs:58,0.14419970749946)
--(axis cs:56,0.136345058581631)
--(axis cs:54,0.141107570404037)
--(axis cs:52,0.124598619706054)
--(axis cs:50,0.123621513630216)
--(axis cs:48,0.117314832610429)
--(axis cs:46,0.11320023995661)
--(axis cs:44,0.118789019757571)
--(axis cs:42,0.121699509282298)
--(axis cs:40,0.152577930432876)
--cycle;

\path [draw=color2, fill=color2, bars]
(axis cs:40,0.167486612370971)
--(axis cs:40,0.106688428876076)
--(axis cs:42,0.0784022682019458)
--(axis cs:44,0.0643765014745411)
--(axis cs:46,0.0685752212558613)
--(axis cs:48,0.0729738230721517)
--(axis cs:50,0.08407024878105)
--(axis cs:52,0.084605761884122)
--(axis cs:54,0.102920260418912)
--(axis cs:56,0.0960488972132564)
--(axis cs:58,0.108545480032762)
--(axis cs:60,0.119748664649029)
--(axis cs:62,0.125551334940846)
--(axis cs:64,0.137219466129468)
--(axis cs:66,0.14449797161566)
--(axis cs:68,0.152314807587251)
--(axis cs:70,0.162465812787486)
--(axis cs:72,0.174991854188361)
--(axis cs:74,0.178711937560595)
--(axis cs:76,0.194097050883399)
--(axis cs:78,0.202202761137199)
--(axis cs:80,0.200005056509171)
--(axis cs:82,0.212641907404687)
--(axis cs:84,0.218965732447428)
--(axis cs:86,0.221931945888501)
--(axis cs:88,0.236803107454494)
--(axis cs:90,0.241472017064648)
--(axis cs:92,0.248634800011284)
--(axis cs:94,0.25789405297675)
--(axis cs:96,0.263625522582024)
--(axis cs:98,0.272627492573197)
--(axis cs:100,0.274579621954306)
--(axis cs:100,0.313500172146046)
--(axis cs:100,0.313500172146046)
--(axis cs:98,0.306349232770223)
--(axis cs:96,0.295142460078391)
--(axis cs:94,0.293462363176388)
--(axis cs:92,0.28433089949431)
--(axis cs:90,0.276884775956292)
--(axis cs:88,0.268655671294138)
--(axis cs:86,0.268729764871666)
--(axis cs:84,0.259338377510838)
--(axis cs:82,0.24631890800067)
--(axis cs:80,0.238967736139066)
--(axis cs:78,0.241740029822469)
--(axis cs:76,0.235293702936981)
--(axis cs:74,0.217372250719388)
--(axis cs:72,0.210120163998633)
--(axis cs:70,0.205596454269956)
--(axis cs:68,0.198332319142763)
--(axis cs:66,0.18519715081348)
--(axis cs:64,0.174916605343276)
--(axis cs:62,0.168373091814203)
--(axis cs:60,0.163377579355063)
--(axis cs:58,0.155870213282805)
--(axis cs:56,0.14861199639697)
--(axis cs:54,0.150809895631645)
--(axis cs:52,0.135345656662274)
--(axis cs:50,0.132745250448349)
--(axis cs:48,0.128819064019562)
--(axis cs:46,0.131713319252773)
--(axis cs:44,0.13603777937386)
--(axis cs:42,0.146214390645067)
--(axis cs:40,0.167486612370971)
--cycle;

\path [draw=color3, fill=color3, bars]
(axis cs:40,0.241886940895305)
--(axis cs:40,0.155207314218929)
--(axis cs:42,0.11928419536995)
--(axis cs:44,0.124258982328071)
--(axis cs:46,0.0714404717531174)
--(axis cs:48,0.0359677987528458)
--(axis cs:50,0.0183781153119509)
--(axis cs:52,0.023532360306569)
--(axis cs:54,0.0189096909399133)
--(axis cs:56,0.0233046176329023)
--(axis cs:58,0.0281695885521018)
--(axis cs:60,0.0328381089248815)
--(axis cs:62,0.0342148073974884)
--(axis cs:64,0.0445532477903284)
--(axis cs:66,0.0622972554424961)
--(axis cs:68,0.0655070193389973)
--(axis cs:70,0.0872377187264658)
--(axis cs:72,0.0904437098010844)
--(axis cs:74,0.0965077361054431)
--(axis cs:76,0.119622361289043)
--(axis cs:78,0.129525030271045)
--(axis cs:80,0.149484906430495)
--(axis cs:82,0.152355387054733)
--(axis cs:84,0.164903822309356)
--(axis cs:86,0.177077739712524)
--(axis cs:88,0.18101861072821)
--(axis cs:90,0.198021086770059)
--(axis cs:92,0.202831000200942)
--(axis cs:94,0.212658283366708)
--(axis cs:96,0.224953306122384)
--(axis cs:98,0.237091755299386)
--(axis cs:100,0.244782064587142)
--(axis cs:100,0.28842231935696)
--(axis cs:100,0.28842231935696)
--(axis cs:98,0.286601846638422)
--(axis cs:96,0.274737751896103)
--(axis cs:94,0.266837469555046)
--(axis cs:92,0.264278020466337)
--(axis cs:90,0.251554955382825)
--(axis cs:88,0.244297057224967)
--(axis cs:86,0.242351299354329)
--(axis cs:84,0.235811443712899)
--(axis cs:82,0.225778970652453)
--(axis cs:80,0.221720356920867)
--(axis cs:78,0.220757554358495)
--(axis cs:76,0.212590105627553)
--(axis cs:74,0.200377431376398)
--(axis cs:72,0.19860196995275)
--(axis cs:70,0.193070195652404)
--(axis cs:68,0.184377799831274)
--(axis cs:66,0.1759159925785)
--(axis cs:64,0.16861888020097)
--(axis cs:62,0.157250981504737)
--(axis cs:60,0.158893961902127)
--(axis cs:58,0.148990649483293)
--(axis cs:56,0.143374882980673)
--(axis cs:54,0.139307914474069)
--(axis cs:52,0.131234018426125)
--(axis cs:50,0.126125567511198)
--(axis cs:48,0.124101781047202)
--(axis cs:46,0.127499117498723)
--(axis cs:44,0.186371473946466)
--(axis cs:42,0.224021730590158)
--(axis cs:40,0.241886940895305)
--cycle;

\addplot [line0]
table {%
40 0.163302303338367
42 0.146002607232286
44 0.136994815761558
46 0.129107585471708
48 0.129859629174112
50 0.139201870708449
52 0.139880710400001
54 0.154466152244028
56 0.152004056947412
58 0.157935032137309
60 0.166015164645275
62 0.169546764788814
64 0.177766743316938
66 0.187302572755763
68 0.194977573015742
70 0.200337460258631
72 0.208235194748879
74 0.214408727374789
76 0.226269762703775
78 0.231182633460603
80 0.231873871910056
82 0.238365506788044
84 0.247240666400264
86 0.258388506633717
88 0.255891088149174
90 0.263705913871581
92 0.269521530119645
94 0.278113110835877
96 0.283191537310025
98 0.288556682187016
100 0.290437865800176
} node[linelabel,pos=0.75,above left] {LSTM};
\addlegendentry{LSTM}
\addplot [line1]
table {%
40 0.139299105096179
42 0.110416123322316
44 0.100953988790678
46 0.0964675336714365
48 0.100275166690388
50 0.105292378520949
52 0.109178580742549
54 0.121691358407223
56 0.124187711732569
58 0.129259865448193
60 0.137051858981213
62 0.143987249604843
64 0.151368054046919
66 0.162228464978775
68 0.169218647521716
70 0.174968745135864
72 0.184054300109122
74 0.192918558983741
76 0.199975431906078
78 0.210882603412526
80 0.212129792320212
82 0.21746379423088
84 0.224839124826603
86 0.238235174965839
88 0.240832427815367
90 0.248914279756998
92 0.253894608839074
94 0.264636756115133
96 0.272917845496874
98 0.279701841227195
100 0.281802738847051
} coordinate[pos=1] (gref);
\draw[<-,shorten <=2pt] (gref) to +(6pt,-2pt) node[linelabel,right] {Gref};
\addlegendentry{Gref}
\addplot [line2]
table {%
40 0.137087520623523
42 0.112308329423506
44 0.1002071404242
46 0.100144270254317
48 0.100896443545857
50 0.108407749614699
52 0.109975709273198
54 0.126865078025278
56 0.122330446805113
58 0.132207846657783
60 0.141563122002046
62 0.146962213377524
64 0.156068035736372
66 0.16484756121457
68 0.175323563365007
70 0.184031133528721
72 0.192556009093497
74 0.198042094139991
76 0.21469537691019
78 0.221971395479834
80 0.219486396324119
82 0.229480407702679
84 0.239152054979133
86 0.245330855380084
88 0.252729389374316
90 0.25917839651047
92 0.266482849752797
94 0.275678208076569
96 0.279383991330207
98 0.28948836267171
100 0.294039897050176
} coordinate[pos=1] (jm);
\draw[<-,shorten <=2pt] (jm) to +(6pt,2pt) node[linelabel,right] {JM};
\addlegendentry{JM}
\addplot [line3]
table {%
40 0.198547127557117
42 0.171652962980054
44 0.155315228137269
46 0.0994697946259202
48 0.0800347899000236
50 0.0722518414115745
52 0.077383189366347
54 0.0791088027069913
56 0.0833397503067874
58 0.0885801190176973
60 0.0958660354135042
62 0.0957328944511128
64 0.106586063995649
66 0.119106624010498
68 0.124942409585136
70 0.140153957189435
72 0.144522839876917
74 0.14844258374092
76 0.166106233458298
78 0.17514129231477
80 0.185602631675681
82 0.189067178853593
84 0.200357633011127
86 0.209714519533427
88 0.212657833976588
90 0.224788021076442
92 0.233554510333639
94 0.239747876460877
96 0.249845529009244
98 0.261846800968904
100 0.266602191972051
} node[linelabel,pos=1,below right] {Ours};
\addlegendentry{Ours}
\legend{} 
\end{axis}

\end{tikzpicture}

%% file: figures/test-padded-reversal.tex
\begin{tikzpicture}

\begin{axis}[
title={Padded Reversal},
ylabel={Difference in Cross Entropy},
ymin=0,
ymax=0.273867973455597,
]
\path [draw=color0, fill=color0, bars]
(axis cs:40,0.102426763883698)
--(axis cs:40,0.0731691377703081)
--(axis cs:41,0.0663246674071087)
--(axis cs:42,0.0758162645893886)
--(axis cs:43,0.0578082209615011)
--(axis cs:44,0.0587663265918057)
--(axis cs:45,0.0602719003579985)
--(axis cs:46,0.0513690933342016)
--(axis cs:47,0.0725257997492907)
--(axis cs:48,0.0784386122050984)
--(axis cs:49,0.0416019517553801)
--(axis cs:50,0.0692685948250318)
--(axis cs:51,0.068653988081107)
--(axis cs:52,0.0675825880250679)
--(axis cs:53,0.0513771992300213)
--(axis cs:54,0.076904439166578)
--(axis cs:55,0.0617470767794215)
--(axis cs:56,0.0717780283280657)
--(axis cs:57,0.0872126599319593)
--(axis cs:58,0.0732036656555888)
--(axis cs:59,0.0709703890081449)
--(axis cs:60,0.0752413401110318)
--(axis cs:61,0.0850405030414462)
--(axis cs:62,0.0780201673335957)
--(axis cs:63,0.0848875742796538)
--(axis cs:64,0.0946052729199026)
--(axis cs:65,0.0917180814182568)
--(axis cs:66,0.113335216926843)
--(axis cs:67,0.0981035702809249)
--(axis cs:68,0.0950926677458477)
--(axis cs:69,0.106064626531598)
--(axis cs:70,0.128285645011811)
--(axis cs:71,0.121325929610659)
--(axis cs:72,0.1159021881988)
--(axis cs:73,0.131927773706042)
--(axis cs:74,0.0994394822772161)
--(axis cs:75,0.119908241800158)
--(axis cs:76,0.147138314313579)
--(axis cs:77,0.143476596982251)
--(axis cs:78,0.154112733596264)
--(axis cs:79,0.145114282420556)
--(axis cs:80,0.155357766111191)
--(axis cs:81,0.142167535941383)
--(axis cs:82,0.129169874984635)
--(axis cs:83,0.159237133658511)
--(axis cs:84,0.155144493417912)
--(axis cs:85,0.163855588363519)
--(axis cs:86,0.169202376806545)
--(axis cs:87,0.167234712269447)
--(axis cs:88,0.18845656616389)
--(axis cs:89,0.199713719773669)
--(axis cs:90,0.191134371931458)
--(axis cs:91,0.189393958290563)
--(axis cs:92,0.207827642364714)
--(axis cs:93,0.193605003735587)
--(axis cs:94,0.218520972707918)
--(axis cs:95,0.223068895596183)
--(axis cs:96,0.180382680133031)
--(axis cs:97,0.208479703225512)
--(axis cs:98,0.228704072990064)
--(axis cs:99,0.223307352483139)
--(axis cs:100,0.232750864106579)
--(axis cs:100,0.247907612040571)
--(axis cs:100,0.247907612040571)
--(axis cs:99,0.23970683345971)
--(axis cs:98,0.246432425271482)
--(axis cs:97,0.225417552447608)
--(axis cs:96,0.198236159510372)
--(axis cs:95,0.241833014387402)
--(axis cs:94,0.235066305276204)
--(axis cs:93,0.211050657331138)
--(axis cs:92,0.223223740818082)
--(axis cs:91,0.2042489368456)
--(axis cs:90,0.207050366721057)
--(axis cs:89,0.203744935898656)
--(axis cs:88,0.202578536025369)
--(axis cs:87,0.185260129404335)
--(axis cs:86,0.184350133821565)
--(axis cs:85,0.175856964979179)
--(axis cs:84,0.165899039805129)
--(axis cs:83,0.171762025131685)
--(axis cs:82,0.149419920590774)
--(axis cs:81,0.158772140725851)
--(axis cs:80,0.173426973293988)
--(axis cs:79,0.164608421322485)
--(axis cs:78,0.172504441392469)
--(axis cs:77,0.161755147390495)
--(axis cs:76,0.168833020133972)
--(axis cs:75,0.139349224354915)
--(axis cs:74,0.121186936018214)
--(axis cs:73,0.15222023560849)
--(axis cs:72,0.138474818566576)
--(axis cs:71,0.14291858411188)
--(axis cs:70,0.15319818316827)
--(axis cs:69,0.125365251841587)
--(axis cs:68,0.119273621802457)
--(axis cs:67,0.128035270764013)
--(axis cs:66,0.138419528372674)
--(axis cs:65,0.11599519996886)
--(axis cs:64,0.122243821828709)
--(axis cs:63,0.111418125835924)
--(axis cs:62,0.104791042036321)
--(axis cs:61,0.106789888327243)
--(axis cs:60,0.104112058027374)
--(axis cs:59,0.0991893407623213)
--(axis cs:58,0.0995064011621948)
--(axis cs:57,0.118640090657299)
--(axis cs:56,0.0989189562632128)
--(axis cs:55,0.0886592682058033)
--(axis cs:54,0.105922733130431)
--(axis cs:53,0.081781107141952)
--(axis cs:52,0.0970183342236969)
--(axis cs:51,0.100117296441497)
--(axis cs:50,0.100753240219427)
--(axis cs:49,0.073187274637056)
--(axis cs:48,0.110089219030103)
--(axis cs:47,0.104467048511703)
--(axis cs:46,0.0775273104820673)
--(axis cs:45,0.0894447049948405)
--(axis cs:44,0.0945382306443989)
--(axis cs:43,0.090698235570767)
--(axis cs:42,0.106374014779789)
--(axis cs:41,0.0939010805441843)
--(axis cs:40,0.102426763883698)
--cycle;

\path [draw=color1, fill=color1, bars]
(axis cs:40,0.0982882418006164)
--(axis cs:40,0.0628839293846395)
--(axis cs:41,0.0563789256340231)
--(axis cs:42,0.061323400878339)
--(axis cs:43,0.0377631955432516)
--(axis cs:44,0.0339850210659774)
--(axis cs:45,0.0421109132960451)
--(axis cs:46,0.0264941757796225)
--(axis cs:47,0.0502955033026166)
--(axis cs:48,0.0476407573606018)
--(axis cs:49,0.00435461294392306)
--(axis cs:50,0.0327185620145453)
--(axis cs:51,0.0346206586185624)
--(axis cs:52,0.0247460401319555)
--(axis cs:53,0.00963117221945402)
--(axis cs:54,0.0247226481655333)
--(axis cs:55,0.0156956620174971)
--(axis cs:56,0.0155931149920216)
--(axis cs:57,0.0229080296814216)
--(axis cs:58,0.00369179082623868)
--(axis cs:59,0.00951582917836112)
--(axis cs:60,0.0137474817624599)
--(axis cs:61,0.0216850950602428)
--(axis cs:62,0.00583326451737571)
--(axis cs:63,0.000586733177225289)
--(axis cs:64,0.0199636755972429)
--(axis cs:65,0.0163413579924323)
--(axis cs:66,0.0267027546836708)
--(axis cs:67,0.0130965662547841)
--(axis cs:68,0.0129767185862048)
--(axis cs:69,0.0242734113752386)
--(axis cs:70,0.0339199194753302)
--(axis cs:71,0.0242729836785491)
--(axis cs:72,0.0239745906686346)
--(axis cs:73,0.032143587401529)
--(axis cs:74,0.0256647460068778)
--(axis cs:75,0.0330248354678448)
--(axis cs:76,0.0511209505759901)
--(axis cs:77,0.0450312708749757)
--(axis cs:78,0.0661218150531991)
--(axis cs:79,0.0446309314997673)
--(axis cs:80,0.0615096294757913)
--(axis cs:81,0.0408129331340893)
--(axis cs:82,0.043230187993803)
--(axis cs:83,0.0607986014397161)
--(axis cs:84,0.0671712926892381)
--(axis cs:85,0.0687602468756889)
--(axis cs:86,0.0728191552333753)
--(axis cs:87,0.0749878300649258)
--(axis cs:88,0.0968418443915979)
--(axis cs:89,0.109158875037323)
--(axis cs:90,0.108457562224145)
--(axis cs:91,0.116777440556844)
--(axis cs:92,0.113071632682238)
--(axis cs:93,0.121729736912668)
--(axis cs:94,0.12038701976259)
--(axis cs:95,0.140681756506037)
--(axis cs:96,0.101300118959692)
--(axis cs:97,0.134218072615577)
--(axis cs:98,0.158294820465556)
--(axis cs:99,0.149231502920141)
--(axis cs:100,0.150715526650083)
--(axis cs:100,0.240833730747067)
--(axis cs:100,0.240833730747067)
--(axis cs:99,0.24045883911488)
--(axis cs:98,0.234723792751347)
--(axis cs:97,0.218404094462183)
--(axis cs:96,0.185068985169388)
--(axis cs:95,0.230611549447942)
--(axis cs:94,0.221887623165016)
--(axis cs:93,0.200792618122472)
--(axis cs:92,0.21128531902977)
--(axis cs:91,0.197650632292232)
--(axis cs:90,0.195826861758231)
--(axis cs:89,0.202962120434861)
--(axis cs:88,0.190156847552633)
--(axis cs:87,0.167328132298511)
--(axis cs:86,0.169136323690519)
--(axis cs:85,0.161115139647157)
--(axis cs:84,0.148364834934845)
--(axis cs:83,0.153773364202889)
--(axis cs:82,0.127533709620477)
--(axis cs:81,0.137382566739163)
--(axis cs:80,0.151565991277044)
--(axis cs:79,0.140125803592166)
--(axis cs:78,0.151915018889861)
--(axis cs:77,0.134585429362218)
--(axis cs:76,0.138729341519587)
--(axis cs:75,0.110701901520562)
--(axis cs:74,0.0923459232387208)
--(axis cs:73,0.122448329768311)
--(axis cs:72,0.105246776448304)
--(axis cs:71,0.108765832970927)
--(axis cs:70,0.115330049887787)
--(axis cs:69,0.0990535599553741)
--(axis cs:68,0.0822188448591583)
--(axis cs:67,0.0822972596315719)
--(axis cs:66,0.100143077263573)
--(axis cs:65,0.0850749432264156)
--(axis cs:64,0.0838130619736347)
--(axis cs:63,0.073902374676448)
--(axis cs:62,0.0661951670047584)
--(axis cs:61,0.071963175406807)
--(axis cs:60,0.0650559000999039)
--(axis cs:59,0.0559079614369144)
--(axis cs:58,0.0594635548169759)
--(axis cs:57,0.0734497750470907)
--(axis cs:56,0.0593525181065114)
--(axis cs:55,0.0504671194023868)
--(axis cs:54,0.0680763032836747)
--(axis cs:53,0.0402101843441466)
--(axis cs:52,0.0576408083487804)
--(axis cs:51,0.0616570980241397)
--(axis cs:50,0.0597652261549133)
--(axis cs:49,0.0369673880818039)
--(axis cs:48,0.0738437490861886)
--(axis cs:47,0.0717563334889621)
--(axis cs:46,0.0576732984340649)
--(axis cs:45,0.0656710783415163)
--(axis cs:44,0.0667589671006249)
--(axis cs:43,0.0751821577004991)
--(axis cs:42,0.0983967682787851)
--(axis cs:41,0.0910977036053797)
--(axis cs:40,0.0982882418006164)
--cycle;

\path [draw=color2, fill=color2, bars]
(axis cs:40,0.0820086022655702)
--(axis cs:40,0.0703253120837481)
--(axis cs:41,0.0624544172380357)
--(axis cs:42,0.0644433892431728)
--(axis cs:43,0.0367172202607491)
--(axis cs:44,0.0244492024050601)
--(axis cs:45,0.0337262365991916)
--(axis cs:46,0.0215664683460859)
--(axis cs:47,0.0374564775318101)
--(axis cs:48,0.0409760243729261)
--(axis cs:49,-0.00600533917968686)
--(axis cs:50,0.0225431647842782)
--(axis cs:51,0.0247922188034026)
--(axis cs:52,0.0113045035904971)
--(axis cs:53,0.00165722505108931)
--(axis cs:54,0.0105019943514541)
--(axis cs:55,0.00454858582297173)
--(axis cs:56,0.00383864385260006)
--(axis cs:57,0.0130226564492696)
--(axis cs:58,-0.00715327613331132)
--(axis cs:59,-0.000745052890912985)
--(axis cs:60,0.00336090525303425)
--(axis cs:61,0.0119144998771403)
--(axis cs:62,0.00371883924198114)
--(axis cs:63,-0.00563965836052892)
--(axis cs:64,0.0146678082367857)
--(axis cs:65,0.0106774133369751)
--(axis cs:66,0.0211675113017378)
--(axis cs:67,0.00632366181435128)
--(axis cs:68,0.00729910765200807)
--(axis cs:69,0.0114141509767586)
--(axis cs:70,0.0297630234496424)
--(axis cs:71,0.0164514746188222)
--(axis cs:72,0.0133675303683662)
--(axis cs:73,0.028309235220042)
--(axis cs:74,0.0127299260594551)
--(axis cs:75,0.0249712984986495)
--(axis cs:76,0.03896093788374)
--(axis cs:77,0.0411368673492102)
--(axis cs:78,0.0546536398233323)
--(axis cs:79,0.0453284302114036)
--(axis cs:80,0.0534357917154883)
--(axis cs:81,0.0474090212315819)
--(axis cs:82,0.0350266793428923)
--(axis cs:83,0.058849422420735)
--(axis cs:84,0.070651450356036)
--(axis cs:85,0.0804959563034416)
--(axis cs:86,0.0876190838373525)
--(axis cs:87,0.0779881781264169)
--(axis cs:88,0.110249778237433)
--(axis cs:89,0.125085842910645)
--(axis cs:90,0.121549694936724)
--(axis cs:91,0.138972015199137)
--(axis cs:92,0.137142371932057)
--(axis cs:93,0.133167993876528)
--(axis cs:94,0.142783297319154)
--(axis cs:95,0.159405943057084)
--(axis cs:96,0.121405368448932)
--(axis cs:97,0.157034787064428)
--(axis cs:98,0.186811147108265)
--(axis cs:99,0.182341338164162)
--(axis cs:100,0.183518691583574)
--(axis cs:100,0.221977870501076)
--(axis cs:100,0.221977870501076)
--(axis cs:99,0.217068444369596)
--(axis cs:98,0.214926873793587)
--(axis cs:97,0.195217494381887)
--(axis cs:96,0.166274353762831)
--(axis cs:95,0.214054416762027)
--(axis cs:94,0.198071434122414)
--(axis cs:93,0.183062662989929)
--(axis cs:92,0.186354536046934)
--(axis cs:91,0.17523136242329)
--(axis cs:90,0.170021436944958)
--(axis cs:89,0.173765061708309)
--(axis cs:88,0.163849288795576)
--(axis cs:87,0.147259716708284)
--(axis cs:86,0.14522122884381)
--(axis cs:85,0.136587288675286)
--(axis cs:84,0.127841162108076)
--(axis cs:83,0.130823646855153)
--(axis cs:82,0.106294216175351)
--(axis cs:81,0.114899325863893)
--(axis cs:80,0.127208713314691)
--(axis cs:79,0.115978937791922)
--(axis cs:78,0.127410663070048)
--(axis cs:77,0.113621650055191)
--(axis cs:76,0.117017900675653)
--(axis cs:75,0.0953124697397572)
--(axis cs:74,0.0772992582832719)
--(axis cs:73,0.105121669535414)
--(axis cs:72,0.0843908376166275)
--(axis cs:71,0.0927208835139288)
--(axis cs:70,0.101833597699189)
--(axis cs:69,0.0827968783248687)
--(axis cs:68,0.0723068849064065)
--(axis cs:67,0.0722909546503629)
--(axis cs:66,0.091048053126567)
--(axis cs:65,0.0712745549491805)
--(axis cs:64,0.0754995085577247)
--(axis cs:63,0.0617208731090434)
--(axis cs:62,0.0583177513023303)
--(axis cs:61,0.068209340518188)
--(axis cs:60,0.0573321494608921)
--(axis cs:59,0.0567891593165699)
--(axis cs:58,0.0550229183147803)
--(axis cs:57,0.0718985220456901)
--(axis cs:56,0.057154622825732)
--(axis cs:55,0.0521290482247531)
--(axis cs:54,0.0631487366116426)
--(axis cs:53,0.0444065666540207)
--(axis cs:52,0.0579763596889167)
--(axis cs:51,0.0641618073705494)
--(axis cs:50,0.0620834847133055)
--(axis cs:49,0.033432938499419)
--(axis cs:48,0.0755194887470414)
--(axis cs:47,0.0731381721753272)
--(axis cs:46,0.0505951146481721)
--(axis cs:45,0.0590029013057308)
--(axis cs:44,0.0579596916564285)
--(axis cs:43,0.0609507324561121)
--(axis cs:42,0.0800699695530882)
--(axis cs:41,0.0739801840859708)
--(axis cs:40,0.0820086022655702)
--cycle;

\path [draw=color3, fill=color3, bars]
(axis cs:40,0.0725092942459742)
--(axis cs:40,0.0614953354353754)
--(axis cs:41,0.0531409556456884)
--(axis cs:42,0.0599182385794346)
--(axis cs:43,0.0382456870966099)
--(axis cs:44,0.0307089801178797)
--(axis cs:45,0.0411601369830055)
--(axis cs:46,0.0292017259175673)
--(axis cs:47,0.0527037258781713)
--(axis cs:48,0.0556691263179592)
--(axis cs:49,0.0137628385024633)
--(axis cs:50,0.0404386347828087)
--(axis cs:51,0.041688459000825)
--(axis cs:52,0.0367538731408679)
--(axis cs:53,0.0203646369229839)
--(axis cs:54,0.0416820446040142)
--(axis cs:55,0.0305010871681456)
--(axis cs:56,0.0354639885924023)
--(axis cs:57,0.0458123981601308)
--(axis cs:58,0.0329442166290936)
--(axis cs:59,0.0323479442270263)
--(axis cs:60,0.0402561773986469)
--(axis cs:61,0.0484234353515656)
--(axis cs:62,0.0394797286671083)
--(axis cs:63,0.0424558911714284)
--(axis cs:64,0.056877766557401)
--(axis cs:65,0.0523376359963765)
--(axis cs:66,0.0712296592851475)
--(axis cs:67,0.0505590270456319)
--(axis cs:68,0.0485476950919018)
--(axis cs:69,0.0611562272856758)
--(axis cs:70,0.0829197983969742)
--(axis cs:71,0.0701828892127437)
--(axis cs:72,0.0658238909014146)
--(axis cs:73,0.086306940328278)
--(axis cs:74,0.0556811298228379)
--(axis cs:75,0.0817667664926301)
--(axis cs:76,0.0925236084996058)
--(axis cs:77,0.0894690308712842)
--(axis cs:78,0.116429684726429)
--(axis cs:79,0.0989068482794028)
--(axis cs:80,0.107679710287741)
--(axis cs:81,0.100441344968201)
--(axis cs:82,0.087568942859258)
--(axis cs:83,0.11348179392917)
--(axis cs:84,0.119402985122127)
--(axis cs:85,0.125022094225119)
--(axis cs:86,0.139116563450493)
--(axis cs:87,0.130846050817233)
--(axis cs:88,0.156389582310105)
--(axis cs:89,0.159781249236844)
--(axis cs:90,0.162551562328538)
--(axis cs:91,0.170949999722198)
--(axis cs:92,0.172949785370247)
--(axis cs:93,0.167063024233018)
--(axis cs:94,0.193535335384623)
--(axis cs:95,0.201689721232207)
--(axis cs:96,0.166848089758535)
--(axis cs:97,0.190963401264544)
--(axis cs:98,0.20893857640494)
--(axis cs:99,0.202157817416208)
--(axis cs:100,0.207393209145192)
--(axis cs:100,0.260486009189459)
--(axis cs:100,0.260486009189459)
--(axis cs:99,0.24411719949255)
--(axis cs:98,0.25212328457982)
--(axis cs:97,0.232361508242338)
--(axis cs:96,0.200322385627056)
--(axis cs:95,0.248195186284273)
--(axis cs:94,0.234917571755084)
--(axis cs:93,0.216028981759783)
--(axis cs:92,0.228010586433472)
--(axis cs:91,0.209375845276603)
--(axis cs:90,0.207453684136477)
--(axis cs:89,0.210760128960059)
--(axis cs:88,0.208240423998472)
--(axis cs:87,0.18568086027106)
--(axis cs:86,0.181629227064971)
--(axis cs:85,0.175210668216844)
--(axis cs:84,0.167947874644812)
--(axis cs:83,0.172025436208917)
--(axis cs:82,0.150663645208528)
--(axis cs:81,0.159657318485298)
--(axis cs:80,0.170549987613532)
--(axis cs:79,0.162358848936739)
--(axis cs:78,0.17166780877192)
--(axis cs:77,0.158512485214124)
--(axis cs:76,0.16184107504334)
--(axis cs:75,0.13243228820411)
--(axis cs:74,0.116207851817186)
--(axis cs:73,0.15280097641348)
--(axis cs:72,0.131115534483753)
--(axis cs:71,0.13455990520962)
--(axis cs:70,0.146173139716143)
--(axis cs:69,0.125370627989683)
--(axis cs:68,0.113403052751439)
--(axis cs:67,0.113083618220388)
--(axis cs:66,0.127781270910203)
--(axis cs:65,0.105240939861894)
--(axis cs:64,0.107220244817188)
--(axis cs:63,0.0982832786862131)
--(axis cs:62,0.0885664384901064)
--(axis cs:61,0.0961236408275743)
--(axis cs:60,0.0865136432658002)
--(axis cs:59,0.0821063018967238)
--(axis cs:58,0.0823777928745736)
--(axis cs:57,0.103445471672548)
--(axis cs:56,0.0787059399860414)
--(axis cs:55,0.0732385763540111)
--(axis cs:54,0.0848488404389437)
--(axis cs:53,0.0633019794430932)
--(axis cs:52,0.0779379633206372)
--(axis cs:51,0.0759640295658967)
--(axis cs:50,0.0746905147147749)
--(axis cs:49,0.0489193446820647)
--(axis cs:48,0.0858653679217999)
--(axis cs:47,0.0798452851155351)
--(axis cs:46,0.0568004226337559)
--(axis cs:45,0.0629804267031668)
--(axis cs:44,0.0606493879315351)
--(axis cs:43,0.0564596361689432)
--(axis cs:42,0.073155097430368)
--(axis cs:41,0.0681233069778608)
--(axis cs:40,0.0725092942459742)
--cycle;

\addplot [line0]
table {%
40 0.0877979508270029
41 0.0801128739756465
42 0.0910951396845889
43 0.0742532282661341
44 0.0766522786181023
45 0.0748583026764195
46 0.0644482019081344
47 0.0884964241304968
48 0.0942639156176009
49 0.0573946131962181
50 0.0850109175222294
51 0.084385642261302
52 0.0823004611243824
53 0.0665791531859866
54 0.0914135861485044
55 0.0752031724926124
56 0.0853484922956392
57 0.102926375294629
58 0.0863550334088918
59 0.0850798648852331
60 0.0896766990692028
61 0.0959151956843445
62 0.0914056046849582
63 0.098152850057789
64 0.108424547374306
65 0.103856640693559
66 0.125877372649758
67 0.113069420522469
68 0.107183144774152
69 0.115714939186593
70 0.140741914090041
71 0.13212225686127
72 0.127188503382688
73 0.142074004657266
74 0.110313209147715
75 0.129628733077537
76 0.157985667223775
77 0.152615872186373
78 0.163308587494367
79 0.15486135187152
80 0.16439236970259
81 0.150469838333617
82 0.139294897787704
83 0.165499579395098
84 0.16052176661152
85 0.169856276671349
86 0.176776255314055
87 0.176247420836891
88 0.195517551094629
89 0.201729327836162
90 0.199092369326257
91 0.196821447568082
92 0.215525691591398
93 0.202327830533363
94 0.226793638992061
95 0.232450954991793
96 0.189309419821702
97 0.21694862783656
98 0.237568249130773
99 0.231507092971424
100 0.240329238073575
} coordinate[pos=1] (lstm);
\draw[<-,shorten <=2pt] (gref) to +(6pt,4pt) node[linelabel,right] {LSTM};
\addlegendentry{LSTM}
\addplot [line1]
table {%
40 0.0805860855926279
41 0.0737383146197014
42 0.0798600845785621
43 0.0564726766218753
44 0.0503719940833012
45 0.0538909958187807
46 0.0420837371068437
47 0.0610259183957893
48 0.0607422532233952
49 0.0206610005128635
50 0.0462418940847293
51 0.048138878321351
52 0.041193424240368
53 0.0249206782818003
54 0.046399475724604
55 0.033081390709942
56 0.0374728165492665
57 0.0481789023642561
58 0.0315776728216073
59 0.0327118953076378
60 0.0394016909311819
61 0.0468241352335249
62 0.036014215761067
63 0.0372445539268367
64 0.0518883687854388
65 0.0507081506094239
66 0.0634229159736221
67 0.047696912943178
68 0.0475977817226815
69 0.0616634856653063
70 0.0746249846815586
71 0.0665194083247382
72 0.0646106835584691
73 0.0772959585849199
74 0.0590053346227993
75 0.0718633684942034
76 0.0949251460477886
77 0.0898083501185969
78 0.10901841697153
79 0.0923783675459666
80 0.106537810376418
81 0.0890977499366262
82 0.08538194880714
83 0.107285982821302
84 0.107768063812041
85 0.114937693261423
86 0.120977739461947
87 0.121157981181718
88 0.143499345972115
89 0.156060497736092
90 0.152142211991188
91 0.157214036424538
92 0.162178475856004
93 0.16126117751757
94 0.171137321463803
95 0.18564665297699
96 0.14318455206454
97 0.17631108353888
98 0.196509306608452
99 0.19484517101751
100 0.195774628698575
} coordinate[pos=1] (gref);
\draw[<-,shorten <=2pt] (gref) to +(6pt,-3pt) node[linelabel,right] {Gref};
\addplot [line2]
table {%
40 0.0761669571746592
41 0.0682173006620033
42 0.0722566793981305
43 0.0488339763584306
44 0.0412044470307443
45 0.0463645689524612
46 0.036080791497129
47 0.0552973248535686
48 0.0582477565599838
49 0.0137137996598661
50 0.0423133247487918
51 0.044477013086976
52 0.0346404316397069
53 0.023031895852555
54 0.0368253654815484
55 0.0283388170238624
56 0.030496633339166
57 0.0424605892474798
58 0.0239348210907345
59 0.0280220532128284
60 0.0303465273569632
61 0.0400619201976642
62 0.0310182952721557
63 0.0280406073742573
64 0.0450836583972552
65 0.0409759841430778
66 0.0561077822141524
67 0.0393073082323571
68 0.0398029962792073
69 0.0471055146508136
70 0.0657983105744158
71 0.0545861790663755
72 0.0488791839924969
73 0.0667154523777281
74 0.0450145921713635
75 0.0601418841192034
76 0.0779894192796965
77 0.0773792587022008
78 0.0910321514466903
79 0.0806536840016628
80 0.0903222525150895
81 0.0811541735477373
82 0.0706604477591218
83 0.094836534637944
84 0.0992463062320561
85 0.108541622489364
86 0.116420156340581
87 0.112623947417351
88 0.137049533516504
89 0.149425452309477
90 0.145785565940841
91 0.157101688811213
92 0.161748453989496
93 0.158115328433228
94 0.170427365720784
95 0.186730179909556
96 0.143839861105881
97 0.176126140723158
98 0.200869010450926
99 0.199704891266879
100 0.202748281042325
} coordinate[pos=1] (jm);
\draw[<-,shorten <=2pt] (jm) to +(6pt,2pt) node[linelabel,right] {JM};
\addplot [line3]
table {%
40 0.0670023148406748
41 0.0606321313117746
42 0.0665366680049013
43 0.0473526616327765
44 0.0456791840247074
45 0.0520702818430862
46 0.0430010742756616
47 0.0662745054968532
48 0.0707672471198796
49 0.031341091592264
50 0.0575645747487918
51 0.0588262442833608
52 0.0573459182307525
53 0.0418333081830386
54 0.0632654425214789
55 0.0518698317610783
56 0.0570849642892219
57 0.0746289349163395
58 0.0576610047518336
59 0.0572271230618751
60 0.0633849103322236
61 0.0722735380895699
62 0.0640230835786073
63 0.0703695849288208
64 0.0820490056872943
65 0.0787892879291355
66 0.0995054650976751
67 0.0818213226330101
68 0.0809753739216705
69 0.0932634276376795
70 0.114546469056559
71 0.102371397211182
72 0.0984697126925837
73 0.119553958370879
74 0.0859444908200122
75 0.10709952734837
76 0.127182341771473
77 0.123990758042704
78 0.144048746749174
79 0.130632848608071
80 0.139114848950636
81 0.13004933172675
82 0.119116294033893
83 0.142753615069043
84 0.14367542988347
85 0.150116381220982
86 0.160372895257732
87 0.158263455544147
88 0.182315003154289
89 0.185270689098452
90 0.185002623232507
91 0.1901629224994
92 0.20048018590186
93 0.1915460029964
94 0.214226453569854
95 0.22494245375824
96 0.183585237692795
97 0.211662454753441
98 0.23053093049238
99 0.223137508454379
100 0.233939609167325
} coordinate[pos=1] (ours);
\draw[<-,shorten <=2pt] (ours) to +(6pt,-4pt) node[linelabel,right] {Ours};
\legend{} 
\end{axis}

\end{tikzpicture}

%% file: figures/test-dyck.tex
\begin{tikzpicture}

\begin{axis}[
title={Dyck},
ylabel={Difference in Cross Entropy},
ymin=-0.02,
ymax=0.12,
]
\path [draw=color0, fill=color0, bars]
(axis cs:40,0.00397435457325653)
--(axis cs:40,-0.0018207756612623)
--(axis cs:42,-0.0076937538589557)
--(axis cs:44,-0.0127886337303192)
--(axis cs:46,-0.0161481346410777)
--(axis cs:48,-0.0164528043966118)
--(axis cs:50,-0.0147326763235661)
--(axis cs:52,-0.0136045899485064)
--(axis cs:54,-0.0145494239396404)
--(axis cs:56,-0.0100510404350069)
--(axis cs:58,-0.0100076372773604)
--(axis cs:60,-0.00988139830351791)
--(axis cs:62,-0.00418288574183191)
--(axis cs:64,0.00203121168041852)
--(axis cs:66,-0.000847079656041668)
--(axis cs:68,0.00190121930370493)
--(axis cs:70,0.00993220718830984)
--(axis cs:72,0.0132089864281447)
--(axis cs:74,0.0138719119687483)
--(axis cs:76,0.018320071886356)
--(axis cs:78,0.0228945518926805)
--(axis cs:80,0.0200687256779975)
--(axis cs:82,0.0259733753481284)
--(axis cs:84,0.0384534015764209)
--(axis cs:86,0.0413594105789218)
--(axis cs:88,0.0506751317012664)
--(axis cs:90,0.0506512339062814)
--(axis cs:92,0.0609880526801557)
--(axis cs:94,0.0768359561303223)
--(axis cs:96,0.0771798420990601)
--(axis cs:98,0.0668813464291176)
--(axis cs:100,0.0877214132540295)
--(axis cs:100,0.109072868500684)
--(axis cs:100,0.109072868500684)
--(axis cs:98,0.0893016981049654)
--(axis cs:96,0.0951347763842185)
--(axis cs:94,0.100337567576246)
--(axis cs:92,0.0803032922156338)
--(axis cs:90,0.0735850274618142)
--(axis cs:88,0.0664746380237882)
--(axis cs:86,0.0675287525854638)
--(axis cs:84,0.0545143693105924)
--(axis cs:82,0.047903616256777)
--(axis cs:80,0.0391361238204906)
--(axis cs:78,0.0441139469417605)
--(axis cs:76,0.0400040249086916)
--(axis cs:74,0.0318763068223729)
--(axis cs:72,0.0335509021064202)
--(axis cs:70,0.0283461919206483)
--(axis cs:68,0.0175162723536913)
--(axis cs:66,0.0151993775044417)
--(axis cs:64,0.0161228341867879)
--(axis cs:62,0.0100328238744378)
--(axis cs:60,0.000393826289150102)
--(axis cs:58,0.00357845242560461)
--(axis cs:56,0.000174427600167019)
--(axis cs:54,-0.00437738751205838)
--(axis cs:52,-0.00366339023878588)
--(axis cs:50,-0.00249206225791725)
--(axis cs:48,-0.00543468673365508)
--(axis cs:46,-0.00275956212975761)
--(axis cs:44,-0.00180220597420127)
--(axis cs:42,0.000429170377621888)
--(axis cs:40,0.00397435457325653)
--cycle;

\path [draw=color1, fill=color1, bars]
(axis cs:40,0.0134687146360565)
--(axis cs:40,0.00601337990093773)
--(axis cs:42,-0.000419198364600862)
--(axis cs:44,-0.00464820037938045)
--(axis cs:46,-0.00949409140329263)
--(axis cs:48,-0.0115684770626649)
--(axis cs:50,-0.0141324472522716)
--(axis cs:52,-0.0153596816690244)
--(axis cs:54,-0.0171431300667384)
--(axis cs:56,-0.0158526806901884)
--(axis cs:58,-0.017473422110998)
--(axis cs:60,-0.0177318764600542)
--(axis cs:62,-0.0164518449421501)
--(axis cs:64,-0.0146945612024745)
--(axis cs:66,-0.013562496633242)
--(axis cs:68,-0.0150918465726973)
--(axis cs:70,-0.0121155309646416)
--(axis cs:72,-0.010040497553915)
--(axis cs:74,-0.00826725757777704)
--(axis cs:76,-0.00433927328046305)
--(axis cs:78,-0.00173633390405762)
--(axis cs:80,0.000417989006801754)
--(axis cs:82,0.00208059444953874)
--(axis cs:84,0.00525602840907601)
--(axis cs:86,0.0114782738333444)
--(axis cs:88,0.0134597494033058)
--(axis cs:90,0.0179566426500777)
--(axis cs:92,0.0216493198613006)
--(axis cs:94,0.0230576608619413)
--(axis cs:96,0.0294636803512422)
--(axis cs:98,0.0292275620913601)
--(axis cs:100,0.0325480545122617)
--(axis cs:100,0.0466848600549522)
--(axis cs:100,0.0466848600549522)
--(axis cs:98,0.0422679425830289)
--(axis cs:96,0.0416636822726614)
--(axis cs:94,0.0337330503446273)
--(axis cs:92,0.0325908828809562)
--(axis cs:90,0.0264425744471846)
--(axis cs:88,0.0221171776370896)
--(axis cs:86,0.0190827863150527)
--(axis cs:84,0.0124994796431158)
--(axis cs:82,0.00775926396176918)
--(axis cs:80,0.00562067396824891)
--(axis cs:78,0.00333049479779356)
--(axis cs:76,-0.000899258419555261)
--(axis cs:74,-0.00486643874934503)
--(axis cs:72,-0.0070453441632563)
--(axis cs:70,-0.0101366223817573)
--(axis cs:68,-0.0132765073581418)
--(axis cs:66,-0.012207423794873)
--(axis cs:64,-0.0129645643170378)
--(axis cs:62,-0.0145560168244375)
--(axis cs:60,-0.0154360015438971)
--(axis cs:58,-0.0154737617978698)
--(axis cs:56,-0.0130301751692051)
--(axis cs:54,-0.0147525941453771)
--(axis cs:52,-0.0121388987286045)
--(axis cs:50,-0.00994018195421175)
--(axis cs:48,-0.00693725951031025)
--(axis cs:46,-0.00439201484444497)
--(axis cs:44,0.00118415843764419)
--(axis cs:42,0.00520500085796941)
--(axis cs:40,0.0134687146360565)
--cycle;

\path [draw=color2, fill=color2, bars]
(axis cs:40,0.0122389907395511)
--(axis cs:40,0.00901278153181818)
--(axis cs:42,0.00263735907887086)
--(axis cs:44,-0.00198155933081797)
--(axis cs:46,-0.00684419645674724)
--(axis cs:48,-0.00993699199128213)
--(axis cs:50,-0.0111671365086461)
--(axis cs:52,-0.0125817398615038)
--(axis cs:54,-0.0155703631520077)
--(axis cs:56,-0.0144120753687733)
--(axis cs:58,-0.0170728204963647)
--(axis cs:60,-0.0171245428070105)
--(axis cs:62,-0.0165650767967294)
--(axis cs:64,-0.0147298588792239)
--(axis cs:66,-0.0136369890884257)
--(axis cs:68,-0.0157706333515687)
--(axis cs:70,-0.0126427471839025)
--(axis cs:72,-0.0106425489497968)
--(axis cs:74,-0.00933388621243551)
--(axis cs:76,-0.00587328000491545)
--(axis cs:78,-0.00294736898470888)
--(axis cs:80,-0.00206291425915856)
--(axis cs:82,-0.000112054685588073)
--(axis cs:84,0.00367121121393739)
--(axis cs:86,0.00901721685772161)
--(axis cs:88,0.0116990076930362)
--(axis cs:90,0.015769938303987)
--(axis cs:92,0.0203872047595101)
--(axis cs:94,0.0213905895143245)
--(axis cs:96,0.0274724256530911)
--(axis cs:98,0.0273555090574962)
--(axis cs:100,0.0314792560290724)
--(axis cs:100,0.0416183460381416)
--(axis cs:100,0.0416183460381416)
--(axis cs:98,0.0370577252087295)
--(axis cs:96,0.0370728650307083)
--(axis cs:94,0.0291428078624567)
--(axis cs:92,0.027575059803399)
--(axis cs:90,0.023682664209942)
--(axis cs:88,0.0185591160803137)
--(axis cs:86,0.0158871163648036)
--(axis cs:84,0.00873947726980204)
--(axis cs:82,0.0053228877539082)
--(axis cs:80,0.0031232325076469)
--(axis cs:78,0.000504220182931917)
--(axis cs:76,-0.00298410963095814)
--(axis cs:74,-0.00675372692211891)
--(axis cs:72,-0.00830055188195785)
--(axis cs:70,-0.0111794117428537)
--(axis cs:68,-0.0147089912594175)
--(axis cs:66,-0.0125354822393105)
--(axis cs:64,-0.0132044131246633)
--(axis cs:62,-0.0144597960585679)
--(axis cs:60,-0.014600301342774)
--(axis cs:58,-0.0142553574857791)
--(axis cs:56,-0.0113659046535666)
--(axis cs:54,-0.0130677836584874)
--(axis cs:52,-0.00995190288588481)
--(axis cs:50,-0.00886393019783711)
--(axis cs:48,-0.00603796251789105)
--(axis cs:46,-0.00259821074207723)
--(axis cs:44,0.0012561981419226)
--(axis cs:42,0.00503560394277143)
--(axis cs:40,0.0122389907395511)
--cycle;

\path [draw=color3, fill=color3, bars]
(axis cs:40,0.00915211050864473)
--(axis cs:40,0.004239334614287)
--(axis cs:42,-0.00468814041089317)
--(axis cs:44,-0.00790152480947355)
--(axis cs:46,-0.0120021856448589)
--(axis cs:48,-0.0134653493469781)
--(axis cs:50,-0.0139095637033049)
--(axis cs:52,-0.0140473561896638)
--(axis cs:54,-0.0165461016088443)
--(axis cs:56,-0.0137834666193513)
--(axis cs:58,-0.016309273171637)
--(axis cs:60,-0.0159510344149431)
--(axis cs:62,-0.0158076687041489)
--(axis cs:64,-0.013921280379241)
--(axis cs:66,-0.0134302790770526)
--(axis cs:68,-0.014959389741883)
--(axis cs:70,-0.0122926303763315)
--(axis cs:72,-0.0104673435982308)
--(axis cs:74,-0.0089669511659228)
--(axis cs:76,-0.00537916060674653)
--(axis cs:78,-0.00280044052962489)
--(axis cs:80,-0.000746712727080017)
--(axis cs:82,0.00101603112057103)
--(axis cs:84,0.00561015280876592)
--(axis cs:86,0.0104847432327164)
--(axis cs:88,0.0140867923249296)
--(axis cs:90,0.018237925710533)
--(axis cs:92,0.0234197638686928)
--(axis cs:94,0.0245607021004629)
--(axis cs:96,0.0322125201290493)
--(axis cs:98,0.0333006534219282)
--(axis cs:100,0.0373451950559627)
--(axis cs:100,0.0572669968550011)
--(axis cs:100,0.0572669968550011)
--(axis cs:98,0.0490800752639405)
--(axis cs:96,0.047037095912823)
--(axis cs:94,0.0367681989332333)
--(axis cs:92,0.0343388661833468)
--(axis cs:90,0.0270645465950624)
--(axis cs:88,0.0222555155748408)
--(axis cs:86,0.01715623606539)
--(axis cs:84,0.00968378995324729)
--(axis cs:82,0.00566936368555402)
--(axis cs:80,0.00272714816306813)
--(axis cs:78,0.000646704788745305)
--(axis cs:76,-0.00233225863439028)
--(axis cs:74,-0.00629512353957754)
--(axis cs:72,-0.00791339286720447)
--(axis cs:70,-0.0100598298897102)
--(axis cs:68,-0.0142510770106474)
--(axis cs:66,-0.0111650734059869)
--(axis cs:64,-0.012096182542615)
--(axis cs:62,-0.0137382789999388)
--(axis cs:60,-0.0137950662452581)
--(axis cs:58,-0.0146849372727912)
--(axis cs:56,-0.0116237509866939)
--(axis cs:54,-0.0143419547791971)
--(axis cs:52,-0.0117735837332055)
--(axis cs:50,-0.0104477334719284)
--(axis cs:48,-0.00842570216740337)
--(axis cs:46,-0.00645168554853076)
--(axis cs:44,-0.00148017515783078)
--(axis cs:42,0.00160331418402354)
--(axis cs:40,0.00915211050864473)
--cycle;

\addplot [line0]
table {%
40 0.00107678945599712
42 -0.0036322917406669
44 -0.00729541985226023
46 -0.00945384838541765
48 -0.0109437455651334
50 -0.00861236929074167
52 -0.00863399009364616
54 -0.00946340572584938
56 -0.00493830641741992
58 -0.00321459242587792
60 -0.0047437860071839
62 0.00292496906630295
64 0.0090770229336032
66 0.00717614892420002
68 0.0097087458286981
70 0.0191391995544791
72 0.0233799442672824
74 0.0228741093955606
76 0.0291620483975238
78 0.0335042494172205
80 0.0296024247492441
82 0.0369384958024527
84 0.0464838854435066
86 0.0544440815821928
88 0.0585748848625273
90 0.0621181306840478
92 0.0706456724478947
94 0.0885867618532842
96 0.0861573092416393
98 0.0780915222670415
100 0.0983971408773569
} node[linelabel,right,pos=1] {LSTM};
\addlegendentry{LSTM}
\addplot [line1]
table {%
40 0.00974104726849712
42 0.00239290124668428
44 -0.00173202097086813
46 -0.0069430531238688
48 -0.00925286828648757
50 -0.0120363146032417
52 -0.0137492901988144
54 -0.0159478621060578
56 -0.0144414279296968
58 -0.0164735919544339
60 -0.0165839390019756
62 -0.0155039308832938
64 -0.0138295627597562
66 -0.0128849602140575
68 -0.0141841769654196
70 -0.0111260766731994
72 -0.00854292085858563
74 -0.00656684816356103
76 -0.00261926585000916
78 0.00079708044686797
80 0.00301933148752533
82 0.00491992920565396
84 0.0088777540260959
86 0.0152805300741985
88 0.0177884635201977
90 0.0221996085486311
92 0.0271201013711284
94 0.0283953556032843
96 0.0355636813119518
98 0.0357477523371945
100 0.039616457283607
} coordinate[pos=1] (gref);
\draw[<-,shorten <=2pt] (gref) to +(6pt,0pt) node[linelabel,right] {Gref};
\addlegendentry{Gref}
\addplot [line2]
table {%
40 0.0106258861356846
42 0.00383648151082114
44 -0.000362680594447684
46 -0.00472120359941224
48 -0.00798747725458659
50 -0.0100155333532416
52 -0.0112668213736943
54 -0.0143190734052475
56 -0.01288899001117
58 -0.0156640889910719
60 -0.0158624220748922
62 -0.0155124364276487
64 -0.0139671360019436
66 -0.0130862356638681
68 -0.0152398123054931
70 -0.0119110794633781
72 -0.00947155041587733
74 -0.00804380656727721
76 -0.0044286948179368
78 -0.00122157440088848
80 0.000530159124244167
82 0.00260541653416007
84 0.00620534424186972
86 0.0124521666112626
88 0.015129061886675
90 0.0197263012569645
92 0.0239811322814545
94 0.0252666986883906
96 0.0322726453418997
98 0.0322066171331129
100 0.036548801033607
} coordinate[pos=1] (jm);
\draw[<-,shorten <=2pt] (jm) to +(6pt,-6pt) node[linelabel,right] {JM};
\addlegendentry{JM}
\addplot [line3]
table {%
40 0.00669572256146587
42 -0.00154241311343482
44 -0.00469084998365217
46 -0.00922693559669485
48 -0.0109455257571907
50 -0.0121786485876166
52 -0.0129104699614346
54 -0.0154440281940207
56 -0.0127036088030226
58 -0.0154971052222141
60 -0.0148730503301006
62 -0.0147729738520439
64 -0.013008731460928
66 -0.0122976762415197
68 -0.0146052333762652
70 -0.0111762301330209
72 -0.00919036823271762
74 -0.00763103735275017
76 -0.0038557096205684
78 -0.00107686787043979
80 0.000990217717994057
82 0.00334269740306252
84 0.00764697138100661
86 0.0138204896490532
88 0.0181711539498852
90 0.0226512361527977
92 0.0288793150260198
94 0.0306644505168481
96 0.0396248080209362
98 0.0411903643429343
100 0.0473060959554819
} coordinate[pos=1] (ours);
\draw[<-,shorten <=2pt] (ours) to +(6pt,+4pt) node[linelabel,right] {Ours};
\addlegendentry{Ours}
\legend{} 
\end{axis}

\end{tikzpicture}

%% file: figures/test-hardest-cfl.tex
\begin{tikzpicture}

\begin{axis}[
title={Hardest CFL},
xlabel={Length},
ylabel={Difference in Cross Entropy},
ymin=0,ymax=0.07,ytick distance=0.05,       ]
\path [draw=color0, fill=color0, bars]
(axis cs:40,0.0312083422807144)
--(axis cs:40,0.0227982845173131)
--(axis cs:41,0.0140771364442471)
--(axis cs:42,0.023592489459965)
--(axis cs:43,0.0155282199395767)
--(axis cs:44,0.0197272244417683)
--(axis cs:45,0.0215180371801665)
--(axis cs:46,0.0187213158799968)
--(axis cs:47,0.0221074743526268)
--(axis cs:48,0.0193414766612952)
--(axis cs:49,0.019035939565811)
--(axis cs:50,0.0157058067704478)
--(axis cs:51,0.0192995516747909)
--(axis cs:52,0.0202341164475398)
--(axis cs:53,0.0179485080986109)
--(axis cs:54,0.018397222746021)
--(axis cs:55,0.0178009573044113)
--(axis cs:56,0.0236904700032682)
--(axis cs:57,0.0172180633228192)
--(axis cs:58,0.022737611206227)
--(axis cs:59,0.0189683668762277)
--(axis cs:60,0.0227839637723913)
--(axis cs:61,0.0277730226102553)
--(axis cs:62,0.0218488780989876)
--(axis cs:63,0.0240580496001254)
--(axis cs:64,0.0262353817296755)
--(axis cs:65,0.0245991566058443)
--(axis cs:66,0.0286893942375568)
--(axis cs:67,0.0211024950686153)
--(axis cs:68,0.0331147553661004)
--(axis cs:69,0.0295848374478537)
--(axis cs:70,0.0317986752728043)
--(axis cs:71,0.034112016982277)
--(axis cs:72,0.0301383642316965)
--(axis cs:73,0.0307365916078531)
--(axis cs:74,0.0329805953793256)
--(axis cs:75,0.0411044650988959)
--(axis cs:76,0.0391695630008519)
--(axis cs:77,0.0398667112892108)
--(axis cs:78,0.0361354675693368)
--(axis cs:79,0.0345719926689623)
--(axis cs:80,0.0418565886557804)
--(axis cs:81,0.0411689495762991)
--(axis cs:82,0.0364961819220056)
--(axis cs:83,0.0406428205429784)
--(axis cs:84,0.0395373156684554)
--(axis cs:85,0.0426797149431217)
--(axis cs:86,0.0458633653742985)
--(axis cs:87,0.0443410760845772)
--(axis cs:88,0.0477297827621797)
--(axis cs:89,0.0475922273529072)
--(axis cs:90,0.0461618709446303)
--(axis cs:91,0.0501269063808383)
--(axis cs:92,0.0496039116847296)
--(axis cs:93,0.0533748806993265)
--(axis cs:94,0.0505520747391012)
--(axis cs:95,0.0542894128026079)
--(axis cs:96,0.0500928016902358)
--(axis cs:97,0.0519412554621561)
--(axis cs:98,0.0528374107287711)
--(axis cs:99,0.0534757067371086)
--(axis cs:100,0.0573993909092524)
--(axis cs:100,0.0643871896460444)
--(axis cs:100,0.0643871896460444)
--(axis cs:99,0.0598679915569015)
--(axis cs:98,0.0586555848219616)
--(axis cs:97,0.0556905272820667)
--(axis cs:96,0.0551008455291883)
--(axis cs:95,0.0602314400067769)
--(axis cs:94,0.0565027618150242)
--(axis cs:93,0.0591330201661494)
--(axis cs:92,0.055293169610107)
--(axis cs:91,0.0550915152888002)
--(axis cs:90,0.0515261978980201)
--(axis cs:89,0.0550260483217272)
--(axis cs:88,0.0522576808310449)
--(axis cs:87,0.0479262896005987)
--(axis cs:86,0.0494112002601222)
--(axis cs:85,0.0474979728408349)
--(axis cs:84,0.0450962779948037)
--(axis cs:83,0.0440609334700357)
--(axis cs:82,0.041301773788034)
--(axis cs:81,0.045730426977909)
--(axis cs:80,0.0464061703701945)
--(axis cs:79,0.0382673326774206)
--(axis cs:78,0.0405718703884957)
--(axis cs:77,0.0423517630338665)
--(axis cs:76,0.0432932309745197)
--(axis cs:75,0.0445553133876034)
--(axis cs:74,0.0397390073478209)
--(axis cs:73,0.0331316127784819)
--(axis cs:72,0.0397518223767252)
--(axis cs:71,0.0366623846589042)
--(axis cs:70,0.0339163499713538)
--(axis cs:69,0.0340437397241369)
--(axis cs:68,0.0383971951789337)
--(axis cs:67,0.0271624795910003)
--(axis cs:66,0.03260487542251)
--(axis cs:65,0.0292002316576012)
--(axis cs:64,0.0328868424939367)
--(axis cs:63,0.0296164432111803)
--(axis cs:62,0.0280615834439049)
--(axis cs:61,0.0360273202714862)
--(axis cs:60,0.0314471967978726)
--(axis cs:59,0.0237925608481651)
--(axis cs:58,0.0305063003011626)
--(axis cs:57,0.0222797029246334)
--(axis cs:56,0.0302208597515649)
--(axis cs:55,0.0257239818167211)
--(axis cs:54,0.0244648097908872)
--(axis cs:53,0.0287251468899925)
--(axis cs:52,0.0291479671662089)
--(axis cs:51,0.0229844272306283)
--(axis cs:50,0.0243309448786321)
--(axis cs:49,0.0252777539280753)
--(axis cs:48,0.0260884241973513)
--(axis cs:47,0.0303705334020162)
--(axis cs:46,0.026957642688638)
--(axis cs:45,0.0291502938891635)
--(axis cs:44,0.0288802167270955)
--(axis cs:43,0.0236897319171943)
--(axis cs:42,0.0313891560522103)
--(axis cs:41,0.0211451249915235)
--(axis cs:40,0.0312083422807144)
--cycle;

\path [draw=color1, fill=color1, bars]
(axis cs:40,0.0239942234886119)
--(axis cs:40,0.0168720712781657)
--(axis cs:41,0.0110759805859209)
--(axis cs:42,0.0188791175051311)
--(axis cs:43,0.0130287839582167)
--(axis cs:44,0.0183606143237481)
--(axis cs:45,0.018842392124823)
--(axis cs:46,0.0166358758953143)
--(axis cs:47,0.0164147845913987)
--(axis cs:48,0.0137627666291408)
--(axis cs:49,0.0146919526123218)
--(axis cs:50,0.012668531871296)
--(axis cs:51,0.0159102305343128)
--(axis cs:52,0.0177739252769992)
--(axis cs:53,0.0127668038447523)
--(axis cs:54,0.0114104256240672)
--(axis cs:55,0.0147435043268505)
--(axis cs:56,0.0207511648156289)
--(axis cs:57,0.0141631045861952)
--(axis cs:58,0.0193419870676727)
--(axis cs:59,0.0150693579391797)
--(axis cs:60,0.0204129230415081)
--(axis cs:61,0.0259087635291725)
--(axis cs:62,0.0164057955845395)
--(axis cs:63,0.0186181373516649)
--(axis cs:64,0.0227085108013025)
--(axis cs:65,0.0214046355090824)
--(axis cs:66,0.0222321374933356)
--(axis cs:67,0.0183525769394238)
--(axis cs:68,0.0253799034320128)
--(axis cs:69,0.0239797617317199)
--(axis cs:70,0.0273455382682737)
--(axis cs:71,0.028756986130603)
--(axis cs:72,0.0262050093972349)
--(axis cs:73,0.0229504892947294)
--(axis cs:74,0.0307022033507314)
--(axis cs:75,0.0325993534246331)
--(axis cs:76,0.0325500483449624)
--(axis cs:77,0.0307879255861017)
--(axis cs:78,0.0301962687972591)
--(axis cs:79,0.0301307329886347)
--(axis cs:80,0.0354342463403947)
--(axis cs:81,0.0372830859559425)
--(axis cs:82,0.0293008116613279)
--(axis cs:83,0.0385310215498588)
--(axis cs:84,0.0375931036584464)
--(axis cs:85,0.0390947895835198)
--(axis cs:86,0.0412606394738903)
--(axis cs:87,0.0430915555511168)
--(axis cs:88,0.0434613198063121)
--(axis cs:89,0.0441041798240289)
--(axis cs:90,0.0425320570736619)
--(axis cs:91,0.0450872413400661)
--(axis cs:92,0.0470079150121993)
--(axis cs:93,0.0507073346462229)
--(axis cs:94,0.04650561355906)
--(axis cs:95,0.0542237614644957)
--(axis cs:96,0.0470661373664319)
--(axis cs:97,0.0463016209031141)
--(axis cs:98,0.0511800670675399)
--(axis cs:99,0.0494772401553373)
--(axis cs:100,0.0565891586299999)
--(axis cs:100,0.0614051172377968)
--(axis cs:100,0.0614051172377968)
--(axis cs:99,0.0544270247422083)
--(axis cs:98,0.0565230687893152)
--(axis cs:97,0.0526459640318303)
--(axis cs:96,0.0539955111550755)
--(axis cs:95,0.0567822064764678)
--(axis cs:94,0.0518811711333634)
--(axis cs:93,0.0559303125230164)
--(axis cs:92,0.0506071095570938)
--(axis cs:91,0.0521079145328692)
--(axis cs:90,0.0476415586439884)
--(axis cs:89,0.0478023244320663)
--(axis cs:88,0.0481908277357762)
--(axis cs:87,0.0455981790277373)
--(axis cs:86,0.0461982012332049)
--(axis cs:85,0.0445601500386721)
--(axis cs:84,0.042553278323265)
--(axis cs:83,0.0409102607011071)
--(axis cs:82,0.0361735445822484)
--(axis cs:81,0.0432800637464138)
--(axis cs:80,0.0409415009668301)
--(axis cs:79,0.0358512940982544)
--(axis cs:78,0.0349702538560862)
--(axis cs:77,0.0386089172434692)
--(axis cs:76,0.0410764997422514)
--(axis cs:75,0.0389181854785329)
--(axis cs:74,0.0349906362851989)
--(axis cs:73,0.0293438709135234)
--(axis cs:72,0.031285057853548)
--(axis cs:71,0.0330542222887471)
--(axis cs:70,0.0312279133151701)
--(axis cs:69,0.0315715396250533)
--(axis cs:68,0.0357017277196388)
--(axis cs:67,0.0236968543433262)
--(axis cs:66,0.0269812846288525)
--(axis cs:65,0.0247641638120555)
--(axis cs:64,0.0291295337348098)
--(axis cs:63,0.024134542463609)
--(axis cs:62,0.0218231513414175)
--(axis cs:61,0.0327028626517495)
--(axis cs:60,0.0267629640912558)
--(axis cs:59,0.0222911460563995)
--(axis cs:58,0.0240166200216135)
--(axis cs:57,0.0193132800823101)
--(axis cs:56,0.0263323189570612)
--(axis cs:55,0.0219709589420091)
--(axis cs:54,0.0201318007785818)
--(axis cs:53,0.0199572638797001)
--(axis cs:52,0.0214385369425187)
--(axis cs:51,0.0220774095230674)
--(axis cs:50,0.0205252510277842)
--(axis cs:49,0.022642786799932)
--(axis cs:48,0.0232493282399222)
--(axis cs:47,0.0252236703042018)
--(axis cs:46,0.0248679383119075)
--(axis cs:45,0.0263067549167293)
--(axis cs:44,0.0246755377826156)
--(axis cs:43,0.021215330689252)
--(axis cs:42,0.0284629260725204)
--(axis cs:41,0.0207430796303377)
--(axis cs:40,0.0239942234886119)
--cycle;

\path [draw=color2, fill=color2, bars]
(axis cs:40,0.0256673724228297)
--(axis cs:40,0.0102313442189477)
--(axis cs:41,0.0113965372414018)
--(axis cs:42,0.0162301341115933)
--(axis cs:43,0.011329398619354)
--(axis cs:44,0.0183934423143628)
--(axis cs:45,0.0161723581555756)
--(axis cs:46,0.012255338541766)
--(axis cs:47,0.016739074987843)
--(axis cs:48,0.0123452526047258)
--(axis cs:49,0.0117357625414463)
--(axis cs:50,0.0136677560724533)
--(axis cs:51,0.0131151049728831)
--(axis cs:52,0.017453337631506)
--(axis cs:53,0.00872521338173836)
--(axis cs:54,0.0115394517247728)
--(axis cs:55,0.0151072050540866)
--(axis cs:56,0.0204551990360321)
--(axis cs:57,0.0115280383183852)
--(axis cs:58,0.0147819177237906)
--(axis cs:59,0.014363051859801)
--(axis cs:60,0.0152932897412865)
--(axis cs:61,0.0207131993158736)
--(axis cs:62,0.0142618782255151)
--(axis cs:63,0.0174747236675624)
--(axis cs:64,0.020289152136196)
--(axis cs:65,0.019194801436119)
--(axis cs:66,0.0206505976565337)
--(axis cs:67,0.0176185237266953)
--(axis cs:68,0.0241546914480832)
--(axis cs:69,0.02262624891396)
--(axis cs:70,0.022403293975775)
--(axis cs:71,0.02754585193563)
--(axis cs:72,0.026030036971554)
--(axis cs:73,0.0224355885764889)
--(axis cs:74,0.0272628293463458)
--(axis cs:75,0.034477340357588)
--(axis cs:76,0.0319415683207698)
--(axis cs:77,0.0283471016737049)
--(axis cs:78,0.0304170822253676)
--(axis cs:79,0.0307030482520599)
--(axis cs:80,0.0357891005132732)
--(axis cs:81,0.0364152292646489)
--(axis cs:82,0.0307297039419653)
--(axis cs:83,0.0363121012010316)
--(axis cs:84,0.037426002586768)
--(axis cs:85,0.0374333405121027)
--(axis cs:86,0.0422987359702568)
--(axis cs:87,0.0399271735856695)
--(axis cs:88,0.0395839136566398)
--(axis cs:89,0.0438050945115261)
--(axis cs:90,0.0410306304169744)
--(axis cs:91,0.0442634028695585)
--(axis cs:92,0.0456517848907546)
--(axis cs:93,0.0498438976705303)
--(axis cs:94,0.0458449221240219)
--(axis cs:95,0.0500254771426012)
--(axis cs:96,0.0445295631201424)
--(axis cs:97,0.0438370883934471)
--(axis cs:98,0.0477009347131195)
--(axis cs:99,0.0468065167658535)
--(axis cs:100,0.0507730919766979)
--(axis cs:100,0.0620664963910988)
--(axis cs:100,0.0620664963910988)
--(axis cs:99,0.0545000129675506)
--(axis cs:98,0.0549496262075111)
--(axis cs:97,0.0533399585260333)
--(axis cs:96,0.0550161348805317)
--(axis cs:95,0.0549018312588889)
--(axis cs:94,0.0516449210790398)
--(axis cs:93,0.0540996130739778)
--(axis cs:92,0.0525175875046255)
--(axis cs:91,0.050343926767113)
--(axis cs:90,0.047885954050676)
--(axis cs:89,0.0483911744917602)
--(axis cs:88,0.0479268098797665)
--(axis cs:87,0.0434919521569778)
--(axis cs:86,0.04601757203335)
--(axis cs:85,0.0443853399189129)
--(axis cs:84,0.0442453049901814)
--(axis cs:83,0.0390203237457175)
--(axis cs:82,0.0355889263107572)
--(axis cs:81,0.0423184451290652)
--(axis cs:80,0.0393197522627016)
--(axis cs:79,0.0380071730595128)
--(axis cs:78,0.0343371307324648)
--(axis cs:77,0.0359919083636582)
--(axis cs:76,0.0387747721184177)
--(axis cs:75,0.039515979795578)
--(axis cs:74,0.0354780993943142)
--(axis cs:73,0.0285269544228598)
--(axis cs:72,0.0313597698625621)
--(axis cs:71,0.0341728168710441)
--(axis cs:70,0.0309097223398116)
--(axis cs:69,0.0295134242725233)
--(axis cs:68,0.0347254806043037)
--(axis cs:67,0.0228366911381442)
--(axis cs:66,0.0304051540111089)
--(axis cs:65,0.026689502692711)
--(axis cs:64,0.0278909334155413)
--(axis cs:63,0.0282195484096163)
--(axis cs:62,0.0258870535794742)
--(axis cs:61,0.0352652301437369)
--(axis cs:60,0.0265521937456441)
--(axis cs:59,0.0250304897417104)
--(axis cs:58,0.029086859085323)
--(axis cs:57,0.0202131489816989)
--(axis cs:56,0.0290838511429081)
--(axis cs:55,0.0229458945784093)
--(axis cs:54,0.0210891823746356)
--(axis cs:53,0.0249038513946008)
--(axis cs:52,0.0274834334822427)
--(axis cs:51,0.0223167752805753)
--(axis cs:50,0.0215625893266269)
--(axis cs:49,0.021984499829991)
--(axis cs:48,0.0237854946080873)
--(axis cs:47,0.0273817186710554)
--(axis cs:46,0.0232968791980643)
--(axis cs:45,0.0283533079831987)
--(axis cs:44,0.0282590095079101)
--(axis cs:43,0.019243658615324)
--(axis cs:42,0.0279705404184391)
--(axis cs:41,0.0192781822736373)
--(axis cs:40,0.0256673724228297)
--cycle;

\path [draw=color3, fill=color3, bars]
(axis cs:40,0.0213209943406778)
--(axis cs:40,0.0106438355823497)
--(axis cs:41,0.00753568220080007)
--(axis cs:42,0.0127893237837207)
--(axis cs:43,0.00710587894937156)
--(axis cs:44,0.0143134815597651)
--(axis cs:45,0.00657788437742362)
--(axis cs:46,0.00997352044245157)
--(axis cs:47,0.0105491086585029)
--(axis cs:48,0.00873834661577473)
--(axis cs:49,0.0097853072628466)
--(axis cs:50,0.00788045066999157)
--(axis cs:51,0.0106504542066572)
--(axis cs:52,0.0094479765477507)
--(axis cs:53,0.00577231974581063)
--(axis cs:54,0.00789179380083946)
--(axis cs:55,0.00759414314295863)
--(axis cs:56,0.0124049765739834)
--(axis cs:57,0.010159497350726)
--(axis cs:58,0.0109981591610483)
--(axis cs:59,0.00712353747174191)
--(axis cs:60,0.0130460209372146)
--(axis cs:61,0.0214317904323492)
--(axis cs:62,0.00883894224088467)
--(axis cs:63,0.0143535933762751)
--(axis cs:64,0.0153617111090212)
--(axis cs:65,0.015873768658428)
--(axis cs:66,0.0132167356182866)
--(axis cs:67,0.0133157983398861)
--(axis cs:68,0.0223302130930913)
--(axis cs:69,0.0227480927188236)
--(axis cs:70,0.0214995692894152)
--(axis cs:71,0.0265868617014207)
--(axis cs:72,0.0235461322435073)
--(axis cs:73,0.0200078386531443)
--(axis cs:74,0.0261093274375591)
--(axis cs:75,0.0359336249448912)
--(axis cs:76,0.0313292462699991)
--(axis cs:77,0.0290828552937232)
--(axis cs:78,0.0279331632016877)
--(axis cs:79,0.0280581810533865)
--(axis cs:80,0.0337531903273873)
--(axis cs:81,0.0359828585433731)
--(axis cs:82,0.0287807386049938)
--(axis cs:83,0.0329995566405915)
--(axis cs:84,0.038073183019017)
--(axis cs:85,0.0369045927589973)
--(axis cs:86,0.0431470069278672)
--(axis cs:87,0.0430197893830477)
--(axis cs:88,0.0446214286429529)
--(axis cs:89,0.0428207745397828)
--(axis cs:90,0.0447066456774364)
--(axis cs:91,0.0469956705392647)
--(axis cs:92,0.0497641458496252)
--(axis cs:93,0.0521835773557933)
--(axis cs:94,0.0492857692443927)
--(axis cs:95,0.0533544929651567)
--(axis cs:96,0.049280058742073)
--(axis cs:97,0.0486145560895383)
--(axis cs:98,0.0528204836074641)
--(axis cs:99,0.0531147713575967)
--(axis cs:100,0.0590987779376581)
--(axis cs:100,0.0699415916801387)
--(axis cs:100,0.0699415916801387)
--(axis cs:99,0.0630359029465146)
--(axis cs:98,0.0604725358590848)
--(axis cs:97,0.0569910104820041)
--(axis cs:96,0.0571019429695386)
--(axis cs:95,0.0624916476731755)
--(axis cs:94,0.0574926793443073)
--(axis cs:93,0.0629906121521556)
--(axis cs:92,0.0550701806897767)
--(axis cs:91,0.0568632902787254)
--(axis cs:90,0.0518982634429918)
--(axis cs:89,0.0508384070337282)
--(axis cs:88,0.0516054259303854)
--(axis cs:87,0.0480632641613237)
--(axis cs:86,0.0510529765844606)
--(axis cs:85,0.0443742898779005)
--(axis cs:84,0.0445748730698373)
--(axis cs:83,0.0425037079145913)
--(axis cs:82,0.036425134406875)
--(axis cs:81,0.0431009798163905)
--(axis cs:80,0.0430944319798376)
--(axis cs:79,0.0361187629638825)
--(axis cs:78,0.0351355028811448)
--(axis cs:77,0.0386370902144192)
--(axis cs:76,0.0377902602876094)
--(axis cs:75,0.0397417785416082)
--(axis cs:74,0.0340410185328305)
--(axis cs:73,0.0294540622229167)
--(axis cs:72,0.0318732969864422)
--(axis cs:71,0.0363666772637041)
--(axis cs:70,0.0270662372047429)
--(axis cs:69,0.029776431011138)
--(axis cs:68,0.0335998210916484)
--(axis cs:67,0.0227941506667444)
--(axis cs:66,0.0302456211629923)
--(axis cs:65,0.0250349946050175)
--(axis cs:64,0.0269305570599035)
--(axis cs:63,0.0238610160024909)
--(axis cs:62,0.0233800500479756)
--(axis cs:61,0.0305449740682449)
--(axis cs:60,0.023713134424716)
--(axis cs:59,0.0223096624984134)
--(axis cs:58,0.0242041979282378)
--(axis cs:57,0.0184790309581301)
--(axis cs:56,0.0277243358817425)
--(axis cs:55,0.0200296951259009)
--(axis cs:54,0.0185312919768096)
--(axis cs:53,0.0214467361861889)
--(axis cs:52,0.0246386217895557)
--(axis cs:51,0.0197028414879777)
--(axis cs:50,0.0184616525415884)
--(axis cs:49,0.022465447774407)
--(axis cs:48,0.0219538019642259)
--(axis cs:47,0.0200083538833742)
--(axis cs:46,0.0228137516452049)
--(axis cs:45,0.026588363358573)
--(axis cs:44,0.0222007422795532)
--(axis cs:43,0.0169854831544924)
--(axis cs:42,0.024351361907026)
--(axis cs:41,0.0203377796922877)
--(axis cs:40,0.0213209943406778)
--cycle;

\addplot [line0]
table {%
40 0.0270033133990137
41 0.0176111307178853
42 0.0274908227560876
43 0.0196089759283855
44 0.0243037205844319
45 0.025334165534665
46 0.0228394792843174
47 0.0262390038773215
48 0.0227149504293232
49 0.0221568467469431
50 0.02001837582454
51 0.0211419894527096
52 0.0246910418068744
53 0.0233368274943017
54 0.0214310162684541
55 0.0217624695605662
56 0.0269556648774166
57 0.0197488831237263
58 0.0266219557536948
59 0.0213804638621964
60 0.0271155802851319
61 0.0319001714408708
62 0.0249552307714462
63 0.0268372464056528
64 0.0295611121118061
65 0.0268996941317227
66 0.0306471348300334
67 0.0241324873298078
68 0.035755975272517
69 0.0318142885859953
70 0.0328575126220791
71 0.0353872008205906
72 0.0349450933042108
73 0.0319341021931675
74 0.0363598013635733
75 0.0428298892432496
76 0.0412313969876858
77 0.0411092371615387
78 0.0383536689789162
79 0.0364196626731915
80 0.0441313795129874
81 0.043449688277104
82 0.0388989778550198
83 0.0423518770065071
84 0.0423167968316295
85 0.0450888438919783
86 0.0476372828172104
87 0.046133682842588
88 0.0499937317966123
89 0.0513091378373172
90 0.0488440344213252
91 0.0526092108348192
92 0.0524485406474183
93 0.0562539504327379
94 0.0535274182770627
95 0.0572604264046924
96 0.0525968236097121
97 0.0538158913721114
98 0.0557464977753663
99 0.0566718491470051
100 0.0608932902776484
} coordinate[pos=1] (p);
\draw[<-,shorten <=2pt] (p) -- ++(0.2cm,0.1cm) node[linelabel,right] {LSTM};
\addlegendentry{LSTM}
\addplot [line1]
table {%
40 0.0204331473833888
41 0.0159095301081293
42 0.0236710217888257
43 0.0171220573237343
44 0.0215180760531819
45 0.0225745735207762
46 0.0207519071036109
47 0.0208192274478002
48 0.0185060474345315
49 0.0186673697061269
50 0.0165968914495401
51 0.0189938200286901
52 0.0196062311097589
53 0.0163620338622262
54 0.0157711132013245
55 0.0183572316344298
56 0.0235417418863451
57 0.0167381923342526
58 0.0216793035446431
59 0.0186802519977896
60 0.0235879435663819
61 0.029305813090461
62 0.0191144734629785
63 0.0213763399076369
64 0.0259190222680561
65 0.0230843996605689
66 0.024606711061094
67 0.021024715641375
68 0.0305408155758258
69 0.0277756506783866
70 0.0292867257917219
71 0.0309056042096751
72 0.0287450336253914
73 0.0261471801041264
74 0.0328464198179652
75 0.035758769451583
76 0.0368132740436069
77 0.0346984214147854
78 0.0325832613266726
79 0.0329910135434446
80 0.0381878736536124
81 0.0402815748511781
82 0.0327371781217881
83 0.0397206411254829
84 0.0400731909908557
85 0.041827469811096
86 0.0437294203535476
87 0.0443448672894271
88 0.0458260737710442
89 0.0459532521280476
90 0.0450868078588252
91 0.0485975779364677
92 0.0488075122846465
93 0.0533188235846197
94 0.0491933923462117
95 0.0555029839704818
96 0.0505308242607537
97 0.0494737924674722
98 0.0538515679284275
99 0.0519521324487728
100 0.0589971379338984
} coordinate[pos=1] (p);
\draw[<-,shorten <=2pt] (p) -- ++(0.2cm,-0.1cm) node[linelabel,right] {Gref};
\addlegendentry{Gref}
\addplot [line2]
table {%
40 0.0179493583208887
41 0.0153373597575195
42 0.0221003372650162
43 0.015286528617339
44 0.0233262259111364
45 0.0222628330693872
46 0.0177761088699152
47 0.0220603968294492
48 0.0180653736064066
49 0.0168601311857187
50 0.0176151726995401
51 0.0177159401267292
52 0.0224683855568744
53 0.0168145323881696
54 0.0163143170497042
55 0.019026549816248
56 0.0247695250894701
57 0.015870593650042
58 0.0219343884045568
59 0.0196967708007557
60 0.0209227417434653
61 0.0279892147298053
62 0.0200744659024946
63 0.0228471360385894
64 0.0240900427758686
65 0.022942152064415
66 0.0255278758338213
67 0.0202276074324197
68 0.0294400860261935
69 0.0260698365932416
70 0.0266565081577933
71 0.0308593344033371
72 0.028694903417058
73 0.0254812714996743
74 0.03137046437033
75 0.036996660076583
76 0.0353581702195937
77 0.0321695050186815
78 0.0323771064789162
79 0.0343551106557864
80 0.0375544263879874
81 0.0393668371968571
82 0.0331593151263613
83 0.0376662124733745
84 0.0408356537884747
85 0.0409093402155078
86 0.0441581540018034
87 0.0417095628713236
88 0.0437553617682032
89 0.0460981345016431
90 0.0444582922338252
91 0.0473036648183358
92 0.04908468619769
93 0.0519717553722541
94 0.0487449216015309
95 0.0524636542007451
96 0.049772849000337
97 0.0485885234597402
98 0.0513252804603153
99 0.0506532648667021
100 0.0564197941838984
} coordinate[pos=1] (p);
\draw[<-,shorten <=2pt] (p) -- ++(0.2cm,-0.3cm) node[linelabel,right] {JM};
\addlegendentry{JM}
\addplot [line3]
table {%
40 0.0159824149615137
41 0.0139367309465439
42 0.0185703428453734
43 0.012045681051932
44 0.0182571119196592
45 0.0165831238679983
46 0.0163936360438282
47 0.0152787312709385
48 0.0153460742900003
49 0.0161253775186268
50 0.01317105160579
51 0.0151766478473175
52 0.0170432991686532
53 0.0136095279659998
54 0.0132115428888245
55 0.0138119191344298
56 0.020064656227863
57 0.014319264154428
58 0.017601178544643
59 0.0147165999850777
60 0.0183795776809653
61 0.025988382250297
62 0.0161094961444301
63 0.019107304689383
64 0.0211461340844624
65 0.0204543816317227
66 0.0217311783906394
67 0.0180549745033152
68 0.0279650170923699
69 0.0262622618649808
70 0.024282903247079
71 0.0314767694825624
72 0.0277097146149748
73 0.0247309504380305
74 0.0300751729851948
75 0.0378377017432497
76 0.0345597532788043
77 0.0338599727540712
78 0.0315343330414162
79 0.0320884720086345
80 0.0384238111536125
81 0.0395419191798818
82 0.0326029365059344
83 0.0377516322775914
84 0.0413240280444271
85 0.0406394413184489
86 0.0470999917561639
87 0.0455415267721857
88 0.0481134272866691
89 0.0468295907867555
90 0.0483024545602141
91 0.0519294804089951
92 0.0524171632697009
93 0.0575870947539745
94 0.05338922429435
95 0.0579230703191661
96 0.0531910008558058
97 0.0528027832857712
98 0.0566465097332745
99 0.0580753371520556
100 0.0645201848088984
} coordinate[pos=1] (p);
\draw[<-,shorten <=2pt] (p) -- ++(0.2cm,+0.2cm) node[linelabel,right] {Ours};
\addlegendentry{Ours}
\end{axis}

\end{tikzpicture}

%% file: paper.bbl
\begin{thebibliography}{25}
\expandafter\ifx\csname natexlab\endcsname\relax\def\natexlab#1{#1}\fi

\bibitem[{Abney et~al.(1999)Abney, McAllester, and Pereira}]{abney+al:1999}
Steven Abney, David McAllester, and Fernando Pereira. 1999.
\newblock \href {https://doi.org/10.3115/1034678.1034759} {Relating
  probabilistic grammars and automata}.
\newblock In \emph{Proc. ACL}, pages 542--549.

\bibitem[{Aguinaga et~al.(2019)Aguinaga, Chiang, and Weninger}]{aguinaga:2018}
Salvador Aguinaga, David Chiang, and Tim Weninger. 2019.
\newblock \href {https://doi.org/10.1109/TPAMI.2018.2810877} {Learning
  hyperedge replacement grammars for graph generation}.
\newblock \emph{IEEE Trans. Pattern Analysis and Machine Intelligence},
  41(3):625--638.

\bibitem[{Autebert et~al.(1997)Autebert, Berstel, and Boasson}]{autebert+:1997}
Jean-Michel Autebert, Jean Berstel, and Luc Boasson. 1997.
\newblock \href {https://doi.org/10.1007/978-3-642-59136-5_3} {Context-free
  languages and pushdown automata}.
\newblock In Grzegorz Rozenberg and Arto Salomaa, editors, \emph{Handbook of
  Formal Languages}, pages 111--174. Springer.

\bibitem[{Earley(1970)}]{earley:1970}
Jay Earley. 1970.
\newblock \href {https://doi.org/10.1145/362007.362035} {An efficient
  context-free parsing algorithm}.
\newblock \emph{Comm. ACM}, 13(2):94--102.

\bibitem[{Futrell et~al.(2019)Futrell, Wilcox, Morita, Qian, Ballesteros, and
  Levy}]{futrell+al:2018}
Richard Futrell, Ethan Wilcox, Takashi Morita, Peng Qian, Miguel Ballesteros,
  and Roger Levy. 2019.
\newblock \href {https://doi.org/10.18653/v1/N19-1004} {Neural language models
  as psycholinguistic subjects: Representations of syntactic state}.
\newblock In \emph{Proc. NAACL HLT}, pages 32--42.

\bibitem[{Glorot and Bengio(2010)}]{glorot+bengio:2010}
Xavier Glorot and Yoshua Bengio. 2010.
\newblock \href {http://proceedings.mlr.press/v9/glorot10a/glorot10a.pdf}
  {Understanding the difficulty of training deep feedforward neural networks}.
\newblock In \emph{Proc. AISTATS}, pages 249--256.

\bibitem[{Goodman(1999)}]{goodman:1999}
Joshua Goodman. 1999.
\newblock \href {http://dl.acm.org/citation.cfm?id=973226.973230} {Semiring
  parsing}.
\newblock \emph{Computational Linguistics}, 25(4):573--605.

\bibitem[{Grefenstette et~al.(2015)Grefenstette, Hermann, Suleyman, and
  Blunsom}]{grefenstette+al:2015}
Edward Grefenstette, Karl~Moritz Hermann, Mustafa Suleyman, and Phil Blunsom.
  2015.
\newblock \href
  {https://papers.nips.cc/paper/5648-learning-to-transduce-with-unbounded-memory.pdf}
  {Learning to transduce with unbounded memory}.
\newblock In \emph{Proc. NeurIPS}, volume~2, pages 1828--1836.

\bibitem[{Greibach(1973)}]{greibach:1973}
Sheila~A. Greibach. 1973.
\newblock \href {https://doi.org/10.1137/0202025} {The hardest context-free
  language}.
\newblock \emph{SIAM J. Comput.}, 2(4):304--310.

\bibitem[{Hao et~al.(2018)Hao, Merrill, Angluin, Frank, Amsel, Benz, and
  Mendelsohn}]{hao+al:2018}
Yiding Hao, William Merrill, Dana Angluin, Robert Frank, Noah Amsel, Andrew
  Benz, and Simon Mendelsohn. 2018.
\newblock \href {https://doi.org/10.18653/v1/W18-5433} {Context-free
  transductions with neural stacks}.
\newblock In \emph{Proc. {B}lackbox{NLP}}, pages 306--315.

\bibitem[{Hu et~al.(2020)Hu, Gauthier, Qian, Wilcox, and Levy}]{hu+al:2020}
Jennifer Hu, Jon Gauthier, Peng Qian, Ethan Wilcox, and Roger Levy. 2020.
\newblock \href {https://www.aclweb.org/anthology/2020.acl-main.158} {A
  systematic assessment of syntactic generalization in neural language models}.
\newblock In \emph{Proc. ACL}, pages 1725--1744.

\bibitem[{Joulin and Mikolov(2015)}]{joulin+mikolov:2015}
Armand Joulin and Tomas Mikolov. 2015.
\newblock \href
  {https://papers.nips.cc/paper/5857-inferring-algorithmic-patterns-with-stack-augmented-recurrent-nets.pdf}
  {Inferring algorithmic patterns with stack-augmented recurrent nets}.
\newblock In \emph{Proc. NeurIPS}, volume~1, pages 190--198.

\bibitem[{Kingma and Ba(2015)}]{kingma+ba:2015}
Diederik~P. Kingma and Jimmy~Lei Ba. 2015.
\newblock \href {https://arxiv.org/abs/1412.6980} {{A}dam: A method for
  stochastic optimization}.
\newblock In \emph{Proc. ICLR}.

\bibitem[{Lang(1974)}]{lang:1974}
Bernard Lang. 1974.
\newblock \href {https://doi.org/10.1007/978-3-662-21545-6_18} {Deterministic
  techniques for efficient non-deterministic parsers}.
\newblock In \emph{Proc.~Colloquium on Automata, Languages, and Programming},
  pages 255--269.

\bibitem[{Levy(2008)}]{levy:2008}
Roger Levy. 2008.
\newblock \href {https://doi.org/10.1016/j.cognition.2007.05.006}
  {Expectation-based syntactic comprehension}.
\newblock \emph{Cognition}, 106:1126--77.

\bibitem[{McCoy et~al.(2020)McCoy, Frank, and Linzen}]{mccoy+al:2020}
Richard McCoy, Robert~H. Frank, and Tal Linzen. 2020.
\newblock \href {https://www.mitpressjournals.org/doi/pdf/10.1162/tacl_a_00304}
  {Does syntax need to grow on trees? {S}ources of hierarchical inductive bias
  in sequence-to-sequence networks}.
\newblock \emph{Trans. ACL}, 8:125--140.

\bibitem[{Paszke et~al.(2019)Paszke, Gross, Massa, Lerer, Bradbury, Chanan,
  Killeen, Lin, Gimelshein, Antiga, Desmaison, Kopf, Yang, DeVito, Raison,
  Tejani, Chilamkurthy, Steiner, Fang, Bai, and Chintala}]{pytorch}
Adam Paszke, Sam Gross, Francisco Massa, Adam Lerer, James Bradbury, Gregory
  Chanan, Trevor Killeen, Zeming Lin, Natalia Gimelshein, Luca Antiga, Alban
  Desmaison, Andreas Kopf, Edward Yang, Zachary DeVito, Martin Raison, Alykhan
  Tejani, Sasank Chilamkurthy, Benoit Steiner, Lu~Fang, Junjie Bai, and Soumith
  Chintala. 2019.
\newblock \href
  {http://papers.neurips.cc/paper/9015-pytorch-an-imperative-style-high-performance-deep-learning-library.pdf}
  {{P}y{T}orch: An imperative style, high-performance deep learning library}.
\newblock In \emph{Proc. NeurIPS}, pages 8024--8035.

\bibitem[{van Schijndel et~al.(2019)van Schijndel, Mueller, and
  Linzen}]{schijndel+al:2019}
Marten van Schijndel, Aaron Mueller, and Tal Linzen. 2019.
\newblock \href {https://doi.org/10.18653/v1/D19-1592} {Quantity doesn{'}t buy
  quality syntax with neural language models}.
\newblock In \emph{Proc. EMNLP-IJCNLP}, pages 5831--5837.

\bibitem[{Shieber et~al.(1995)Shieber, Schabes, and Pereira}]{shieber+:1995}
Stuart~M. Shieber, Yves Schabes, and Fernando C.~N. Pereira. 1995.
\newblock \href {https://doi.org/https://doi.org/10.1016/0743-1066(95)00035-I}
  {Principles and implementation of deductive parsing}.
\newblock \emph{Journal of Logic Programming}, 24(1):3--36.

\bibitem[{Stolcke(1995)}]{stolcke:1995}
Andreas Stolcke. 1995.
\newblock \href {http://www.aclweb.org/anthology/J95-2002.pdf} {An efficient
  probabilistic context-free parsing algorithm that computes prefix
  probabilities}.
\newblock \emph{Computational Linguistics}, 21(2):165--201.

\bibitem[{Sun et~al.(1995)Sun, Giles, Chen, and Lee}]{sun+al:1995}
G.~Z. Sun, C.~Lee Giles, H.~H. Chen, and Y.~C. Lee. 1995.
\newblock \href {https://arxiv.org/abs/1711.05738} {The neural network pushdown
  automaton: Model, stack, and learning simulations}.
\newblock Technical Report UMIACS-TR-93-77 and CS-TR-3118, University of
  Maryland.
\newblock Revised version.

\bibitem[{Suzgun et~al.(2019)Suzgun, Gehrmann, Belinkov, and
  Shieber}]{suzgun+:2019}
Mirac Suzgun, Sebastian Gehrmann, Yonatan Belinkov, and Stuart~M. Shieber.
  2019.
\newblock \href {https://arxiv.org/abs/1911.03329} {Memory-augmented recurrent
  neural networks can learn generalized {D}yck languages}.
\newblock {a}rXiv:1922.03329.

\bibitem[{Tomita(1987)}]{tomita:1987}
Masaru Tomita. 1987.
\newblock \href {https://www.aclweb.org/anthology/J87-1004.pdf} {An efficient
  augmented context-free parsing algorithm}.
\newblock \emph{Computational Linguistics}, 13(1--2):31--46.

\bibitem[{Wilcox et~al.(2019)Wilcox, Levy, and Futrell}]{wilcox+al:2019}
Ethan Wilcox, Roger Levy, and Richard Futrell. 2019.
\newblock \href {https://doi.org/10.18653/v1/W19-4819} {Hierarchical
  representation in neural language models: Suppression and recovery of
  expectations}.
\newblock In \emph{Proc. BlackboxNLP}, pages 181--190.

\bibitem[{Yogatama et~al.(2018)Yogatama, Miao, Melis, Ling, Kuncoro, Dyer, and
  Blunsom}]{yogatama+al:2018}
Dani Yogatama, Yishu Miao, G{\'a}bor Melis, Wang Ling, Adhiguna Kuncoro, Chris
  Dyer, and Phil Blunsom. 2018.
\newblock \href {https://openreview.net/pdf?id=SkFqf0lAZ} {Memory architectures
  in recurrent neural network language models}.
\newblock In \emph{Proc. ICLR}.

\end{thebibliography}
